%% file: main.tex
\documentclass[11pt]{article}
\usepackage{ra}

\begin{document}

\ratitleblock
  {RelateAnything: Real-Time Open-Vocabulary\\Relation Prediction From Any Inputs}
  {Maëlic Neau}
  {Independent Researcher \\ Correspondence: \texttt{maelicneau@gmail.com}}
  {\input{sec/00_abstract}}
  {Code: \url{https://github.com/Maelic/RelateAnything}\\
   Models: \url{https://huggingface.co/collections/maelic/relateanything}\\
   Dataset: \url{https://huggingface.co/datasets/maelic/RA-4M}}

\input{sec/01_intro}
\input{sec/02_related}
\input{sec/03_model}
\input{sec/04_corpus}
\input{sec/05_protocol}
\input{sec/06_results}
\input{sec/07_decoupling}
\input{sec/08_datalessons}
\input{sec/09_efficiency}
\input{sec/10_discussion}
\input{sec/11_conclusion}

\clearpage
{\small\bibliography{main}}

\clearpage
\appendix
\input{app/a_arch}
\clearpage
\input{app/b_objective}
\clearpage
\input{app/c_corpus}
\clearpage
\input{app/d_protocol}
\clearpage
\input{app/e_results}
\clearpage
\input{app/f_probes}
\clearpage
\input{app/g_scaling}
\clearpage
\input{app/h_ablations}
\clearpage
\input{app/i_cost}
\clearpage
\input{app/j_prompts}

\end{document}

%% file: sec/00_abstract.tex
Open-vocabulary object detection accepts an arbitrary class list at inference,
and promptable segmentation returns regions carrying no class names at all: in
both, the taxonomy has left the model and become an input. Relation prediction
has not made that transition. Scene-graph models are still trained and
evaluated on the 50 or 56 predicates of a single annotation style, with the
relation head conditioned on object class labels and therefore tied to one
detector and one label space. Three obstacles account for this, and they are
not primarily modelling ones: no relation corpus is both free-text and
verified, an architecture conditioned on labels cannot accept a vocabulary it
was not trained on, and the standard metric rewards agreement with the training
corpus, so that enlarging the vocabulary is scored as a regression.

We present \textbf{\model{}}, a 53M-parameter relation model that takes an
image and a set of regions from any source and returns scored relations over a
predicate vocabulary supplied at inference as a list of strings. Object class
labels are never an input, so the region source can be replaced without
retraining, and the vocabulary is a bank of text embeddings rather than a
learned classifier, so changing it is a matrix substitution. The model runs at
20 ms per frame on a single GPU. Training over 19{,}103 predicates requires
two departures from standard practice: the supervision must be treated as
positive-unlabeled, since at this vocabulary size an unannotated predicate is
frequently true, and the text encoder must be corrected, since contrastive
encoders embed antonyms such as \pred{above} and \pred{below} at a cosine
similarity of 0.95.

To supply the supervision we build \textbf{\dataset{}}, a corpus of 474k images
and 4.3M relations over 10{,}102 free-text predicates, generated by a
vision--language model against numbered box markers and filtered by a
deterministic geometric check. To make the result measurable we build
\textbf{\bench{}}, whose six axes are each scored across datasets and none
satisfiable by the priors that standard recall rewards: a frequency table over
ground-truth object categories, which sees no pixels, exceeds a trained model
on the metric by which leaderboards are ordered.

On three benchmarks evaluated cross-dataset and a fourth evaluated
zero-shot, \model{} outperforms the strongest open-vocabulary method of comparable
scale, with
$2.3$--$3.5\times$ the mean recall, and the margins persist when a real
detector supplies the regions. Against a scene-graph model built on a 3B
vision--language model it leads on both metrics on all three, at under 2\% of
the parameters, and trails on micro recall only when both are scored through a
free-text matcher. In-domain measurement overstates cross-dataset
gains by approximately $5\times$. The model, the corpus and the benchmark are
publicly released.

%% file: sec/01_intro.tex
\section{Introduction}
\label{sec:intro}

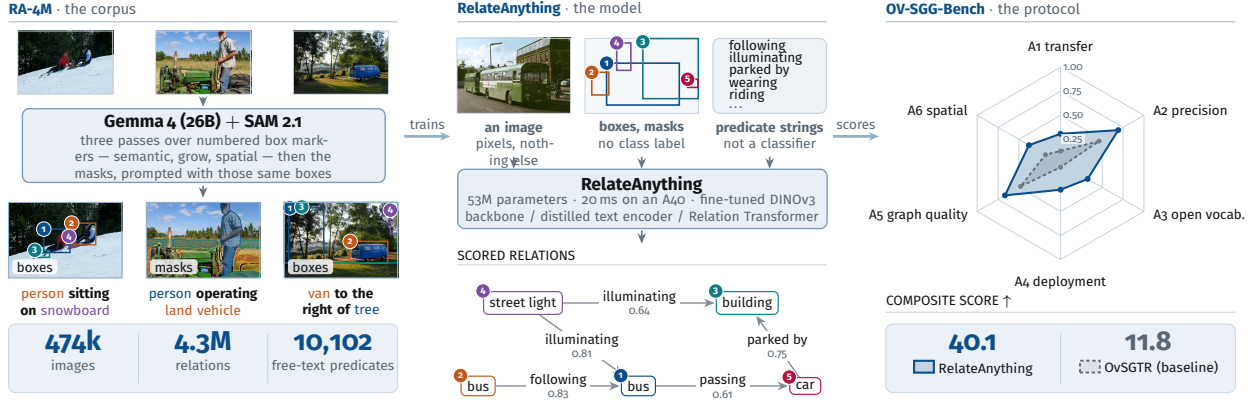
\begin{figure*}[t]
\centerline{\resizebox{1.0\textwidth}{!}{\input{figures/overview.tex}}}
\caption{\textbf{Overview.} \emph{Left}, \dataset{}: three corpus images, the
annotator, and one verified free-text relation per image. \emph{Middle},
\model{}: its three inputs are separable (pixels, regions from any source,
predicate strings); the regions and the five scored relations are model
output, and none of the predicates is among the 50 of VG150 \citep{xu2017}.
\emph{Right}, \bench{}: six chance-corrected axes (centre chance, rim perfect,
rings 0.25 apart) and the composite over A1, A2, A4, A5 and A6. The baseline
drawn is OvSGTR, the one system runnable on every axis and so the only one that can carry a composite; on A1 the stronger baseline is ROBIN-3B, and
\cref{tab:sixaxes} reports the best available on each axis. A5 is
\cref{eq:truebits} at equal graph length (\cref{sec:results:a5}).}
\label{fig:overview}
\end{figure*}

Open-vocabulary object detection accepts an arbitrary class list at inference
\citep{vild,detic,owlvit,glip,groundingdino}, including at real-time rates
\citep{yoloworld,yoloe}, and promptable segmentation returns regions carrying
no class names at all \citep{sam,fastsam}. In both cases the taxonomy has left
the model and become an input, which is what allows either to be pointed at a
domain it was not built for. Relation prediction has not made that transition yet.
Scene-graph generation, the literature that would supply it, is still trained
and evaluated on the 50 predicates of VG150 \citep{xu2017} or the 56 of PSG
\citep{psg}, drawn from one annotation style; its dominant architecture
predicts objects and relations from a shared backbone and conditions the
relation head on predicted object classes \citep{motifs,vctree,ovsgtr}, which
fixes the label space and excludes the region sources a deployed system
actually has, such as a different detector, a class-agnostic segmenter or a
human annotation; and where the vocabulary is open at all, it arrives as a
single caption whose predicate names are encoded jointly, which bounds it at
roughly 150 strings and keeps a language model in the inference path.


Three obstacles have kept relation prediction at the scale of a benchmark
vocabulary, and none of them is primarily a modelling obstacle. The first is
that the supervision does not exist. Relation annotation is the most expensive
and least reliable form of annotation in vision, and the machine-generated
corpora that do exist either project onto a fixed class list or leave the
annotator unchecked (\cref{tab:gen} in \cref{app:corpus}). The second is that
the architecture cannot use it: a relation head conditioned on object labels
inherits a label space, and a vocabulary encoded as one caption cannot be
enlarged. The third, and the one that has received least attention, is that the
measurement cannot see it. Scene-graph recall is computed against a benchmark
whose predicate vocabulary is also its training vocabulary, so it rewards
agreement with a corpus rather than the relation in the image, and a model that
enlarges its vocabulary is penalised for answering with a string the annotation
does not use. In this work, we present a model, a corpus and a benchmark that together remove all three obstacles, and we measure the effect of each on the other two (\cref{fig:overview}).

\paragraph{A relation model whose inputs are separable}
We present \model{}, a 53M-parameter relation model that takes an image and a
set of regions from any source and returns scored relations between pairs of
those regions, over a predicate vocabulary supplied at inference as a list of
strings (\cref{fig:overview}). Two properties of that interface account for
most of the design. Object class labels are never an input, at any stage, so
the region source can be replaced without retraining and the model cannot
inherit a detector's label distribution; and the vocabulary is not a learned
classifier but a bank of text embeddings that no learned layer transforms, so
changing it is a matrix substitution rather than a retraining. The relation model runs
at 20 ms per frame on a single GPU, alongside a real-time object detector \citep{yoloe}. Training it over 19{,}103 free-text predicates, rather
than the few dozen usual in this literature, changes the learning problem in
two ways that must both be treated. The supervision becomes
positive-unlabeled, because at this vocabulary size a pair annotated
\pred{riding} is also, unstated, \pred{sitting on}, and penalising the
unstated predicate teaches the model to suppress correct answers. We address
this with a batch-local InfoNCE loss whose positive is the annotated
predicate's synonym group and whose negatives are discounted by an estimated
probability, fitted on the corpus's multiply-annotated pairs, that they are
also true.
The target text space itself is also defective: contrastive text encoders embed antonyms
such as \pred{above} and \pred{below} at a cosine similarity of 0.95, which
upper-bounds what any model regressing onto those directions can express,
whoever trains it. We correct the text encoder by distilling a student whose
objective adds an antonym-repulsion term, so that inverse pairs separate
while synonym groups stay together.

\paragraph{Supervision without a label budget}
Such a model requires a corpus whose vocabulary is not that of a benchmark and
whose annotation is dense enough that unannotated pairs are rare. An image with
$N$ objects contains $N(N-1)$ candidate pairs, and for most of them whether the
relation holds is a matter of judgement, which is why the community kept the 50
most frequent of the 36{,}549 predicate strings Visual Genome's annotators wrote
\citep{vg,xu2017}, and why the discarded remainder is exactly the part an
open-vocabulary model is meant to serve. A vision--language model (VLM) is what makes
the alternative affordable. We generate
\dataset{}, a corpus of 474k images and 4.3M relations over 10{,}102 free-text
predicates. The VLM annotates against numbered markers placed on the boxes,
so that grounding is an input to the annotator rather than an inference from
its output, and every proposed relation then passes a deterministic geometric
gate that rejects what box geometry contradicts and leaves what geometry
cannot constrain unchecked and counted. On the same images and the same boxes
as the annotations it replaces, the result is $1.7\times$ denser with
$107\times$ the vocabulary.

\paragraph{Measuring relation prediction rather than corpus agreement}
Measurement has to come first, since a protocol that rewards agreement with a
corpus cannot register whether either of the other obstacles has been removed.
Scene-graph recall proves to be, to a large extent, a prior-matching score
(\cref{sec:priors}): a frequency table over ground-truth object categories,
given no pixels, outperforms a trained model on the metric that orders
leaderboards while falling well below it per predicate; the vocabulary a
training corpus shares with a benchmark predicts recall on that benchmark; and
a model that learns which pairs an annotator wrote down learns the annotator
rather than the image. Two scoring conventions amplify these effects
(\cref{sec:protocol}), a matcher that assigns detections to ground-truth
objects without constraint and a detector operating point that goes unreported
yet moves attainable recall by more than the spread between published methods.
Each finding constrains the design of \bench{}. Because shared triplet mass
predicts recall, every axis is scored across datasets and none of the benchmarks
contributes a training image, a setting scene-graph generation has left
largely unexplored, since results there are almost always reported in-domain
\citep{sggsurvey}. Because a model can profit from what an annotator chose to
write down, two axes are scored against negatives that a human adjudicated, on
which a confident false positive costs as much as a miss. And because each
axis can be satisfied on its own by a different shortcut, whether corpus
match, uniform caution, collapse onto the frequent predicates or the operating
point of the detector, the six are reported together and aggregated in a way
that penalises imbalance across them.

\paragraph{Overview of contributions}
The paper contributes (i) \model{} (\cref{sec:model}), the relation model
described above, together with the objective that makes a five-figure
predicate vocabulary trainable; (ii) \dataset{} (\cref{sec:corpus}), the first
relation corpus we are aware of that is both free-text and verified against
the geometry of the boxes it names; (iii) \bench{} (\cref{sec:bench}), the
evaluation protocol and the measurements of prior and of scoring convention
that motivate it; and (iv) an empirical study
(\cref{sec:results,sec:decouple,sec:datalessons,sec:efficiency}) organised
around three questions, namely what scene-graph recall measures
(\cref{sec:bench}), whether relation prediction benefits from being coupled to
object prediction (\cref{sec:decouple}), and whether in-domain measurement
predicts cross-dataset transfer (\cref{sec:datalessons}). On three benchmarks
evaluated cross-dataset and a fourth evaluated zero-shot, \model{}
outperforms OvSGTR \citep{ovsgtr}, the strongest open-vocabulary method of
comparable scale, on every recall and precision metric, with mean recall higher by a
factor of $2.3$--$3.5$ and rare-predicate recall higher by a factor of
$5$--$21$, and the margins persist with the same structure when a real
detector supplies the regions. On the question of coupling we find that
relation supervision sharpens object identity in the dense features without
creating relational ones, and that the relation score is 87--93\% pair context
against 0.1\% object identity, so there is no shared representation to justify
a shared backbone. On the question of measurement we find that in-domain gains
overstate cross-dataset gains by approximately $5\times$, and that a change
selected in-domain can lose out of it. Model, corpus and benchmark are public.

\paragraph{Evaluation setting}
All results in this paper are obtained under cross-dataset transfer, on
benchmarks the model was not trained on, with four marked exceptions: the
HICO-DET row trained on a 5\% relation share of its training split
(\cref{tab:a1}) and the three fine-tuned rows of \cref{tab:react}.
\Cref{sec:datalessons} shows that in-domain measurement can favour changes
that reduce transfer performance.

%% file: figures/overview.tex
\input{figures/overview_data}%
\def\ftitle{\fontsize{8}{9}\selectfont\sffamily\bfseries}%
\def\fnote{\fontsize{6.2}{7.2}\selectfont\sffamily}%
\def\fsmall{\fontsize{7}{8}\selectfont\sffamily}%
\def\fband{\fontsize{6.8}{7.8}\selectfont\sffamily\bfseries}%
\def\fbig{\fontsize{12.5}{13}\selectfont\sffamily\bfseries}%
%
\def\ZaX{0.0}\def\ZaW{5.90}
\def\ZbX{6.80}\def\ZbW{5.60}
\def\ZcX{13.30}\def\ZcW{5.30}
\def\BandY{6.36}
\def\ArrY{4.55}
%
\def\PW{3.70}\def\PH{2.467}\def\PDx{0.60}\def\PDy{0.66}%
\def\PfX{0.0}\def\PfY{3.70}
\newcommand{\cpx}[2]{({\PfX+#1*\PW},{\PfY+#2*\PH})}
%
\def\RCX{15.95}\def\RCY{4.14}\def\RR{1.46}%
\newcommand{\rv}[2]{({\RCX+\RR*#2*cos(#1)},{\RCY+\RR*#2*sin(#1)})}%
\def\AngA{90}\def\AngB{30}\def\AngC{-30}%
\def\AngD{-90}\def\AngE{-150}\def\AngF{150}%

\begin{tikzpicture}[x=1cm, y=1cm,
  font=\fnote,
  >={Stealth[length=1.6mm, width=1.3mm]},
  flow/.style={->, line width=0.7pt, draw=raMuted!90},
  big/.style={->, line width=1.2pt, draw=raAccent!45},
  panel/.style={draw=raRule, fill=raPanel, rounded corners=3pt, line width=0.5pt},
  mod/.style={draw=raAccent!55, fill=raAccent!9, rounded corners=3pt,
              line width=0.6pt, align=center, inner sep=3pt},
  title/.style={font=\ftitle, text=raInk, align=center, inner sep=1pt},
  note/.style={font=\fnote, text=raMuted, align=center, inner sep=1pt},
  lnote/.style={note, align=left},
  band/.style={font=\fband, text=raAccent, anchor=south west, inner sep=0pt},
  chip/.style={draw=raRule, fill=white, rounded corners=1.5pt, anchor=west,
               inner xsep=2.4pt, inner ysep=1.5pt, font=\fnote, text=raInk},
  ptag/.style={font=\fnote, text=raInk, fill=white, fill opacity=0.82,
               text opacity=1, inner xsep=2pt, inner ysep=1pt,
               rounded corners=1.5pt},
  pass/.style={font=\fnote, text=raAccent, fill=raAccent!10, inner xsep=2.4pt,
               inner ysep=1.2pt, rounded corners=1.5pt},
  ovbadge/.style={circle, draw=white, line width=0.35pt, inner sep=0pt,
                  minimum size=2.2mm,
                  font=\fontsize{4.4}{4.4}\selectfont\sffamily\bfseries, text=white},
  dot/.style={circle, inner sep=0pt, minimum size=2.9mm,
              font=\fontsize{5.4}{5.4}\selectfont\sffamily\bfseries, text=white},
  gchip/.style={fill=white, rounded corners=2pt, line width=0.5pt,
                inner xsep=2.6pt, inner ysep=2.0pt, font=\fnote, text=raInk},
  gedge/.style={->, line width=0.6pt, draw=raMuted!75, shorten >=0.5pt},
  glab/.style={font=\fnote, text=raInk, align=center, fill=white,
               inner xsep=1.6pt, inner ysep=1.0pt},
]

\node[band] at (\ZaX,\BandY) {\dataset{} \textcolor{raMuted}{\mdseries$\cdot$ the corpus}};
\draw[raAccent!35, line width=0.5pt] (\ZaX,{\BandY-0.07}) -- ({\ZaX+\ZaW},{\BandY-0.07});
\ovRegionColours

\def\AW{1.72}\def\AH{1.147}\def\AStep{2.09}%
\def\UY{5.19}
\def\AY{2.40}

\def\UW{1.44}\def\UH{0.96}%
\node[anchor=south west, inner sep=0] at ({0*\AStep+(\AW-\UW)/2},\UY)
  {\includegraphics[width=\UW cm]{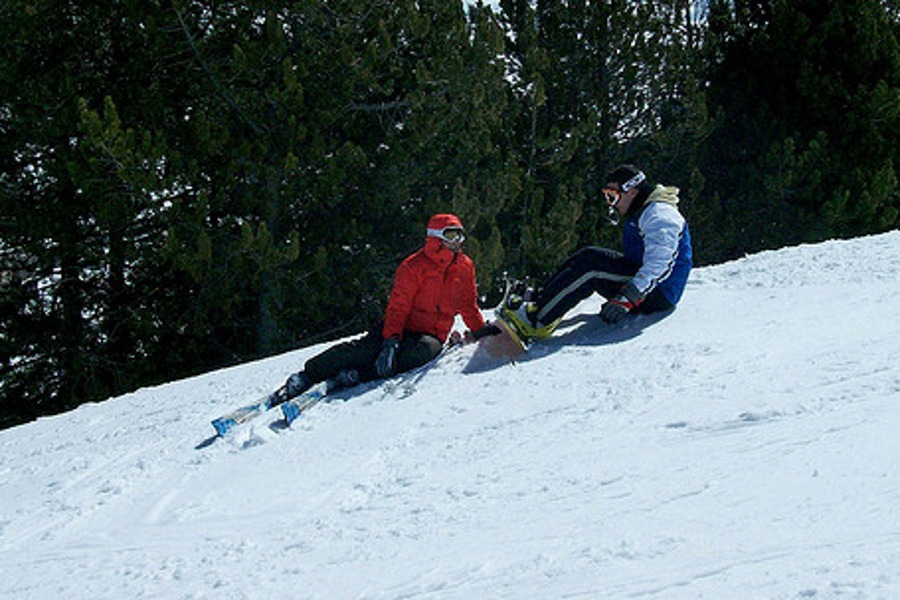}};
\draw[raRule, line width=0.4pt] ({0*\AStep+(\AW-\UW)/2},\UY)
  rectangle ({0*\AStep+(\AW+\UW)/2},{\UY+\UH});
\node[anchor=south west, inner sep=0] at ({1*\AStep+(\AW-\UW)/2},\UY)
  {\includegraphics[width=\UW cm]{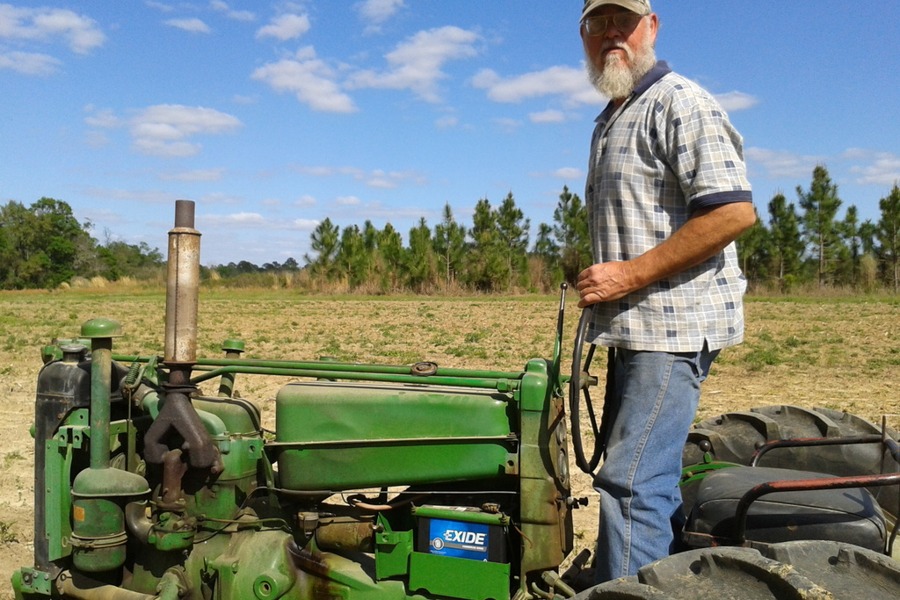}};
\draw[raRule, line width=0.4pt] ({1*\AStep+(\AW-\UW)/2},\UY)
  rectangle ({1*\AStep+(\AW+\UW)/2},{\UY+\UH});
\node[anchor=south west, inner sep=0] at ({2*\AStep+(\AW-\UW)/2},\UY)
  {\includegraphics[width=\UW cm]{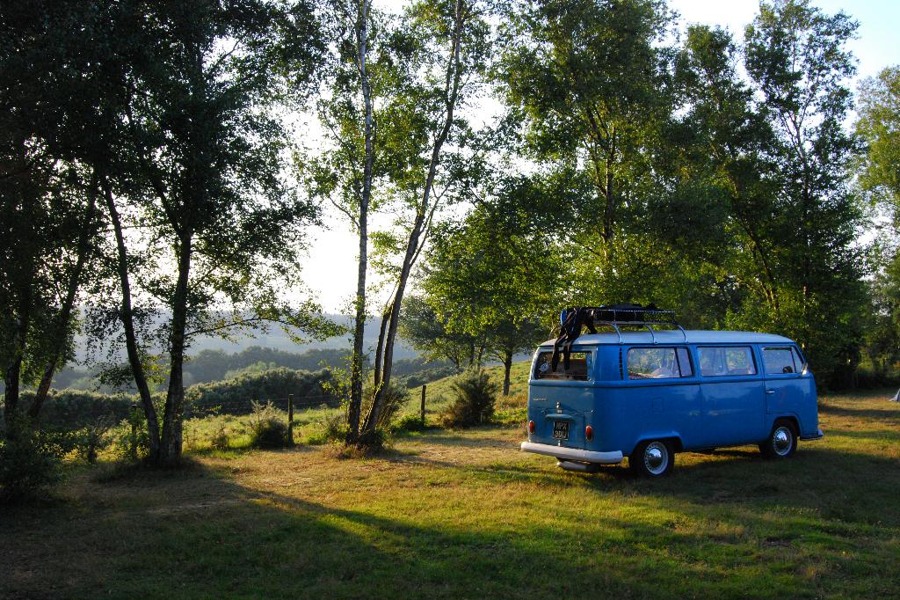}};
\draw[raRule, line width=0.4pt] ({2*\AStep+(\AW-\UW)/2},\UY)
  rectangle ({2*\AStep+(\AW+\UW)/2},{\UY+\UH});

\draw[flow] (2.95,5.15) -- (2.95,5.00);
\node[mod, anchor=south west, minimum width=5.50cm, minimum height=1.12cm]
  at (0.20,3.84) {};
\node[title] at (2.95,4.78) {\ovGenModel{} (\ovGenSize{}) $+$ \ovSegmenter{}};
\node[note, anchor=north, text width=5.24cm] at (2.95,4.62)
  {\hyphenpenalty=10000 three passes over numbered box markers ---
   \ovGenPasses{} --- then the masks, prompted with those same boxes};
\draw[flow] (2.95,3.80) -- (2.95,3.62);

\def\colX{0}%
\newcommand{\bxp}[2]{({\colX+#1*\AW},{\AY+#2*\AH})}%
\newcommand{\ovpanel}[4]{%
  \pgfmathsetmacro{\colX}{(#1-1)*\AStep}%
  \global\let\colX\colX
  \node[anchor=south west, inner sep=0] at (\colX,\AY)
    {#2};
  #4%
  \draw[raAccent!45, line width=0.5pt]
    (\colX,\AY) rectangle ({\colX+\AW},{\AY+\AH});%
  \node[ptag, anchor=south west] at ({\colX+0.05},{\AY+0.05}) {#3};%
}
\newcommand{\ovtext}[8]{%
  \node[note, anchor=north, text width={\AW cm}] at ({\colX+\AW/2},{\AY-0.10})
    {\textcolor{ovr#2}{#3} \textbf{\textcolor{raInk}{#4}} \textcolor{ovr#6}{#5}};%
}
\ovCorpusRows

\draw[fill=raAccent!8, draw=raAccent!22, rounded corners=3pt, line width=0.4pt]
  (\ZaX,0.66) rectangle ({\ZaX+\ZaW},1.70);
\foreach \x in {1.97, 3.93}
  \draw[raAccent!25, line width=0.4pt] (\x,0.80) -- (\x,1.56);
\newcommand{\ovbig}[3]{%
  \node[font=\fbig, text=raAccent, anchor=base] at (#1,1.30) {#2};%
  \node[note, anchor=north, text width=1.86cm] at (#1,1.20) {#3};%
}
\ovbig{0.98}{\ovImages{}k}{images}
\ovbig{2.95}{\ovRelations{}M}{relations}
\ovbig{4.92}{\ovPredicates{}}{free-text predicates}

\node[band] at (\ZbX,\BandY) {\model{} \textcolor{raMuted}{\mdseries$\cdot$ the model}};
\draw[raAccent!35, line width=0.5pt] (\ZbX,{\BandY-0.07}) -- ({\ZbX+\ZbW},{\BandY-0.07});
\draw[big] ({\ZaX+\ZaW+0.10},\ArrY) -- ({\ZbX-0.10},\ArrY);
\node[note, anchor=south] at ({(\ZaX+\ZaW+\ZbX)/2},{\ArrY+0.07}) {trains};

\def\TW{1.72}\def\TH{1.139}\def\TY{4.90}%
\def\TaX{6.80}\def\TbX{8.74}\def\TcX{10.68}\def\TcW{1.72}%
\newcommand{\apx}[2]{({\TbX+#1*\TW},{\TY+#2*\TH})}

\node[anchor=south west, inner sep=0] at (\TaX,\TY)
  {\includegraphics[width=\TW cm]{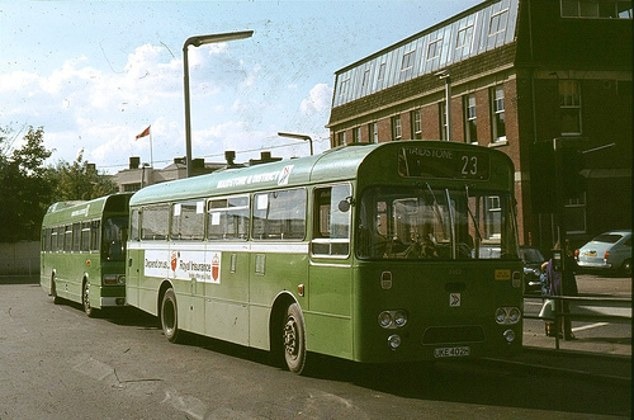}};
\draw[raRule, line width=0.4pt] (\TaX,\TY) rectangle ({\TaX+\TW},{\TY+\TH});
\node[note, anchor=north, text width={\TW cm}] at ({\TaX+\TW/2},{\TY-0.07})
  {\textbf{an image}\\ pixels, nothing else};

\definecolor{rg1}{HTML}{0B4F8A}\definecolor{rg2}{HTML}{C2571A}%
\definecolor{rg3}{HTML}{0E7C86}\definecolor{rg4}{HTML}{7A4E9E}%
\definecolor{rg5}{HTML}{B3123F}%
\draw[raRule, fill=raPanel!55, line width=0.4pt]
  (\TbX,\TY) rectangle ({\TbX+\TW},{\TY+\TH});
\draw[rg1, line width=0.7pt] \apx{0.1905}{0.1110} rectangle \apx{0.8314}{0.6693};
\draw[rg2, line width=0.7pt] \apx{0.0598}{0.2455} rectangle \apx{0.2096}{0.5471};
\draw[rg3, line width=0.7pt] \apx{0.5101}{0.1960} rectangle \apx{0.9984}{0.9971};
\draw[rg4, line width=0.7pt] \apx{0.2798}{0.5733} rectangle \apx{0.4043}{0.9383};
\draw[rg5, line width=0.7pt] \apx{0.9049}{0.3462} rectangle \apx{0.9959}{0.4543};
\node[ovbadge, fill=rg1] at \apx{0.1905}{0.6693} {1};
\node[ovbadge, fill=rg2] at \apx{0.0598}{0.5471} {2};
\node[ovbadge, fill=rg3] at \apx{0.5101}{0.9550} {3};
\node[ovbadge, fill=rg4] at \apx{0.2798}{0.9383} {4};
\node[ovbadge, fill=rg5] at \apx{0.9049}{0.4543} {5};
\node[note, anchor=north, text width={\TW cm}] at ({\TbX+\TW/2},{\TY-0.07})
  {\textbf{boxes, masks}\\ no class label};

\draw[panel] (\TcX,\TY) rectangle ({\TcX+\TcW},{\TY+\TH});
\def\VYo{1.02}%
\newcommand{\ovvocab}[2]{%
  \node[anchor=west, font=\fnote, text=raInk] at ({\TcX+0.11},{\TY+\VYo-#1*0.185}) {#2};%
}
\ovVocabList
\node[anchor=west, font=\fnote, text=raMuted]
  at ({\TcX+0.11},{\TY+\VYo-\ovVocabListN*0.185}) {\dots};
\node[note, anchor=north, text width={\TcW cm}] at ({\TcX+\TcW/2},{\TY-0.07})
  {\textbf{predicate strings}\\ not a classifier};

\foreach \x in {{\TaX+\TW/2}, {\TbX+\TW/2}, {\TcX+\TcW/2}}
  \draw[flow] ({\x},4.24) -- ({\x},4.06);
\node[mod, anchor=south west, minimum width=\ZbW cm, minimum height=0.90cm]
  at (\ZbX,3.14) {};
\node[title] at ({\ZbX+\ZbW/2},3.82) {\model{}};
\node[note, text width=5.36cm] at ({\ZbX+\ZbW/2},3.46)
  {\ovParams{}M parameters $\cdot$ \ovLatency{}\,ms on an \ovDevice{}
   $\cdot$ fine-tuned DINOv3 backbone / distilled text encoder / Relation Transformer};

\draw[flow] ({\ZbX+\ZbW/2},3.12) -- ({\ZbX+\ZbW/2},2.92);
\node[band, text=raInk, anchor=south west, font=\fnote] at (\ZbX,2.66)
  {SCORED RELATIONS};
\draw[raRule, line width=0.4pt] (\ZbX,2.60) -- ({\ZbX+\ZbW},2.60);

\coordinate (p1) at ({\ZbX+2.75},0.76);
\coordinate (p2) at ({\ZbX+0.32},0.76);
\coordinate (p3) at ({\ZbX+4.40},2.02);
\coordinate (p4) at ({\ZbX+1.00},2.02);
\coordinate (p5) at ({\ZbX+5.28},0.76);
\newcommand{\ovgnode}[2]{\node[gchip, draw=rg#1] (g#1) at (p#1) {#2};}
\newcommand{\ovgedge}[4]{%
  \draw[gedge] (g#1) -- node[glab]
    {#3\\[-1.4pt]{\fontsize{5.2}{5.2}\selectfont\sffamily
     \textcolor{raMuted}{#4}}} (g#2);%
}
\ovGraphNodes
\ovGraphEdges
\renewcommand{\ovgnode}[2]{\node[ovbadge, fill=rg#1] at (g#1.north west) {#1};}
\ovGraphNodes

\node[band] at (\ZcX,\BandY) {\bench{} \textcolor{raMuted}{\mdseries$\cdot$ the protocol}};
\draw[raAccent!35, line width=0.5pt] (\ZcX,{\BandY-0.07}) -- ({\ZcX+\ZcW},{\BandY-0.07});
\draw[big] ({\ZbX+\ZbW+0.10},\ArrY) -- ({\ZcX-0.10},\ArrY);
\node[note, anchor=south] at ({(\ZbX+\ZbW+\ZcX)/2},{\ArrY+0.07}) {scores};

\foreach \r in {0.25,0.5,0.75,1.0} {
  \draw[raRule, line width=0.3pt]
    \rv{\AngA}{\r} -- \rv{\AngB}{\r} -- \rv{\AngC}{\r} -- \rv{\AngD}{\r}
    -- \rv{\AngE}{\r} -- \rv{\AngF}{\r} -- cycle;
}
\foreach \a in {\AngA,\AngB,\AngC,\AngD,\AngE,\AngF}
  \draw[raRule, line width=0.3pt] (\RCX,\RCY) -- \rv{\a}{1.0};

\filldraw[raAccent, line width=0.9pt, fill=raAccent!25]
  \rv{\AngA}{\ovOursAone} -- \rv{\AngB}{\ovOursAtwo} -- \rv{\AngC}{\ovOursAthree}
  -- \rv{\AngD}{\ovOursAfour} -- \rv{\AngE}{\ovOursAfive} -- \rv{\AngF}{\ovOursAsix}
  -- cycle;
\foreach \a/\v in {\AngA/\ovOursAone, \AngB/\ovOursAtwo, \AngC/\ovOursAthree,
                   \AngD/\ovOursAfour, \AngE/\ovOursAfive, \AngF/\ovOursAsix}
  \fill[raAccent] \rv{\a}{\v} circle (0.045);

\filldraw[raMuted!95, line width=0.8pt, dash pattern=on 1.5pt off 1.1pt,
          fill=raMuted, fill opacity=0.20]
  \rv{\AngA}{\ovBaseAone} -- \rv{\AngB}{\ovBaseAtwo} -- \rv{\AngD}{\ovBaseAfour}
  -- \rv{\AngE}{\ovBaseAfive} -- \rv{\AngF}{\ovBaseAsix} -- cycle;
\foreach \a/\v in {\AngA/\ovBaseAone, \AngB/\ovBaseAtwo, \AngD/\ovBaseAfour,
                   \AngE/\ovBaseAfive, \AngF/\ovBaseAsix}
  \fill[raMuted!85] \rv{\a}{\v} circle (0.038);

\foreach \r/\lbl in {0.25/0.25, 0.5/0.50, 0.75/0.75, 1.0/1.00}
  \node[font=\fontsize{4.8}{4.8}\selectfont\sffamily, text=raMuted,
        anchor=west, inner xsep=1.0pt, inner ysep=0.5pt,
        fill=white, fill opacity=0.72, text opacity=1] at \rv{\AngA}{\r} {\lbl};

\node[note, anchor=south, text=raInk] at \rv{\AngA}{1.14} {A1 transfer};
\node[note, anchor=west, text=raInk] at \rv{\AngB}{1.08} {A2 precision};
\node[note, anchor=west, text=raInk, align=left] at \rv{\AngC}{1.08}
  {A3 open vocab.};
\node[note, anchor=north, text=raInk] at \rv{\AngD}{1.14} {A4 deployment};
\node[note, anchor=east, text=raInk, align=right] at \rv{\AngE}{1.08}
  {A5 graph quality};
\node[note, anchor=east, text=raInk] at \rv{\AngF}{1.08} {A6 spatial};

\node[band, text=raInk, anchor=south west, font=\fnote] at (\ZcX,1.94)
  {COMPOSITE SCORE $\uparrow$};
\draw[raRule, line width=0.4pt] (\ZcX,1.88) -- ({\ZcX+\ZcW},1.88);
\draw[fill=raAccent!8, draw=raAccent!22, rounded corners=3pt, line width=0.4pt]
  (\ZcX,0.66) rectangle ({\ZcX+\ZcW},1.70);
\draw[raAccent!25, line width=0.4pt]
  ({\ZcX+\ZcW/2},0.80) -- ({\ZcX+\ZcW/2},1.56);
\node[font=\fbig, text=raAccent, anchor=base] at ({\ZcX+1.32},1.30)
  {\ovOursOVS};
\node[note, text=raInk, anchor=north, text width=2.44cm] at ({\ZcX+1.32},1.20)
  {\tikz[baseline=-0.55ex]{\filldraw[raAccent, line width=0.9pt, fill=raAccent!25]
     (0,-0.06) rectangle (0.30,0.12);}\;\model{}};
\node[font=\fbig, text=raMuted, anchor=base] at ({\ZcX+3.98},1.30)
  {\ovBaseOVS};
\node[note, text=raInk, anchor=north, text width=2.44cm] at ({\ZcX+3.98},1.20)
  {\tikz[baseline=-0.55ex]{\filldraw[raMuted!95, line width=0.8pt,
     dash pattern=on 1.5pt off 1.1pt, fill=raMuted!20]
     (0,-0.06) rectangle (0.30,0.12);}\;OvSGTR (baseline)};
\end{tikzpicture}

%% file: sec/02_related.tex
\section{Related work}
\label{sec:related}

\paragraph{Scene graphs and their benchmarks}
Visual relationship detection was formulated by \citet{vrd} as the prediction
of $\langle$subject, predicate, object$\rangle$ triplets, with a language
prior over predicates in the method from the outset. Visual Genome \citep{vg}
supplied the annotations and the VG150 split \citep{xu2017} fixed the
vocabulary at 150 object classes and 50 predicates. Later benchmarks changed
the images or the region type more than the formulation: PSG \citep{psg} uses
panoptic segments, GQA \citep{gqa} normalises the vocabulary, Open Images
\citep{openimages} is sparse and large, IndoorVG \citep{indoorvg} targets
indoor robotics, and HICO-DET \citep{hicodet} and SpatialSense
\citep{spatialsense} isolate human--object interactions and adversarially
collected spatial relations \citep{sggsurvey}. Haystack \citep{haystack}
differs in construction: annotated per predicate rather than per image, built
on SA-1B \citep{sam}, and alone among these in publishing explicit negatives,
which is what allows \cref{sec:results:a2} to measure precision against
adjudicated negatives.

\paragraph{Closed-set generation and its evaluation}
The dominant approach predicts relations over a fixed predicate set with a
classifier conditioned on object labels and layout, using message passing
\citep{xu2017}, sequential context \citep{motifs}, tree structure
\citep{vctree}, or knowledge-routed graphs \citep{kern}. \citet{motifs} also
introduced the frequency baseline that we measure in \cref{sec:priors}.
Evaluation moved from recall at $K$ \citep{xu2017} to mean recall
\citep{kern,vctree} once it became clear that the skew of VG150 allows a model
to score well while predicting little beyond \pred{on} and \pred{has}, and
subsequently to debiasing \citep{tde,bgnn,informativesgg}, compositional
augmentation \citep{knyazev} and label correction \citep{noisylabel}. Each of
these estimates its correction from training label statistics, that is, from
a prior of the same kind as the one it corrects. None addresses the two
observations of \cref{sec:priors,sec:protocol}, namely that benchmarks share
their vocabulary with the training corpora and that the standard matcher
\citep{sgbench} grants credit that object detection has disallowed since COCO
\citep{coco}. From LVIS \citep{lvis} we adopt three evaluation practices:
federated annotation, a support-based rare/common/frequent split, and
precision computed over labelled cells only.

\paragraph{One-stage and real-time generation}
SGTR \citep{sgtr,sgtrplus}, RelTR \citep{reltr}, EGTR \citep{egtr} and PE-Net
\citep{penet} predict subject, object and predicate from a single set of
queries, and one-stage human--object interaction detectors \citep{ppdm,qpic}
apply the same design at real-time rates. REACT \citep{react} and REACT++
\citep{reactpp} are, to our knowledge, the fastest published scene-graph
models; both are closed-set and trained per benchmark, and
\cref{sec:efficiency} compares against their published numbers. All share a
backbone between objects and relations and condition on predicted object
classes.

\paragraph{Open-vocabulary relations}
Replacing the predicate classifier with similarity against text embeddings
admits new predicates at inference time. Approaches include prompting a
vision--language model \citep{ovsggprompt}, building on the visual--semantic
space of a grounded detector \citep{vs3}, composing CLIP \citep{clip} prompts
without relation training \citep{recode}, expanding predicates with a language
model \citep{epic,sdsgg}, generating the graph as a sequence \citep{pgsg}, and
refining the prompting and the pairing \citep{rahp,inova}. OvSGTR
\citep{ovsgtr_eccv} formulated fully open-vocabulary generation over both
objects and predicates and defined the OvR-SGG protocol that we reproduce in
\cref{sec:protocol}. Its extension \citep{ovsgtr} pre-trains on MegaSG, a
machine-annotated corpus, which makes it the closest available control for
supervision quality and our baseline throughout. Scene-Graph ViT \citep{sgvit}
and relational pre-training \citep{rlip,rlipv2} learn relations and objects
jointly from region--text pairs. Our setting differs in three respects: no
object class labels are used at any stage, supervision is free text over a
five-figure vocabulary rather than a projection onto a benchmark's set, and
evaluation is conducted both with the answer set supplied and with it withheld
(\cref{sec:results:a3}). Detectors that accept a class list at inference
\citep{vild,detic,owlvit,glip,groundingdino}, including real-time ones
\citep{yoloworld,yoloe}, and segmenters that return class-agnostic regions
\citep{sam,fastsam} are the region sources available to a relation model. A
relation model that requires object labels can use none of them without a
label space it was trained on, whereas one that reads regions can use all of
them (\cref{sec:anyinputs:regions}).

\paragraph{Machine-generated relation supervision}
Existing approaches differ in how a generated relation is attached to an
instance and in whether the annotator is checked (\cref{tab:gen} in
\cref{app:corpus}). Caption parsing \citep{gpt4sgg} yields ungrounded
triplets, RLIPv2 \citep{rlipv2} grounds parsed text with a learned tagger, and
GPT4SGG \citep{gpt4sgg}, MegaSG \citep{ovsgtr}, Synthetic Visual Genome
\citep{svg} and the All-Seeing Project \citep{asv2} prompt a multimodal model
directly, accepting its output or filtering it with a second model. None
verifies a generated relation against the geometry of the boxes it names;
\dataset{} does (\cref{sec:corpus}), and retains every surface form rather
than projecting onto a class list as GQA and MegaSG do.

\paragraph{Learning at vocabulary scale}
Positive-unlabeled learning \citep{elkan,nnpu,pusurvey} treats an unlabeled
example as a mixture of positive and negative; federated annotation
\citep{lvis} is the same observation applied to a benchmark. We encounter it
in the training signal (\cref{sec:scale}): a pair annotated \pred{riding} is
also, implicitly, \pred{sitting on}. Contrastive text spaces are known to be
anisotropic \citep{abtt} and to exhibit hubness \citep{hubness,csls};
relations expose an additional failure that these works do not address,
namely that antonyms embed almost identically.

%% file: sec/03_model.tex
\section{The \model{} model}
\label{sec:model}

\model{} takes an image and a set of regions and returns scored relations
between pairs of regions, drawn from a predicate vocabulary supplied at
inference. We state the problem (\cref{sec:decouple:contract}), describe the
architecture (\cref{sec:model:arch}) and discuss the two properties of the
learning problem that change at a vocabulary of ten thousand predicates
(\cref{sec:scale}); the evidence that coupling object and relation prediction
does not help is separate (\cref{sec:decouple}).

\subsection{Problem statement}
\label{sec:decouple:contract}

Given an image $I$ and a set of boxes $B = \{b_i\}_{i=1}^{N}$, the model
outputs scored triplets $(i, j, p)$, where $p$ is drawn from a vocabulary
$\mathcal{V}$ supplied at inference as strings. Two constraints follow.

\paragraph{No object labels}
The model receives only pixels and box coordinates. This setting is harder
than the predicate-classification protocol under which most published numbers
are obtained, in which ground-truth classes are supplied and carry the prior
analysed in \cref{sec:priors:freq}, and it corresponds to deployment, where
labels come from a detector and may be noisy, out of vocabulary, or absent.
Object categories are used during training only inside two loss terms, the
pair sampler's relatedness loss and the object--text grounding term
(\cref{sec:scale:pu}, \cref{app:aux}), and never as a network input.

\paragraph{Vocabulary as input}
There is no predicate classifier. Scores are cosine similarities between
visual pair embeddings and text embeddings, and no learned transformation is
applied on the text side, since a projection fitted to the training
vocabulary would be undefined for strings supplied later. Replacing the
vocabulary therefore amounts to substituting one matrix and requires no
retraining.

\subsection{Architecture}
\label{sec:model:arch}

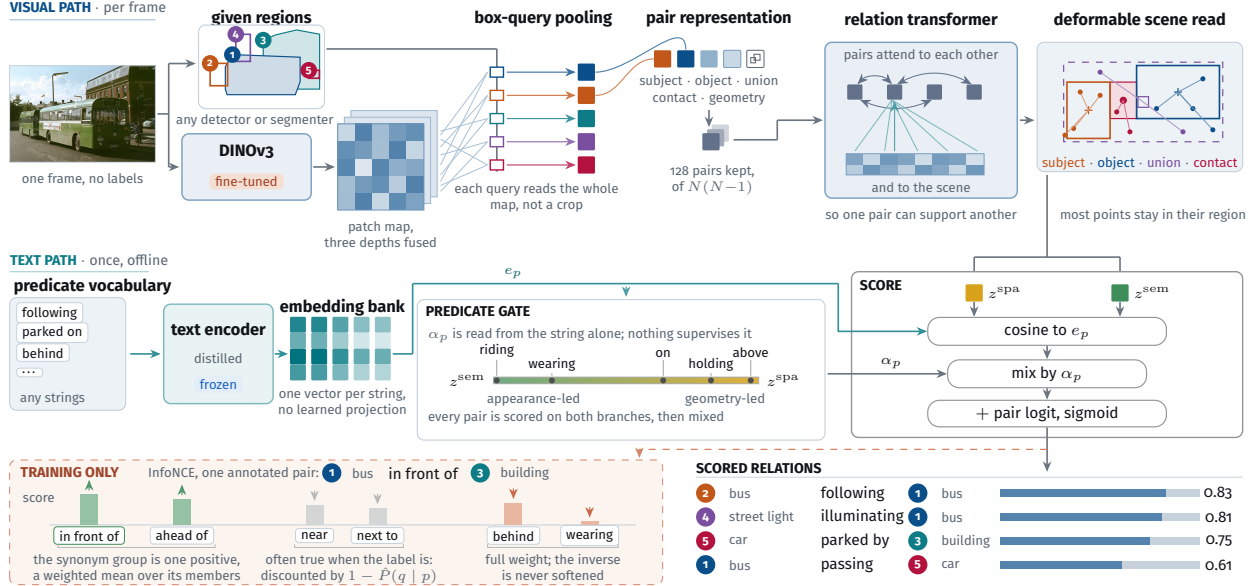
\begin{figure*}[t]
\centering
\resizebox{\textwidth}{!}{\input{figures/architecture.tex}}
\caption{Overview of \model{}. \textbf{Band 1, the visual path.} The frame goes to the
backbone and, unchanged, to whatever produces the regions; that producer is
not part of the model and its class labels are discarded. One query per box
pools the patch map, five pooled features form each ordered pair, and the
sampled pairs are refined against one another and against the scene.
\textbf{Band 2, the text path and the score.} Predicate strings supplied at
inference are embedded once by a frozen encoder into a bank of unit vectors
that no learned layer transforms; a gate maps each embedding to the weight
$\alpha_p$ mixing the two branches (\cref{app:arch:pairhead}). \textbf{Band 3,
the loss}, training only: one annotated pair scored against six predicates in
its batch, the annotated string and its synonyms forming one positive, a
negative the corpus often co-annotates with the label discounted, the
directional inverse never. Frame, regions and relations are those of
\cref{fig:overview}.}
\label{fig:arch}
\end{figure*}

The parameters of the model are concentrated in the pair representation
(\cref{fig:arch}); \cref{app:arch} gives the full specification and the
ablation behind each component.

\paragraph{Visual path}
A DINOv3 backbone \citep{dinov3} encodes the image at $448\times448$. Dense
patch features are read at depths $-6$, $-3$ and $-1$, layer-normalised per
tap and fused with learned weights. Per-object features are obtained by
coordinate-aware soft pooling rather than by cropping: a query formed from a
Fourier position encoding \citep{fourierfeatures} of box $b_i$ cross-attends
over all patch tokens, $v_i = \mathrm{Attn}(q(b_i), F, F)$, so that an
object's representation can incorporate the context it interacts with. The
backbone is fully fine-tuned at a learning rate $8\times$ lower than that of
the head.

\paragraph{Pair representation}
For an ordered pair we form
\begin{equation}
x_{ij} = \big[\,v_i;\; v_j;\; v_{\cup(i,j)};\; v_{\cap(i,j)};\;
\mathrm{MLP}(g_{ij})\,\big]\,W_{\mathrm{proj}},
\qquad x_{ij} \in \mathbb{R}^{512},
\end{equation}
where $v_\cup$ pools the union box, $v_\cap$ pools the \emph{contact} region
(the intersection when the boxes overlap and the gap between them otherwise),
and $g_{ij} \in \mathbb{R}^{19}$ collects scale-invariant geometric
features.\footnote{Fifteen box features and four mask-only slots, constant for boxes and encoding mask shape for segments.} A small transformer then refines the sampled pairs with
self-attention over pairs and cross-attention to the scene tokens together
with tokens encoding the pair's own box corners, followed by a block of joint
self-attention over pairs and scene tokens. A final additive
stage allows each pair to read the scene outside its own boxes: it predicts
sampling offsets around its four anchors and reads the feature map at those
locations through a zero-initialised gate \citep{deformabledetr,dabdetr}. This
stage is what allows \pred{parked on} to be decided by the road surface and
\pred{hanging from} by the attachment point above the subject
(\cref{app:arch:deform}).

\paragraph{Two branches and a predicate-conditioned gate}
Spatial and semantic relations rely on different evidence, so the head scores
each pair twice and mixes the two cosine channels per predicate:
\begin{equation}
\ell_{\mathrm{pred}}(i,j,p) = \tau\,\big[\alpha_p \cos(z^{\mathrm{spa}}_{ij}, e_p)
  + (1-\alpha_p) \cos(z^{\mathrm{sem}}_{ij}, e_p)\big] + \beta,
\label{eq:gate}
\end{equation}
where $e_p$ is the text embedding of predicate $p$ and $\alpha_p = g(e_p)$
depends on that embedding alone, so that it is defined for strings the model
has never seen. The gate receives no supervision of its own. The weights it
learns are strongly bimodal: projective and proximity relations are routed to
the geometry branch and actions to the appearance branch (\cref{fig:gate} in
\cref{app:arch:pairhead}). The two channels can also be read out separately at
inference, as a layout graph and a content graph (\cref{tab:decomposed}).

\paragraph{Pair sampling}
A two-stage sampler retains 400 of the $N(N-1)$ ordered pairs by geometric
plausibility and then 128 by a learned relatedness score. It retains 99.79\%
of annotated positives; exhaustive scoring costs $1.02\times$ and yields no
measurable gain. The learned stage is trained to predict which pairs carry
annotations, and hence learns the annotation propensity discussed in
\cref{sec:priors:propensity}. Its logit enters the final score additively
(\cref{app:score}), which allows it to be removed at evaluation time and its
contribution measured (\cref{sec:results:a2}).

\subsection{Training with $10^4$ predicates}
\label{sec:scale}

Open-vocabulary relation models are usually trained on the same few dozen
predicates as closed-set models, with only the head replaced. We instead train
against a bank of 19{,}103 free-text predicate strings: \dataset{}'s own
10{,}102, together with 9{,}001 strings taken from the vocabularies of three
further relation corpora \citep{svg,gqa,spatialsense} whose relations we do not
train on. Since nothing on the text side is learned, the bank is the answer
space rather than the supervision, and making it wider than any single source
costs nothing (\cref{app:corpus:decisions}). Two properties of the learning
problem change at this scale, and both must be addressed for the larger
vocabulary to be beneficial. \Cref{tab:notation} in
\cref{app:objective} summarises the notation.

The primary objective is a batch-local InfoNCE loss \citep{infonce} over
predicate slots. For each annotated pair, the positive is the annotated
predicate's \emph{synonym group}, each member weighted by an estimated
probability of synonymy and the group aggregated as a weighted mean, and the
negatives are the other predicates in the batch. Aligning to a group rather
than to a single string is a trade-off that we quantify in
\cref{app:textspace}: group positives transfer to unseen vocabularies, whereas
exact-string positives improve within-group ranking at a cost on rare
predicates. Four auxiliary terms, namely a per-cell sigmoid loss, a swap hinge
on directional predicates, background suppression and an object--text
grounding term on the pooled features, are specified in \cref{app:aux}, and
the full recipe is given in \cref{tab:recipe}.

\paragraph{Positive-unlabeled supervision}
\label{sec:scale:pu}
With 50 predicates, treating every non-annotated predicate as a negative is
approximately correct. With ten thousand it is systematically wrong: if a
$\langle$man, horse$\rangle$ pair is annotated \pred{riding}, then
\pred{sitting on} and \pred{mounted on} are true but unstated, and penalising
them teaches the model to suppress correct answers. This is the
positive-unlabeled structure of federated annotation \citep{lvis}, encountered
at training time. We address it by estimating, for every ordered pair of
predicates, the probability that $q$ also holds when $p$ is annotated, and by
discounting each negative by that probability. The estimate is fitted on the
706k box pairs of the training mixture that carry more than one annotation,
from co-annotation counts, text-space cosine and frequency, with the
correction of \citet{elkan} for the fact that an unstated predicate is
unlabeled rather than false. The denominator logit of negative $q$ for a pair annotated $p$ becomes
\begin{equation}
\ell_{\mathrm{pred}}(i,j,q) + \log\big(1 - \hat{P}(q \mid p)\big),
\label{eq:softneg}
\end{equation}
so that a predicate that is almost surely also true contributes almost
nothing to the denominator; a pair carrying several annotations combines
their discounts as if independent. Two exemptions are essential: annotated predicates are never
down-weighted, and directional inverses always receive full weight, since the
supervision that separates \pred{above} from \pred{below} is the one signal
that must not be softened. The positive-side synonym weights are an
isotonic fit of the probability of synonymy against text-space cosine on
1{,}177 lexical synonym pairs, and the swap hinge of \cref{app:aux} is
weighted by the estimated probability that a predicate is directional, so no
semantic constant of the objective is set by hand. Object categories enter
training only through the sampler's relatedness loss, which weights an
unannotated pair by one minus the annotation rate of its category pair (floor
0.3), and through the grounding term of \cref{app:aux}; both condition a loss
rather than the forward pass and cannot supply a predicate prior to the
network. A source-aware refinement is described in \cref{app:aux}.

\begin{figure}[t]
\centering
\includegraphics[width=\linewidth]{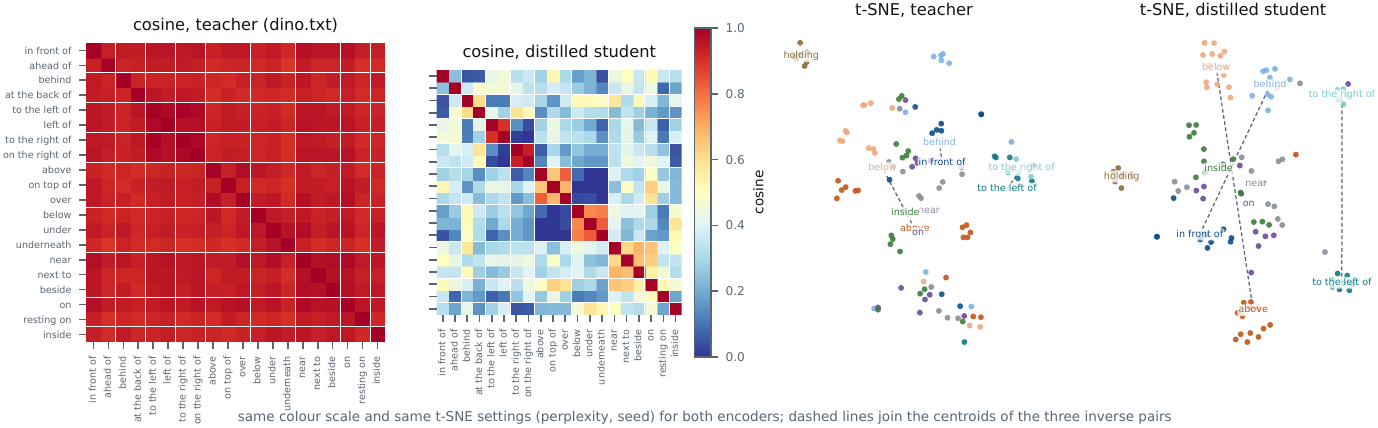}
\caption{\textbf{Direction in the text embedding space.} Left two panels: pairwise
cosines of twenty spatial predicates under the teacher and under the distilled
student, same colour scale. Right two: ten spatial synonym families (107
surface forms) under t-SNE \citep{tsne}, same settings for both. The teacher's
matrix is a near-uniform block above 0.8 in which \pred{above}/\pred{below} is
as close as \pred{above}/\pred{on top of}, and its t-SNE panel interleaves the
members of each inverse pair. The student retains the synonym blocks, drives
the inverse blocks to near zero and splits each inverse pair into two
clusters. Statistics in \cref{tab:textspace} (\cref{app:textspace}).}
\label{fig:textcos}
\end{figure}

\paragraph{Direction in the text embedding space}
\label{sec:scale:text}
Because the head regresses visual features onto text directions, the geometry
of the text space bounds what the model can express. Contrastive text encoders
fail in precisely the dimension on which relations depend: antonyms occur in
near-identical contexts and are embedded almost identically. On our vocabulary
the teacher encoder (dino.txt \citep{dinotxt}, 2048-d) places \pred{above} and
\pred{below} at a cosine similarity of 0.95, \pred{in front of} and
\pred{behind} at 0.94, and \pred{to the left of} and \pred{to the right of} at
0.99, which is indistinguishable from its synonym pairs at 0.96 on average
(\cref{fig:textcos}; \cref{tab:textspace} in \cref{app:textspace}). No visual
encoder trained against such targets can reliably separate \pred{above} from
\pred{below}. This limitation lies in the objective rather than in the visual
model, and it applies to every method that scores relations against
off-the-shelf text embeddings. Removing the mean direction \citep{abtt} does
not resolve it.

We therefore distil a 512-dimensional student encoder over the full CLIP BPE
vocabulary \citep{clip}, so that no predicate string can fall out of
vocabulary. Of its five objectives (\cref{app:textspace}), the
antonym-repulsion hinge on known inverse pairs is the one a frozen teacher
cannot supply, and a neighbourhood-preservation term is what stops the student
from buying that separation cheaply: without the latter, a student reaches an
equally good synonym-versus-inverse AUC at an effective dimension of 24 rather
than 44. After distillation that AUC rises from 0.85 to 0.99, the mean inverse cosine
falls from 0.92 to 0.09 while synonyms stay at 0.71, and hubness falls from
8.1 to 1.3. The student is used offline only; at inference the model carries the embedding bank.

One consequence of this objective affects every closed-vocabulary result
below. Because the model is aligned to a synonym group, it frequently answers
with a member of the correct group other than the annotated one, for instance
\pred{wearing} where VG150 reads \pred{wears}, which exact-string metrics
count as a miss. Such numbers are therefore conservative for our model, and A3
(\cref{sec:results:a3}) provides the corrected view.

%% file: figures/architecture.tex
\definecolor{rg0}{HTML}{0B4F8A}%
\definecolor{rg1}{HTML}{C2571A}%
\definecolor{rg2}{HTML}{0E7C86}%
\definecolor{rg3}{HTML}{7A4E9E}%
\definecolor{rg4}{HTML}{B3123F}%
\definecolor{raPair}{HTML}{62708A}
\definecolor{raSpa}{HTML}{D9A21B}
\definecolor{raSem}{HTML}{3E8E5A}
\def\ftitle{\fontsize{7.5}{8.5}\selectfont\sffamily\bfseries}%
\def\fnote{\fontsize{6}{7}\selectfont\sffamily}%
\def\fsmall{\fontsize{7}{8}\selectfont\sffamily}%
\def\fband{\fontsize{6.5}{7.5}\selectfont\sffamily\bfseries}%
\def\IW{2.20}\def\IH{1.457}\def\IX{0.0}\def\IY{6.49}%
\def\BW{1.76}\def\BH{1.12}\def\BX{2.92}\def\BY{7.37}%
\newcommand{\apx}[2]{({\IX+#1*\IW},{\IY+#2*\IH})}
\newcommand{\bpx}[2]{({\BX+#1*\BW},{\BY+#2*\BH})}

\begin{tikzpicture}[x=1cm, y=1cm,
  font=\fnote,
  >={Stealth[length=1.6mm, width=1.3mm]},
  flow/.style={->, line width=0.7pt, draw=raMuted!90},
  wire/.style={line width=0.7pt, draw=raMuted!90},
  tflow/.style={->, line width=0.7pt, draw=raRef!85},
  twire/.style={line width=0.7pt, draw=raRef!85},
  attn/.style={line width=0.25pt, draw=raAccent!35},
  mod/.style={draw=raAccent!55, fill=raAccent!9, rounded corners=3pt,
              line width=0.6pt, align=center, inner sep=3pt},
  tmod/.style={mod, draw=raRef!55, fill=raRef!9},
  panel/.style={draw=raRule, fill=raPanel, rounded corners=3pt, line width=0.5pt},
  wpanel/.style={draw=raRule, fill=white, rounded corners=3pt, line width=0.5pt},
  opanel/.style={draw=raInk!45, fill=white, rounded corners=3pt, line width=0.6pt},
  title/.style={font=\ftitle, text=raInk, align=center, inner sep=1pt},
  note/.style={font=\fnote, text=raMuted, align=center, inner sep=1pt},
  lnote/.style={note, align=left},
  rnote/.style={note, align=right},
  tok/.style={inner sep=0pt, minimum size=2.6mm, rounded corners=0.6pt},
  band/.style={font=\fband, text=raAccent, anchor=south west, inner sep=0pt},
  tag/.style={font=\fnote, rounded corners=2pt, inner xsep=2.5pt, inner ysep=1.3pt},
  hot/.style={tag, fill=BrickRed!12, text=BrickRed!85!black},
  cold/.style={tag, fill=raLink!12, text=raLink!90!black},
  chip/.style={draw=raRule, fill=white, rounded corners=1.5pt, anchor=west,
               inner xsep=2.4pt, inner ysep=1.5pt, font=\fnote, text=raInk},
  badge/.style={circle, inner sep=0pt, minimum size=2.9mm,
                font=\fontsize{5.5}{5.5}\selectfont\sffamily\bfseries, text=white},
  sbadge/.style={circle, inner sep=0pt, minimum size=2.5mm,
                 font=\fontsize{5}{5}\selectfont\sffamily\bfseries, text=white},
  pill/.style={draw=raInk!45, fill=white, rounded corners=5pt, line width=0.5pt,
               inner xsep=4pt, inner ysep=2.4pt, font=\fsmall, text=raInk, align=center},
]

\node[band] at (0,8.70) {VISUAL PATH \textcolor{raMuted}{\mdseries$\cdot$ per frame}};
\draw[raAccent!35, line width=0.5pt] (0,8.64) -- (2.20,8.64);

\node[anchor=south west, inner sep=0] at (\IX,\IY)
  {\includegraphics[width=\IW cm]{figures/teaser_frame.jpg}};
\draw[raRule, line width=0.4pt] (\IX,\IY) rectangle ({\IX+\IW},{\IY+\IH});
\node[note, anchor=north] at (1.10,6.37) {one frame, no labels};

\draw[wire] (2.23,7.22) -- (2.40,7.22);
\draw[wire] (2.40,6.42) -- (2.40,7.98);
\draw[flow] (2.40,7.98) -- (2.81,7.98);
\draw[flow] (2.40,6.42) -- (2.56,6.42);

\node[title] at (3.80,8.66) {given regions};
\node[wpanel, minimum width=1.90cm, minimum height=1.26cm] at (3.80,7.93) {};
\filldraw[fill=rg2!20, draw=rg2, line width=0.5pt]
              \bpx{0.515}{0.200} -- \bpx{0.998}{0.200} -- \bpx{0.998}{0.940}
           -- \bpx{0.800}{0.997} -- \bpx{0.620}{0.930} -- \bpx{0.515}{0.860} -- cycle;
\filldraw[fill=rg0!20, draw=rg0, line width=0.5pt]
              \bpx{0.205}{0.125} -- \bpx{0.790}{0.125} -- \bpx{0.831}{0.300}
           -- \bpx{0.831}{0.630} -- \bpx{0.250}{0.665} -- \bpx{0.190}{0.500}
           -- \bpx{0.190}{0.200} -- cycle;
\draw[rg1, line width=0.7pt] \bpx{0.0598}{0.2455} rectangle \bpx{0.2096}{0.5471};
\draw[rg3, line width=0.7pt] \bpx{0.2798}{0.5733} rectangle \bpx{0.4043}{0.9383};
\draw[rg4, line width=0.7pt] \bpx{0.9049}{0.3462} rectangle \bpx{0.9959}{0.4543};
\node[sbadge, fill=rg0] at \bpx{0.250}{0.665} {1};
\node[sbadge, fill=rg1] at \bpx{0.0598}{0.5471} {2};
\node[sbadge, fill=rg2] at \bpx{0.515}{0.860} {3};
\node[sbadge, fill=rg3] at \bpx{0.2798}{0.9383} {4};
\node[sbadge, fill=rg4] at \bpx{0.9049}{0.4543} {5};
\node[note, anchor=north] at (3.70,7.26) {any detector or segmenter};

\node[mod, minimum width=1.95cm, minimum height=1.00cm] (bb) at (3.575,6.42) {};
\node[title] at (3.575,6.66) {DINOv3};
\node[hot] at (3.575,6.20) {fine-tuned};
\draw[flow] (4.58,6.42) -- (4.91,6.42);

\foreach \o in {0.20,0.10}
  \fill[raAccent!14, draw=raAccent!40, line width=0.3pt]
    ({4.95+\o},{5.77+\o}) rectangle ({6.25+\o},{7.07+\o});
\foreach \i in {0,...,4} \foreach \j in {0,...,4} {
  \pgfmathtruncatemacro{\t}{14 + 58*abs(sin(73*\i + 131*\j + 25))}
  \fill[raAccent!\t] ({4.95+0.26*\i},{5.77+0.26*\j}) rectangle ++(0.26,0.26);
}
\draw[raAccent!60, line width=0.35pt] (4.95,5.77) rectangle (6.25,7.07);
\node[note, anchor=north] at (5.60,5.65) {patch map,\\ three depths fused};

\node[title] at (8.05,8.66) {box-query pooling};
\draw[wire] (4.78,8.20) -- (7.35,8.20);
\draw[wire] (7.35,8.20) -- (7.35,6.35);
\foreach \k/\c/\y/\ya/\yb in {1/rg0/7.85/6.90/6.45, 2/rg1/7.50/6.72/6.20,
                              3/rg2/7.15/6.96/6.30, 4/rg3/6.80/6.60/6.05,
                              5/rg4/6.45/6.84/6.14} {
  \draw[attn] (6.50,\ya) -- (7.21,\y);
  \draw[attn] (6.50,\yb) -- (7.21,\y);
  \draw[\c, fill=white, line width=0.6pt] (7.25,{\y-0.075}) rectangle (7.45,{\y+0.075});
  \draw[->, \c!80, line width=0.5pt] (7.50,\y) -- (8.54,\y);
  \node[tok, fill=\c] (obj\k) at (8.70,\y) {};
}
\node[note, anchor=north] at (7.95,6.22) {each query reads the whole\\ map, not a crop};

\node[title] at (10.70,8.66) {pair representation};
\node[tok, fill=rg1] at (9.85,8.05) {};
\node[tok, fill=rg0] at (10.20,8.05) {};
\node[tok, fill=raAccent!45] at (10.55,8.05) {};
\node[tok, fill=raAccent!22, draw=raAccent!55, line width=0.35pt] at (10.90,8.05) {};
\node[tok, fill=white, draw=raMuted, line width=0.35pt] at (11.25,8.05) {};
\draw[raMuted, line width=0.3pt] (11.18,7.99) rectangle (11.25,8.07);
\draw[raMuted, line width=0.3pt] (11.23,8.03) rectangle (11.32,8.12);
\draw[rg1!70, line width=0.5pt] (obj2.east) to[out=0, in=180] (9.72,8.05);
\draw[rg0!70, line width=0.5pt] (obj1.east) to[out=0, in=90] (10.20,8.19);
\node[note, anchor=north] at (10.55,7.85)
  {subject $\cdot$ object $\cdot$ union\\ contact $\cdot$ geometry};
\draw[flow] (10.55,7.28) -- (10.55,7.02);
\fill[raPair!30, rounded corners=0.6pt] (10.59,6.80) rectangle (10.85,7.06);
\fill[raPair!60, rounded corners=0.6pt] (10.52,6.73) rectangle (10.78,6.99);
\node[tok, fill=raPair] at (10.58,6.80) {};
\node[note, anchor=north] at (10.60,6.50) {128 pairs kept,\\ of $N(N{-}1)$};

\draw[wire] (10.92,6.83) -- (11.58,6.83);
\draw[wire] (11.58,6.83) -- (11.58,7.125);
\draw[flow] (11.58,7.125) -- (12.27,7.125);
\node[title] at (13.75,8.66) {relation transformer};
\node[mod, minimum width=2.90cm, minimum height=2.35cm] at (13.75,7.125) {};
\foreach \x in {12.75,13.35,13.95,14.55} \node[tok, fill=raPair, minimum size=2.3mm] at (\x,7.55) {};
\draw[<->, line width=0.4pt, draw=raPair] (12.79,7.67) to[bend left=45] (13.31,7.67);
\draw[<->, line width=0.4pt, draw=raPair] (13.39,7.67) to[bend left=28] (14.51,7.67);
\draw[<->, line width=0.4pt, draw=raPair] (12.83,7.43) to[bend right=22] (13.91,7.43);
\node[note, anchor=south] at (13.75,7.96) {pairs attend to each other};
\foreach \i in {0,...,7} \foreach \j in {0,1} {
  \pgfmathtruncatemacro{\t}{16 + 46*abs(sin(59*\i + 113*\j + 17))}
  \fill[raAccent!\t] ({12.62+0.26*\i},{6.30+0.16*\j}) rectangle ++(0.26,0.16);
}
\draw[raAccent!55, line width=0.3pt] (12.62,6.30) rectangle (14.70,6.62);
\foreach \sx in {12.75,13.20,13.70,14.15,14.60}
  \draw[attn, draw=raRef!55] (\sx,6.66) -- (13.31,7.43);
\draw[->, line width=0.45pt, draw=raRef!75] (13.35,6.66) -- (13.35,7.41);
\node[note, anchor=north] at (13.75,6.24) {and to the scene};
\node[note, anchor=north] at (13.75,5.84) {so one pair can support another};

\draw[flow] (15.24,7.125) -- (15.46,7.125);
\node[title] at (17.05,8.66) {deformable scene read};
\node[panel, minimum width=3.10cm, minimum height=1.95cm] at (17.05,7.325) {};
\draw[rg3, line width=0.5pt, dashed] (15.90,6.80) rectangle (18.30,8.00);
\fill[rg4!14] (16.60,7.15) rectangle (17.00,7.70);
\draw[rg4, line width=0.5pt] (16.60,7.15) rectangle (17.00,7.70);
\draw[rg1, line width=0.7pt] (15.95,6.85) rectangle (16.60,7.70);
\draw[rg0, line width=0.7pt] (17.00,7.15) rectangle (18.25,7.95);
\draw[rg1, line width=0.5pt] (16.215,7.275) -- (16.335,7.275) (16.275,7.215) -- (16.275,7.335);
\draw[rg0, line width=0.5pt] (17.565,7.55) -- (17.685,7.55) (17.625,7.49) -- (17.625,7.61);
\draw[rg3, line width=0.5pt] (17.02,7.32) rectangle (17.18,7.48);
\fill[rg4] (16.80,7.425) circle (0.055);
\foreach \ax/\ay/\dx/\dy/\c in {16.275/7.275/-0.225/-0.275/rg1, 16.275/7.275/0.175/0.225/rg1,
                                16.275/7.275/-0.125/0.275/rg1, 16.275/7.275/-0.355/-0.375/rg1,
                                17.625/7.55/-0.325/-0.250/rg0, 17.625/7.55/0.375/0.200/rg0,
                                17.625/7.55/0.225/-0.250/rg0, 17.625/7.55/0.525/-0.570/rg0,
                                17.10/7.40/-0.650/0.450/rg3, 17.10/7.40/0.600/-0.450/rg3,
                                16.80/7.425/-0.100/-0.175/rg4, 16.80/7.425/0.100/-0.475/rg4} {
  \draw[\c!60, line width=0.3pt] (\ax,\ay) -- (\ax+\dx,\ay+\dy);
  \fill[\c] (\ax+\dx,\ay+\dy) circle (0.038);
}
\node[note, anchor=north] at (17.05,6.62) {\textcolor{rg1}{subject}
  $\cdot$ \textcolor{rg0}{object} $\cdot$ \textcolor{rg3}{union}
  $\cdot$ \textcolor{rg4}{contact}};
\node[note, anchor=north] at (17.25,5.84) {most points stay in their region};

\draw[wire] (15.65,6.30) -- (15.65,5.06);
\draw[wire] (14.55,5.06) -- (16.75,5.06);
\draw[flow] (14.55,5.06) -- (14.55,4.68);
\draw[flow] (16.75,5.06) -- (16.75,4.68);

\node[band, text=raRef] at (0,4.90) {TEXT PATH \textcolor{raMuted}{\mdseries$\cdot$ once, offline}};
\draw[raRef!35, line width=0.5pt] (0,4.84) -- (2.10,4.84);

\node[title, anchor=west] at (0.02,4.62) {predicate vocabulary};
\draw[panel] (0,2.75) rectangle (1.75,4.45);
\node[chip] at (0.10,4.22) {following};
\node[chip] at (0.10,3.92) {parked on};
\node[chip] at (0.10,3.62) {behind};
\node[chip] at (0.10,3.32) {\dots};
\node[lnote, anchor=north west] at (0.12,3.08) {any strings};

\draw[tflow] (1.82,3.60) -- (2.28,3.60);
\node[tmod, minimum width=1.65cm, minimum height=1.50cm] at (3.15,3.60) {};
\node[title] at (3.15,3.98) {text encoder};
\node[note] at (3.15,3.54) {distilled};
\node[cold] at (3.15,3.14) {frozen};

\draw[tflow] (4.00,3.60) -- (4.20,3.60);
\node[title] at (5.02,4.30) {embedding bank};
\foreach \x in {4.35,4.67,4.99,5.31,5.63} \foreach \y/\t in {4.04/70, 3.80/40, 3.56/85, 3.32/55} {
  \pgfmathtruncatemacro{\u}{\t + 25*sin(180*\x)}
  \node[tok, fill=raRef!\u, minimum size=2.4mm] at (\x,\y) {};
}
\node[note, anchor=north] at (5.02,3.14) {one vector per string,\\ no learned projection};

\draw[twire] (5.78,3.60) -- (6.08,3.60);
\draw[twire] (6.08,3.60) -- (6.08,4.66);
\draw[twire] (6.08,4.66) -- (12.50,4.66);
\node[note, anchor=south, text=raRef!80!black] at (7.60,4.70) {$e_p$};
\draw[->, dashed, line width=0.5pt, draw=raRef!75] (9.30,4.62) -- (9.30,4.46);

\draw[wpanel] (6.15,2.30) rectangle (12.30,4.42);
\node[band, text=raInk] at (6.27,4.16) {PREDICATE GATE};
\node[lnote, anchor=north west] at (6.27,4.04)
  {$\alpha_p$ is read from the string alone; nothing supervises it};
\shade[left color=raSem!75, right color=raSpa!85] (7.30,3.15) rectangle (11.30,3.25);
\draw[raInk!35, line width=0.3pt] (7.30,3.15) rectangle (11.30,3.25);
\foreach \a/\p/\r in {0.012/riding/1, 0.22/wearing/0, 0.64/on/1, 0.82/holding/0, 0.970/above/1} {
  \pgfmathsetmacro{\px}{7.30 + \a*4.0}
  \fill[raInk!75] (\px,3.20) circle (0.045);
  \ifnum\r=1
    \draw[raInk!40, line width=0.3pt] (\px,3.27) -- (\px,3.51);
    \node[note, text=raInk, anchor=south] at (\px,3.51) {\p};
  \else
    \draw[raInk!40, line width=0.3pt] (\px,3.27) -- (\px,3.31);
    \node[note, text=raInk, anchor=south] at (\px,3.31) {\p};
  \fi
}
\node[note, anchor=east, text=raInk] at (7.22,3.20) {$z^{\mathrm{sem}}$};
\node[note, anchor=west, text=raInk] at (11.38,3.20) {$z^{\mathrm{spa}}$};
\node[lnote, anchor=north west] at (7.16,3.05) {appearance-led};
\node[rnote, anchor=north east] at (11.44,3.05) {geometry-led};
\node[lnote, anchor=north west] at (6.27,2.78)
  {every pair is scored on both branches, then mixed};

\draw[opanel] (12.70,2.34) rectangle (18.60,4.84);
\node[band, text=raInk] at (12.82,4.56) {SCORE};
\node[tok, fill=raSpa] (zspa) at (14.55,4.52) {};
\node[tok, fill=raSem] (zsem) at (16.75,4.52) {};
\node[note, anchor=west, text=raInk] at (14.72,4.52) {$z^{\mathrm{spa}}$};
\node[note, anchor=west, text=raInk] at (16.92,4.52) {$z^{\mathrm{sem}}$};
\node[pill, minimum width=3.60cm] (cos) at (15.65,3.94) {cosine to $e_p$};
\node[pill, minimum width=3.00cm] (mix) at (15.65,3.30) {mix by $\alpha_p$};
\node[pill, minimum width=3.60cm] (sig) at (15.65,2.70) {$+$ pair logit, sigmoid};
\draw[flow] (zspa.south) -- (zspa.south |- cos.north);
\draw[flow] (zsem.south) -- (zsem.south |- cos.north);
\draw[flow] (cos.south) -- (mix.north);
\draw[flow] (mix.south) -- (sig.north);
\draw[tflow] (12.50,4.66) -- (12.50,3.94) -- (cos.west);
\draw[flow] (12.34,3.30) -- (mix.west);
\node[note, anchor=south, text=raInk] at (13.30,3.34) {$\alpha_p$};
\draw[flow] (sig.south) -- (15.65,1.82);
\draw[->, dashed, line width=0.55pt, draw=BrickRed!55] (15.65,2.16) -- (9.50,2.16) -- (9.50,2.03);

\draw[draw=BrickRed!45, fill=BrickRed!4, dashed, rounded corners=3pt, line width=0.5pt]
  (0,0.12) rectangle (9.85,2.00);
\node[band, text=BrickRed!80!black] at (0.16,1.74) {TRAINING ONLY};
\node[lnote, anchor=west] at (2.05,1.805) {InfoNCE, one annotated pair:};
\node[badge, fill=rg0] at (4.86,1.805) {1};
\node[anchor=west, font=\fnote, text=raMuted] at (5.03,1.805) {bus};
\node[anchor=west, font=\fsmall, text=raInk] at (5.58,1.805) {in front of};
\node[badge, fill=rg2] at (7.10,1.805) {3};
\node[anchor=west, font=\fnote, text=raMuted] at (7.27,1.805) {building};
\node[note, anchor=west] at (0.16,1.42) {score};
\draw[raRule, line width=0.4pt] (0.60,1.02) -- (9.45,1.02);
\foreach \x/\h/\p/\d in {1.20/0.46/{in front of}/raSem, 2.60/0.39/{ahead of}/raRule} {
  \fill[raSem!55] ({\x-0.13},1.02) rectangle ({\x+0.13},{1.02+\h});
  \draw[->, line width=0.5pt, draw=raSem!90] (\x,{1.08+\h}) -- (\x,{1.24+\h});
  \node[chip, anchor=center, draw=\d] at (\x,0.86) {\p};
}
\foreach \x/\h/\p in {4.60/0.30/near, 5.55/0.25/{next to}} {
  \fill[raInk!18] ({\x-0.13},1.02) rectangle ({\x+0.13},{1.02+\h});
  \draw[->, line width=0.5pt, draw=raInk!30] (\x,{1.24+\h}) -- (\x,{1.08+\h});
  \node[chip, anchor=center] at (\x,0.86) {\p};
}
\foreach \x/\h/\p in {7.60/0.33/behind, 8.75/0.05/wearing} {
  \fill[BrickRed!30] ({\x-0.13},1.02) rectangle ({\x+0.13},{1.02+\h});
  \draw[->, line width=0.5pt, draw=BrickRed!70] (\x,{1.24+\h}) -- (\x,{1.08+\h});
  \node[chip, anchor=center] at (\x,0.86) {\p};
}
\foreach \a/\b in {0.73/3.10, 4.30/5.94, 7.21/9.18}
  \draw[raInk!30, line width=0.4pt] (\a,0.73) -- (\a,0.68) -- (\b,0.68) -- (\b,0.73);
\node[note, anchor=north] at (1.92,0.63)
  {the synonym group is one positive,\\ a weighted mean over its members};
\node[note, anchor=north] at (5.12,0.63)
  {often true when the label is:\\ discounted by $1-\hat{P}(q \mid p)$};
\node[note, anchor=north] at (8.20,0.63)
  {full weight; the inverse\\ is never softened};

\node[band, text=raInk, anchor=south west] at (10.35,1.80) {SCORED RELATIONS};
\draw[raRule, line width=0.4pt] (10.35,1.74) -- (18.60,1.74);
\foreach \y/\sb/\sc/\sn/\p/\ob/\oc/\on/\s/\w in {%
    1.50/2/rg1/bus/following/1/rg0/bus/0.83/2.49,
    1.14/4/rg3/street light/illuminating/1/rg0/bus/0.81/2.43,
    0.78/5/rg4/car/parked by/3/rg2/building/0.75/2.25,
    0.42/1/rg0/bus/passing/5/rg4/car/0.61/1.83} {
  \node[badge, fill=\sc] at (10.50,\y) {\sb};
  \node[anchor=west, font=\fnote, text=raMuted] at (10.72,\y) {\sn};
  \node[anchor=west, font=\fsmall, text=raInk] at (12.10,\y) {\p};
  \node[badge, fill=\oc] at (13.70,\y) {\ob};
  \node[anchor=west, font=\fnote, text=raMuted] at (13.92,\y) {\on};
  \fill[raRule] (14.95,{\y-0.055}) rectangle (17.95,{\y+0.055});
  \fill[raAccent!70] (14.95,{\y-0.055}) rectangle ({14.95+\w},{\y+0.055});
  \node[anchor=east, font=\fsmall, text=raInk] at (18.58,\y) {\s};
}
\end{tikzpicture}

%% file: sec/04_corpus.tex
\section{\dataset{}: a verified free-text relation corpus}
\label{sec:corpus}

The model described above requires a corpus whose predicate vocabulary is not
that of a benchmark, whose annotation is dense enough that unannotated pairs
are rare, and whose errors are of a kind that can be detected. None exists, so we generated one: 474{,}413 images, 4{,}282{,}531 relations,
10{,}102 distinct free-text predicates and 9.03 relations per image, at a cost
of 104 GPU-hours. The pipeline, the statistics and a comparison with existing
machine-annotated corpora are in \cref{app:corpus}.

\paragraph{Motivation}
Relation annotation is among the most expensive and least reliable forms of
annotation in vision. An object box has one correct answer, whereas a relation
is a claim about a pair, of which an image with $N$ objects contains $N(N-1)$,
and for most pairs whether the claim holds is a matter of judgement. When
Visual Genome's annotators were free to write what they saw they produced
36{,}549 distinct predicate strings \citep{vg}; the community retained the 50
most frequent \citep{xu2017}, and the discarded strings are precisely what an
open-vocabulary model is intended to cover. A vision--language model annotates
no single relation better than a person, but it annotates more densely and
uniformly at marginal cost, and its errors are of a kind a geometric check can detect. We therefore use it as the annotator, not the predictor.

\subsection{Generation and geometric verification}
\label{sec:corpus:gates}

Generation runs an open-weight vision--language model \citep{gemma} on images
with \emph{numbered markers}, one coloured dot carrying the box index at the
centre of each box, so that grounding is an input to the annotator rather
than an inference from its output. Every proposed relation
then passes through a deterministic geometric gate (\cref{alg:verify} in
\cref{app:corpus}), which rejects 11.3\% of raw candidates; \cref{fig:gates}
breaks this figure down by gate. To our knowledge, no other machine-annotated
relation corpus verifies its annotator (\cref{tab:gen}).

The design principle is that a gate fires only when box geometry logically
constrains the predicate. A rejection is then a true negative up to
box-annotation error, whereas predicates geometry cannot constrain pass
through unchecked and are counted as residual risk; a gate that also guessed
on those would replace a known gap in coverage with an unknown error rate.
Verification rejects a relation or swaps its roles; the predicate string
stays the annotator's own, except in the balancing step below. After gating,
0\% of containment and 0.1\% of contact relations have disjoint boxes, and
the corpus has no duplicate triples and no self-loops. Direction repair is
the one non-rejecting gate: for left/right and above/below the roles are
swapped to match the boxes and the surface form kept. Since the annotator
prefers the left and upper object as subject, each verified left/right,
above/below and front/behind relation is then restated from the other
endpoint with probability one half, roles swapped and the predicate replaced
by its inverse (\pred{to the left of} becomes \pred{to the right of}); the
geometric top-up is balanced the same way. This leaves the corpus
direction-balanced, a property the model depends on and no source corpus
has.

The residual error classes are the geometrically unconstrained ones. Gaze is
the largest: roughly 22\% of \pred{looking at}/\pred{watching} annotations
have disjoint boxes and cannot be verified. Wrong-instance grounding,
prior-driven labels on very large boxes, contact predicates between
overlapping but unrelated boxes (\pred{person wearing tent} passes the
contact gate because the boxes do overlap), and mislabelled source boxes all
propagate into the corpus. One of these classes has a measurable downstream
effect, which we trace in \cref{sec:transfer:wearing}.

\subsection{Statistics and training mixture}
\label{sec:corpus:stats}

Relative to the source annotations on the same images and the same boxes,
\dataset{} is $1.7\times$ denser (9.03 versus 5.29 relations per image), has
$107\times$ the vocabulary and 1.4 nats higher predicate entropy
(\cref{tab:corpus}), and reproduces 73 of the 94 source predicate classes as
exact strings. Since the
images and boxes are identical, the comparison isolates the annotation schema.
The distribution remains long-tailed (the top ten predicates carry 57\% of the
mass), but the head predicates differ (\cref{fig:top50} in \cref{app:corpus}):
\pred{behind}, \pred{in front of}, \pred{wearing}, \pred{to the right of} and
\pred{to the left of}, spatial and balanced between inverses by construction, replace \pred{near} and \pred{on}. The mean object
degree rises from 1.88 to 3.20, and the share of images whose graph forms a
single connected component rises from 70.9\% to 82.6\%.

\pred{holding}, \pred{grasping}, \pred{carrying} and 535 further members of
the same family are retained as distinct strings, because a model whose
deployment vocabulary is arbitrary must be trained on targets that populate
the text space densely; canonical groups exist only within the loss and the
gates (\cref{app:corpus:decisions}). Every evaluation split is excluded by
image identity, using explicit VG$\leftrightarrow$COCO id tables and
perceptual near-duplicate search, which removes 8{,}687 corpus images and
32{,}456 raw Visual Genome images (\cref{sec:corpus:leak}).

\paragraph{Training mixture}
\label{sec:corpus:mixture}
\dataset{} is not used alone. We refer to one trained model as a \emph{tower},
that is, a backbone, a pair head and a training mixture, and we evaluate two
towers throughout. The \emph{zero-shot tower} is trained on \dataset{}
together with the leakage-filtered raw Visual Genome annotations, using all
of their free-text predicate strings rather than the 50 of VG150 (40{,}615
images, 758{,}547 relations, 17{,}742 predicate strings); the mixture is 91.9\%/8.1\% by image and
84.6\%/15.4\% by relation. This tower has seen no image from any evaluation split used in this paper
and none of the benchmarks' label sets; its Visual Genome share is raw
free-text annotation on images outside every evaluation split. The \emph{released tower} additionally
includes the HICO-DET training split at 21.0\% of images, which amounts to a
5.0\% relation share because HICO-DET annotates 1.94 relations per image
against 9.02 for the leakage-filtered \dataset{}, and uses the source-aware negatives of
\cref{app:objective}. Its mixture is 72.7/6.3/21.0 by image and 80.6/14.4/5.0
by relation over the three sources (\cref{tab:recipe}); \cref{tab:overlap}
lists each share alongside the shared triplet mass it produces. Both towers are
scored against the same 19{,}103-string predicate bank, so the zero-shot tower
carries 108 of HICO-DET's 116 predicate strings as columns without having been
trained on its relations. Unless a row is marked \emph{zero-shot}, the released tower is the
model reported in every table; on HICO-DET, where it is not zero-shot, the
zero-shot tower is reported alongside it. The two towers differ by less than
one point on every axis other than HICO-DET itself (\cref{tab:scaling} in
\cref{app:scaling}).

%% file: sec/05_protocol.tex
\section{\bench{}: analysis of scene-graph evaluation and a cross-dataset protocol}
\label{sec:bench}

We analyse what scene-graph recall rewards, identifying three priors that a
standard architecture can exploit by construction (\cref{sec:priors}) and two
scoring conventions whose effect exceeds the differences they are used to
measure (\cref{sec:protocol}), and then define the six-axis protocol on which
\cref{sec:results} reports (\cref{sec:bench:axes}). The model and corpus of
the preceding sections serve here only as measurement instruments; no finding
below depends on their design.

\subsection{Metrics and evaluation modes}
\label{sec:bench:metrics}

R@K denotes scene-graph recall, the fraction of an image's annotated
$\langle$subject, predicate, object$\rangle$ triplets appearing among the $K$
highest-scoring predicted triplets, averaged over images; since every
triplet counts equally, it is dominated by frequent predicates. mR@K is the
mean of the per-predicate R@K values, so a rare predicate contributes as much
as \pred{on}; F1@K is their harmonic mean; wR@K weights each predicate's
recall by the logarithm of its inverse frequency in the evaluated split; and
the rare, common and frequent buckets partition predicates by annotation
count in that split (below 50, 50 to 500, above 500), following LVIS
\citep{lvis}. Federated average precision (fAP) and P-AUC are
computed over labelled cells only and used on A2 alone
(\cref{sec:results:a2}).

Under \emph{graph-constrained} scoring each subject--object pair contributes
only its highest-scoring predicate to the $K$ slots, whereas unconstrained
scoring lets one pair occupy several; the difference is 11.6 R@50 on VG150 and
19.0 on PSG for our model. All numbers in this paper are graph-constrained, as
are the baseline's. Results come in one of two modes: with \emph{ground-truth
boxes} the regions are given and only the relations are scored, whereas in
\emph{detection mode} (SGDet) they come from a detector, so the score also
reflects detector recall, an unreported free parameter (\cref{sec:protocol}).

\subsection{Three priors rewarded by recall}
\label{sec:priors}

\paragraph{Shared triplet mass}
\label{sec:priors:overlap}
The VG150 test split uses the same 50 predicate strings as its training split,
so a VG150-trained model encounters no vocabulary novelty at all; nevertheless,
\emph{open-vocabulary} methods are ranked on exactly this number on the OvR-SGG
leaderboard \citep{ovsgtr_eccv}, which we reproduce in \cref{sec:transfer:ovr}.
We quantify this confound as the \emph{shared triplet mass}, the fraction of
a corpus's relation instances whose $\langle$subject category, predicate,
object category$\rangle$ triple the benchmark also annotates
(\cref{tab:overlap} in \cref{app:corpus}). The triple rather than the
predicate, because a predicate string is not an annotation: \pred{on} between
a person and a horse and \pred{on} between a book and a table are different
acts, and two corpora can agree on the string while never agreeing on the pair
it is asserted of. Our mixture shares 53.4\% of its mass with VG150 on the
predicate alone and 12.8\% on the triple. No image and no annotation is
shared; leakage is excluded separately by image identity
(\cref{sec:corpus:leak}). The confound is concentrated in-domain and is very
large there: the baseline's fine-tuning corpus reproduces 90.9\% of its
relation mass as triples VG150 also annotates, against our 12.8\%, whereas on
PSG it sits at 8.6\% against our 10.7\%. Because the triple charges for the
two object names as well, it also charges for taxonomy and domain mismatch,
and \cref{tab:massdecomp} prints the components separately. The statistic is
predictive. An arm trained with a larger share of raw Visual Genome reached
54.3 R@50 on VG150, the best zero-shot figure we are aware of, while being the
worst model we trained on every tail metric.

\paragraph{Object-category prior}
\label{sec:priors:freq}

\begin{table*}[t]
\centering\small
\caption{\textbf{A frequency table exceeds the trained model on the micro protocol and
falls below it on the macro protocol in every cell.} \textsc{freq}
\citep{motifs} returns the most frequent training predicate for a
$\langle$subject, object category$\rangle$ pair from ground-truth labels and
no pixels; ours is the zero-shot tower, which receives no labels. Per-edge
top-1 accuracy over ground-truth pairs and boxes. \emph{only ours}: edges we
predict correctly and \textsc{freq} does not; \emph{union}: the oracle
combination.}
\label{tab:priors}
\input{tables/priors}
\end{table*}

It has been known since \citet{motifs} that a lookup table over object
category pairs is a strong scene-graph baseline. \Cref{tab:priors} shows its
advantage to be confined to the metric by which methods are usually ranked: in
per-edge top-1 accuracy the frequency baseline exceeds our trained model by
15\% on all three benchmarks, while on the per-predicate average of the same
predictions our model exceeds it by 29--86\%. The baseline returns each
category pair's majority predicate, which a micro metric rewards and a macro
metric penalises. Two readings should be avoided. The table does not show that
visual input is unnecessary, since the edge-by-edge join (\cref{fig:join} in
\cref{app:results}) finds the model correct where the baseline is wrong on
6.9--12.8\% of edges; nor that our model is weak, since the baseline receives
oracle object categories we never see. The conclusion is narrower: a metric a
pixel-free lookup table can win does not measure relation understanding, and
it is the metric that orders leaderboards. A model given object class labels
obtains this prior for free, and because it is estimated from one corpus's
label statistics it is also the component that does not survive a change of
corpus.

\paragraph{Annotation propensity}
\label{sec:priors:propensity}
The third prior lies in what the annotation omits. A few pairs per image are
annotated and the rest left empty, but an empty pair is not a negative one: on
PSG test at its maximum-F1 threshold, 78\% of the edges a \model{} checkpoint
emits fall on pairs the corpus never annotated in either direction, against
11\% matching an annotated triple (\cref{app:threshold}). A model trained to
predict which pairs carry annotations therefore learns \emph{annotation
propensity}, which on these corpora reduces to a contact prior. Our
architecture contains such a term by construction, the pair-relatedness score
of \cref{sec:model:arch}, so its contribution can be measured, and its sign
differs by benchmark family: retaining it improves recall on the
annotation-derived benchmarks and degrades truth judgement on the adjudicated
negatives of SpatialSense (\cref{sec:priors:relatedness} in
\cref{app:protocol}). Recall against sparse annotation and truth judgement
against adjudicated negatives are therefore different quantities.

A protocol that is robust to these three priors should report a vector of
measurements rather than a single number, state the shared triplet mass
behind every recall cell, and include at least one axis scored against adjudicated
negatives and one that a frequency prior cannot help.

\subsection{Two scoring conventions with effects larger than those measured}
\label{sec:protocol}

Whereas \cref{sec:priors} concerned what the benchmarks ask, this section
concerns how they score. We encountered both conventions while reproducing
OvR-SGG \citep{ovsgtr_eccv}, the open-vocabulary relation leaderboard, which
holds 15 of the 50 VG150 predicates out of training and scores in detection
mode. The reproduction is faithful: running the released checkpoints through
our inference code and scoring them with a transcription of the baseline's
metric recovers all twelve published numbers to within 0.35 points
(\cref{tab:protocol} in \cref{app:ovr}). The observations below concern the protocol itself.

\paragraph{Duplicate credit in the matcher}
The standard detection-mode matcher, inherited from the reference
implementation \citep{motifs,sgbench} and still used by current
open-vocabulary methods, credits a predicted triplet when its subject and
object boxes each reach IoU $\ge 0.5$ with the corresponding ground-truth
boxes and the label triple matches. No assignment constraint is enforced, so
a ground-truth object covered by $d$ detections offers $d$ opportunities to
recover the same relation; on VG150 test, each recovered ground-truth box is
covered by 2.99 detections on average. Object detection has not permitted
this since COCO \citep{coco}, where each ground-truth instance may be matched
by at most one detection and the remainder count as false positives.
Restoring the constraint reduces the published model's R@50 by 12.0\% and its
novel R@50 by 13.5\%, consistently across both released checkpoints. This
amounts to 2.45 points of Base+Novel R@50, compared with the 2.39 points
gained by upgrading the same method's backbone from Swin-T to Swin-B, which
constitutes a separate leaderboard entry. The correction applied here is the
most generous available, matching each ground-truth object to the
highest-IoU detection of the correct class; a class-agnostic assignment costs
more (25.5 down to 15.2 R@50).

\begin{figure}[t]
\centering
\includegraphics[width=0.7\linewidth]{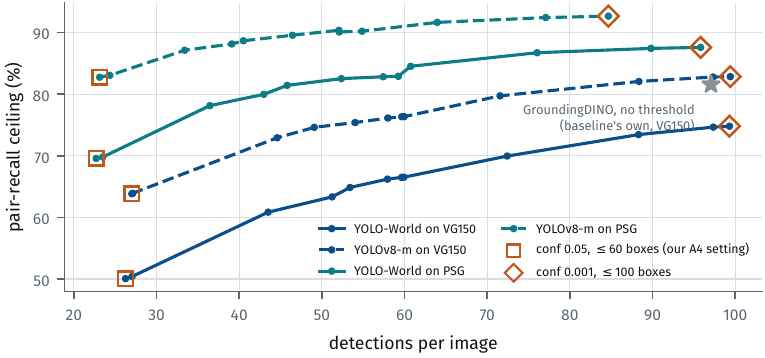}
\caption{\textbf{The pair-recall ceiling as a function of the detector operating
point.} One curve per detector and benchmark, weights fixed, swept over
confidence threshold and box cap; markers are the two operating points of
\cref{tab:boxbudget} (\cref{app:protocol}). The 4.3-point spread between all
published OvR-SGG methods fits inside the vertical extent of any single curve.
The baseline's GroundingDINO \citep{groundingdino} retains every non-background query, the source of its apparent advantage.}
\label{fig:boxbudget}
\end{figure}

\paragraph{Detector operating point}
Detection-mode recall is bounded above by \emph{pair recall}, the fraction of
ground-truth relations for which both endpoints are detected. Pair recall is
quadratic in object recall, bounds every model evaluated on the same boxes
identically, and depends not on the detector itself but on its confidence
threshold and box cap, parameters that are rarely reported. Holding the
detector weights fixed and varying only these two settings moves the ceiling
by 12--49\% relative across four detector/benchmark combinations
(\cref{fig:boxbudget}); on VG150 the change is 19 points, against a 4.3-point
spread between all published methods on the leaderboard. A threshold can also
invert a comparison between detector families. The baseline's GroundingDINO
\citep{groundingdino} appears stronger than a YOLO detector, with a ceiling
of 0.82 against 0.64, because it applies no threshold and retains 97 boxes
per image against the YOLO detector's 27; at a matched budget the YOLO
detector reaches 0.83. Our own A4 is measured at a confidence threshold of
0.05 against a ceiling of 0.696, whereas the same detector reaches 0.876 at a
matched budget. Detection-mode results should therefore state the matcher,
the detector operating point and the resulting pair-recall ceiling;
\cref{sec:discussion} lists the reporting practices this implies.

\subsection{The six axes}
\label{sec:bench:axes}

\bench{} reports a vector of six axes, summarised in \cref{tab:axes}. Each axis
answers a question that no other axis can, and each is listed together with
the way in which it fails when read alone, which is the reason for reporting a
vector. Two axes are prior-blind by construction: A2 is scored against
negatives that a human adjudicated, so annotation propensity cannot help, and
A6 is balanced by design, so the category prior cannot help. Every claim in
this paper about relational competence, as opposed to corpus match, rests on
these two axes.

\begin{table}[t]
\centering\footnotesize
\caption{The six axes of \bench{}; the third column states why no single axis
suffices.}
\label{tab:axes}
\begin{tabularx}{\linewidth}{@{}l>{\raggedright\arraybackslash}X>{\raggedright\arraybackslash}X@{}}
\toprule
& what it measures & how it fails alone \\
\midrule
A1 \note{transfer} &
closed-vocabulary recall on ground-truth boxes, four sources of differing
shared triplet mass & can be satisfied by corpus match \\
A2 \note{precision} &
Haystack \citep{haystack}: federated AP and P-AUC against explicit negative
annotations; the only axis on which a confident false positive on a rare
predicate is visible & invariant to uniform score depression: a model that
never asserts a relation still ranks well \\
A3 \note{open vocab} &
the full 19{,}103-string vocabulary deployed, synonym-tolerant matching &
rewards head collapse: \pred{on} and \pred{in} are members of most synonym
sets \\
A4 \note{deployment} &
detection-mode graphs on a \emph{shared} open-vocabulary detector, against the
measured pair-recall ceiling & bounded by the detector \\
A5 \note{graph quality} &
the information in the relations that a vision--language judge without ground
truth accepts one at a time (\cref{eq:truebits}); the only axis that can credit
a true but unlabelled relation & reported only when the judge's controls pass;
shown whole graphs, the judge prefers short repetitive ones, and that verdict
is reported alongside \\
A6 \note{spatial} &
SpatialSense \citep{spatialsense}: adversarially collected, balanced true/false
triples, chance 50\% & a boxes-only baseline scores 68.8 \emph{without the
image}: a high score indicates resistance to priors rather than visual
reasoning \\
\bottomrule
\end{tabularx}
\end{table}

On A3, a prediction is counted as correct when it matches the ground truth
under a synonym matcher at a cosine threshold calibrated to one embedding
space. A threshold fitted in one space and reused in another accepted 0.6\% of
true synonyms and reduced the axis to zero (\cref{app:closed:directions}).

\paragraph{Composite}
The axes are combined into a single scalar, OVS, over A1, A2, A4, A5 and A6.
Each cell is chance-corrected, A1 entering as F1@50 averaged over its four
benchmarks, and the axes are combined by a harmonic mean,
which penalises imbalance, following the practice in generalised zero-shot
learning. Two cells are normalised by a measured quantity rather than a chance
level, A4 by the pair-recall ceiling of the shared detector and A5 by the
information in the images' own annotations, so that each reports the share of
what is recoverable that a model delivers. A3 is reported but excluded, since the
only baseline runnable on it is scored through a matcher whose choice moves
the result by more than the systems differ (\cref{sec:results:a3}). The composite does
not replace the vector; it exists to keep model selection from optimising one
axis. Withholding each axis in turn leaves the ordering of the two systems
unchanged, at ratios between $2.0$ and $3.7\times$ (\cref{tab:ovsloo}).


\paragraph{Baselines} Each axis is reported against the strongest system we
could run on it. OvSGTR \citep{ovsgtr} pre-trained on MegaSG, the corpus whose
images we re-annotate, is the closest control for supervision quality and the
only system runnable on all six; we run its released checkpoint through our
evaluator in its relation-open-vocabulary mode, the only configuration with no
frequency prior and a head that accepts an arbitrary vocabulary. One residual
asymmetry favours it throughout: its protocol consumes object labels, whereas
ours never does. ROBIN-3B \citep{svg} and four general multimodal models
answer in free text instead, which makes them the only systems runnable on A3
and, in ROBIN's case, the stronger baseline on A1; they need a matcher, whose
choice \cref{sec:results:a3} shows to move the result by more than the systems
differ. The remaining open-vocabulary methods of \cref{sec:related} lack released weights or an inference path that accepts our regions and vocabulary, and appear on the leaderboard of \cref{sec:transfer:ovr} instead; the omission reflects what we could run, not a judgement.

%% file: tables/priors.tex
\begin{tabular}{lr rrr rrr rr}
\toprule
 & & \multicolumn{3}{c}{micro (per edge)} & \multicolumn{3}{c}{macro (per predicate)} & \multicolumn{2}{c}{join} \\
\cmidrule(lr){3-5}\cmidrule(lr){6-8}\cmidrule(lr){9-10}
benchmark & edges & \textsc{freq} & ours & $\Delta$ & \textsc{freq} & ours & $\Delta$ & only ours & union \\
\midrule
VG150 & 152,535 & 68.4 & 57.7 & -15.6\% & 18.9 & 35.1 & +85.7\% & 6.9 & 75.3 \\
PSG & 13,623 & 50.9 & 43.3 & -15.0\% & 20.7 & 31.6 & +52.4\% & 12.8 & 63.7 \\
IndoorVG & 29,175 & 67.9 & 57.3 & -15.5\% & 29.8 & 38.4 & +28.9\% & 8.2 & 76.2 \\
\bottomrule
\end{tabular}

%% file: sec/06_results.tex
\section{Results}
\label{sec:results}
\label{sec:transfer}

\begin{table*}[t]
\centering\small
\caption{\textbf{The six axes of \bench{}}, released model against the best
baseline available on each axis, one evaluator for all. Each baseline cell
names its system: ROBIN-3B \citep{svg} leads OvSGTR \citep{ovsgtr} on A1 and
is the only one of the two runnable on A3, and OvSGTR is the only system run
on A2, A4, A5 and A6. That column is an envelope over two systems rather than
a system, so the composite is computed for OvSGTR alone. A5 is
\cref{eq:truebits}; A6 is the macro mean of per-predicate AUC, pooled in
parentheses; the HICO-DET cell uses the zero-shot tower.}
\label{tab:sixaxes}
{\footnotesize\resizebox{\linewidth}{!}{\input{tables/sixaxes}}}
\end{table*}

\Cref{tab:sixaxes} summarises all six axes, discussed in turn below. No result
is in-domain, the regime in which a corpus-matched prior is correct and the
priors of \cref{sec:priors} are invisible: every cell is measured on a split whose
images and annotations were excluded from training, by a tower that never
trained on the benchmark's label set, the sole exception being the marked
HICO-DET row of \cref{tab:a1}. One evaluator scores
every system, and the shared triplet mass is printed beside each cell
(\cref{sec:priors:overlap}); no images or annotations are shared.

\subsection{A1: cross-dataset transfer on four benchmarks}
\label{sec:transfer:a1}

\begin{table*}[t]
\centering\small
\caption{\textbf{Axis 1: cross-dataset transfer.} Test splits, ground-truth boxes,
graph-constrained, one evaluator, OvSGTR \citep{ovsgtr} throughout.
\emph{triplet mass} is the share of \emph{our} training relations whose
subject--predicate--object triple the benchmark also annotates
(\cref{sec:priors:overlap}); the baseline's is 90.9\%, 8.6\% and 10.4\% on
VG150, PSG and IndoorVG. No images or annotations are shared.
\emph{rare} is recall on the rare support bucket. The baseline receives
ground-truth object labels; \model{} receives none. On HICO-DET the released
model has seen a 5\% relation share of the training split and is reported
alongside the zero-shot tower.}
\label{tab:a1}
\resizebox{\linewidth}{!}{\input{tables/a1_transfer}}
\end{table*}

As \cref{tab:a1} shows, \model{} outperforms OvSGTR on every metric of every
source, including micro recall, though it receives no object labels and the
baseline is given ground-truth ones. The margins vary: head recall is
$1.04$--$1.40\times$ the baseline's, mean recall $2.3$--$3.5\times$ and
rare-bucket recall $5$--$21\times$, the ratio being undefined on VG150 where
the baseline scores exactly 0.0. This is the pattern the prior-matching
account of \cref{sec:priors} predicts: VG150 is OvSGTR's own benchmark, where
90.9\% of its triplet mass is shared against our 12.8\%, and its headline
number there is carried by 21 frequent predicates, the part of the
distribution recoverable without visual input. OvSGTR is not the strongest
baseline on this axis, however. ROBIN-3B leads it on F1@50 on all three
benchmarks both were run on, 19.6 / 27.9 / 22.7 against 16.5 / 13.5 / 20.2,
which is why \cref{tab:sixaxes} carries ROBIN in the A1 cells; it too stays
behind us on both metrics everywhere (\cref{tab:baselinescross}). Our own
micro recall warrants the same scepticism, which is why \cref{tab:a1} carries
the triplet-mass column. The strongest systems we run are not on this axis at
all: the in-domain specialists of \cref{sec:efficiency} lead us on F1@K on PSG
and IndoorVG once a detector supplies the boxes (\cref{tab:react}).

\subsection{A2: precision on adjudicated negatives}
\label{sec:results:a2}

Recall does not penalise hallucination. Since an unlabelled pair is not a
negative pair, a model can assert false relations at no cost in recall. A2
addresses this by scoring against negatives that an annotator adjudicated
rather than negatives assumed by the protocol.

A2 is evaluated on Haystack \citep{haystack}, the only scene-graph test set we
are aware of that publishes explicit negatives, that is, cells an annotator
inspected and marked as not carrying the predicate, as distinct from cells
left blank. Annotating per predicate rather than per image reaches rare
classes that the PSG test split barely contains and makes its positives
sparse: 2{,}870 positives against 23{,}174 explicit negatives over 11{,}368
images. It reuses PSG's 56 predicates and panoptic format, so a model runs on
it unchanged, and its images come from SA-1B \citep{sam}, so it shares no
images or annotation ancestry with VG150, IndoorVG, PSG or the Visual Genome
component of our mixture. Scoring is federated in the sense of LVIS
\citep{lvis}, over the 26{,}044 labelled cells only, on ground-truth boxes; we
report Haystack's own per-predicate ROC-AUC (P-AUC) together with a federated
AP that ranks within each predicate.

The results are given in \cref{tab:axesdetail}. \model{} improves mean
federated AP by 39\% and rare-predicate fAP by 59\%, over 55 predicates with
support of at least five, 31 of which are rare, and raises P-AUC from 83.9 to
93.7. Both models express all 56 predicates, but reach different fractions of
the cells. The baseline scores every ordered pair; our relatedness sampler
reaches 89.0\% of the 26{,}044 labelled cells, and we do not force the rest
through it, since an unsampled pair is one the deployed system emits nothing
for. Those cells are scored zero: a correct rejection for a negative, a miss
for a positive. We note that the same predictions, scored by counting every
unlabelled pair as a false positive, yield an AP of 0.035, whereas scored
against explicit negatives they yield an fAP of 0.56; the difference lies
entirely in what the protocol assumes about unannotated pairs.

\subsection{A3: the full vocabulary with the answer set withheld}
\label{sec:results:a3}

A1 supplies the benchmark's vocabulary to the model, which a deployed open-vocabulary model does not have. A3 removes this assistance: the
model answers from its full 19{,}103-string vocabulary and is scored with
synonym-tolerant matching. Recall does not collapse (56.0 R@50 on VG150, against 53.3 with the benchmark's own 50 strings supplied), and the
median rank of the ground-truth string among 19{,}103 candidates is 1 on VG150
and IndoorVG and 10 on PSG (\cref{tab:axesdetail}).

\paragraph{Applicability to the baseline} OvSGTR scores relations against
text, but receives it as a single caption encoding the predicate names
jointly, so its vocabulary is bounded by the encoder's 512 word pieces, about
150 strings, and splitting 19{,}103 strings across 140 captions does not
recover the measurement, since each name's embedding depends on the others in
its caption. Ours are embedded independently and offline. This axis is one a
model must be designed for.

\paragraph{Free-text systems}
A model that answers in free text can be evaluated on this axis by
construction, which admits a family of baselines that the other axes cannot
accommodate: ROBIN-3B \citep{svg}, a scene-graph model built on a 3B
vision--language model, and general multimodal models prompted for the same output: Qwen3-VL \citep{qwen3vl}, InternVL3.5 \citep{internvl35} and
GLM-4.6V-Flash \citep{glmv}. \Cref{tab:baselines} evaluates them on ground-truth
regions, so that relation prediction is measured independently of
localisation; prompts are in \cref{app:baselines}.

\begin{table}[t]
\centering\small
\caption{\textbf{Free-text systems on the open-vocabulary axis.} Ground-truth
regions for every model, graph-constrained, one scorer; ours is the 53M
ViT-S/16+ tower. PSG test, 2{,}179 images: one set of generations per model scored under the
matchers in circulation, so the rows of a system differ from one another only
in how its free text is mapped onto the benchmark's words. The general models
are reported under our A3 rule alone. Our A3 rows use the $\tau=0.60$ matcher
shared by every system here rather than the calibrated threshold of
\cref{tab:axesdetail}, hence the different PSG figures.
\emph{Pair cov.} is the share of annotated pairs a system reaches at all;
ours at top-$K$ as elsewhere. The cross-dataset comparison, closed-set prompt and SGDet leg are in \cref{app:baselines}.}
\label{tab:baselines}
{\footnotesize\resizebox{\linewidth}{!}{\input{tables/baselines_psg}}}
\end{table}

The matcher is a third scoring convention whose effect exceeds the differences
it is used to measure. \Cref{tab:baselines} is one set of generations per
model read up to three ways: mapping free text onto the benchmark's words is
worth $+11.7$ R@50 to ROBIN, from 26.2 exact to 37.9 at $\tau=0.60$, and
$+23.2$ to us, since a model answering from 19{,}103 strings is the one exact
matching penalises hardest (\cref{sec:scale:region}). The choice decides the
mean-recall ordering of the two purpose-built systems: ROBIN leads under exact
matching, 20.0 against 13.0, and we lead under every synonym-tolerant matcher,
31.3 against 25.1 at $\tau=0.60$. Micro recall does not turn over, ROBIN
staying ahead throughout, so the two metrics disagree in the direction
\cref{sec:priors} predicts. A single row is therefore a choice of scorer
rather than a measurement of a model, and we report the band.

Two things survive the matcher. One is which pairs a system proposes: pair
coverage is 99.7\% for us, 45.6--77.4\% for ROBIN and 23.1--35.5\% for the
general models, and a pair never named cannot carry the right predicate; scale
buys mostly this, 8B to 32B adding $+11.9$ points of it. The other is what
happens off PSG. Run on three benchmarks, each system under its own protocol
(\cref{tab:baselinescross}), we lead ROBIN on both metrics on all three, by
$1.9\times$ on VG150 micro recall and $2.2\times$ on IndoorVG against
$1.1\times$ on PSG, the benchmark on which such systems are usually compared.

\subsection{A4: detection-mode graphs on a shared detector}
\label{sec:transfer:deploy}

\begin{table*}[t]
\centering\small
\caption{\textbf{Axes 2--4 in detail.} \emph{A2} (top left): precision on the explicit
negatives of Haystack \citep{haystack}; fAP ranks within a predicate, P-AUC is
Haystack's own per-predicate ROC-AUC. \emph{Pair coverage} is the share of
labelled cells whose pair a model scores at all: complete for the baseline,
which scores every ordered pair, and 89.0\% for ours, whose sampler prunes and
whose unsampled cells score zero. Both express all 56 predicates. \emph{A3} (top right): all
19{,}103 training strings, synonym-tolerant matching at the calibrated
threshold; \emph{MRR} and \emph{median rank} score the ground-truth string
against that vocabulary. \emph{A4} (bottom): PSG test, graph-constrained, IoU
matching, on a shared YOLO-World detector \citep{yoloworld} prompted with each
benchmark's categories at confidence 0.05 and at most 60 boxes per image; the
pair-recall ceiling bounds both models identically (\cref{fig:boxbudget}), and
the baseline additionally consumes the detector's predicted labels.}
\label{tab:axesdetail}
{\footnotesize\setlength{\tabcolsep}{4pt}%
\begin{minipage}[t]{0.57\linewidth}\centering
{\sffamily\bfseries A2 precision (Haystack, ground-truth boxes)}\\[3pt]
\resizebox{\linewidth}{!}{\input{tables/a2_precision}}
\end{minipage}\hfill
\begin{minipage}[t]{0.41\linewidth}\centering
{\sffamily\bfseries A3 open vocabulary (19{,}103 strings)}\\[3pt]
\resizebox{\linewidth}{!}{\input{tables/a3_openvocab}}
\end{minipage}\\[8pt]
{\sffamily\bfseries A4 detection mode (PSG test, shared YOLO-World boxes)}\\[3pt]
\input{tables/a4_deployment}}
\end{table*}

The transfer results above use ground-truth boxes; in \cref{tab:axesdetail} a
shared open-vocabulary detector supplies the regions and both models are
re-scored against the measured pair-recall ceiling of 0.696. Every margin of
\cref{tab:a1} survives with the same structure (mR@50 $3.3\times$, rare bucket
$14\times$), so the relation model is unaffected by the change of box source
and what it learned transfers. The dominant loss is the detector's. Under a
one-to-one matcher that ignores the detected class, retention relative to
ground-truth boxes is 62--63\% on PSG and 31--32\% on IndoorVG across all
three towers; when the matcher also requires the detected class to agree, as
in \cref{sec:protocol}, it is 46\% and 19\% (\cref{fig:detbox} in
\cref{app:results}). A relation cannot be recovered unless both endpoints are
detected, and neither a better relation model nor more relation annotation
raises pair recall. Both models are scored on identical boxes, so the
comparison is unaffected by the operating point; the absolute numbers are
specific to a confidence threshold of 0.05.

\subsection{A5: graph quality judged without ground truth}
\label{sec:results:a5}

\begin{table*}[t]
\centering\small
\caption{\textbf{A5: \cref{eq:truebits} and its decomposition}, PSG test, the shared
YOLO-World boxes of A4, one judge, restricted-vocabulary arm. Each relation
the judge accepts is credited with its surprisal under the PSG training
marginal over the 56 predicates both systems emit; the last column is the
product. The judge is asked only whether the relation is true. The two blocks
differ in rendering as well as depth, and each carries its own corruption-acceptance control (\cref{app:oracle:relation}). The headline cell of
\cref{tab:sixaxes} is the matched block.}
\label{tab:a5b}
{\footnotesize\input{tables/a5b}}
\end{table*}

\begin{figure*}[t]
\centering
\input{figures/a5_qual.tex}
\caption{\textbf{Two graphs on the same boxes.} Two PSG test images from the A5 run,
the shared YOLO-World boxes of A4 drawn and numbered, and the top ten
relations each system emits from the benchmark's 56 predicates. Object names
are the detector's, shown for the reader only. \pred{on} is set in
grey; each predicate carries its surprisal in bits under the PSG training
marginal, which \cref{eq:truebits} credits when the judge accepts it.
The baseline's graphs consist mostly of \pred{on}; ours convey more per
relation and include claims a judge rejects.}
\label{fig:a5qual}
\end{figure*}

A5 addresses a question none of the other axes can: whether, without ground truth, the emitted graph describes the image well. A
vision--language judge from a different model family than the annotator is
shown the image with numbered boxes and asked about a system's relations (\cref{app:oracle}).

The judge yields three verdicts, two of which favour the baseline. Shown whole
graphs it prefers the baseline's at deployed length and ours at equal length,
and shown single relations it accepts the baseline's at a slightly higher
rate; but each relation it accepts from the baseline conveys about a third
less information, because most of them are \pred{on}. The three verdicts are not in
conflict, and the axis is scored by the third.

\paragraph{Per-relation judgement} The judge is shown one relation at a time,
with the subject and object boxes drawn on the image, and asked whether the
claim holds for that photograph; a fixed share of relations are first
corrupted by substituting another predicate from the same image, and the run
reports how often a corruption is accepted. Truth alone favours \pred{on},
which is almost always true and almost never informative, so a system
answering \pred{on} everywhere would be accepted more often than either model
here. Asking the judge to rate informativeness does not resolve this: it rated
the baseline's \pred{on} relations at 1.05 on a scale of 0--3 against 0.72 for
its other predicates, the reverse of the intended ordering. The redundancy of
the five-hundredth \pred{on} is a property of the distribution it occurs in,
which a judge shown one relation at a time cannot observe.

\paragraph{Measuring informativeness}
The information carried by a predicate is its surprisal under a reference
distribution, and the axis is defined as the total surprisal of the relations
the judge accepted, per image:
\begin{equation}
\label{eq:truebits}
I \;=\; \sum_{r \,:\, \mathrm{judge}(r) \,=\, \mathrm{true}}
        -\log_2 p_{\mathrm{ref}}\!\left(\mathrm{pred}(r)\right).
\end{equation}
The two factors compensate for each other's weakness. Surprisal alone would
reward rare predicates emitted at random, which the judge rejects as false;
truth alone would reward \pred{on}, priced at 2.31 bits against 7.04 for
\pred{riding}. Neither can be increased by repetition. The reference is the PSG
training marginal, shared by both systems and restricted to the 56 predicates
both emit, so no smoothing constant influences the comparison.

\Cref{tab:a5b} gives the decomposition. At matched depth, that is, the top ten
pairs of each system, the judge accepts slightly more of OvSGTR's relations
than of ours, 4.33 against 4.07 per image. Each accepted relation is, however,
worth 3.09 bits against our 4.57, so the axis reads 13.4 bits per image for
the baseline against 18.6 for ours, and 14.1 against 28.7 at deployed depth. A
reference-free estimator that charges a system for the size of the vocabulary
it uses agrees on the ordering (\cref{app:oracle:relation}). Since bits per
image have no upper bound, \cref{tab:sixaxes} and \cref{fig:overview} divide
by the same quantity evaluated on the annotation itself: the ground-truth
relations of those 199 images carry 27.9 bits under the same reference, and
at matched depth we deliver 0.67 of that against 0.48 for OvSGTR. The ratio
is not bounded by one (at deployed depth ours is 1.03) because the annotation
is sparse, which is why the matched-depth cell is the one that enters the
composite.

\paragraph{Corruption control} Each depth carries its own: the judge accepts
0.135 of specific corruptions at deployed depth and 0.201 at matched depth, a
difference due to the rendering rather than the depth
(\cref{app:oracle:relation}). The headline cell is thus the arm with the
weaker control. We report it nonetheless, because that rate is an upper bound
rather than a false-positive rate, a swapped predicate sometimes being true of its new pair, and because the ordering holds under both renderings and depths.

\paragraph{Whole-graph comparison} Shown two whole graphs, in both
presentation orders and with several votes each, the judge chose the
baseline's in 49 of 58 decisive comparisons in the restricted-vocabulary arm
and 35 of 46 in the full-vocabulary arm, with all controls passing
(\cref{tab:a5} in \cref{app:oracle}). Two factors confound the verdict. Our
graphs are longer, 18.3 relations against 10.3, and the baseline's was never
the longer of the two; a separate run in which each system contributes its top
ten pairs reverses the outcome, 40 of 64 decisive comparisons in our favour.
The graphs also differ in composition: 66\% of the baseline's relations are
\pred{on}, and averaged per graph its modal predicate accounts for 0.85 of
the relations against 0.52 for ours (\cref{fig:a5qual}). A graph-level verdict cannot
separate the two readings, that our additional relations are false or that the
judge prefers shorter text, and a verdict at equal length says nothing about
what the additional relations are worth. The axis is therefore scored by
\cref{eq:truebits}, the only verdict of the three that does not depend on
graph length.

\subsection{A6: spatial relations against adversarial negatives}
\label{sec:transfer:spatial}

A6 is the axis on which a frequency prior cannot help. Without it, changes
that moved recall by tens of percent while leaving the projective predicates
near chance (\cref{fig:spatial} in \cref{app:results}) would have looked like
improvements in relational understanding. The sigmoid auxiliary and the
deformable read together are what moved them, raising \pred{in front of} from
0.58 to 0.68, \pred{above} from 0.54 to 0.60 and \pred{to the left of} from
0.56 to 0.64 on the released tower, for a macro AUC of 0.690 and a pooled AUC
of 0.675. This axis remains our weakest.

The baseline row evaluates OvSGTR on the same 2{,}758 cells with the same
metric implementation, from ground-truth boxes and at full coverage: macro AUC
0.591, pooled 0.617. We are ahead on all nine predicates, and on three the
baseline is at or below chance (\pred{above} 0.495, \pred{to the right of}
0.489, \pred{under} 0.515), the insensitivity to projective structure the axis
was designed to detect. Neither model reaches the 68.8\% of the original
authors' boxes-only baseline, which does not see the image: we reach 62.5\%
and the baseline 59.0\%. An AUC near 0.69 is therefore not evidence of visual
spatial reasoning, and \cref{app:results} lists the probes bounding this
claim. We read the remaining gap as a limit of supervision rather than
capacity, since capacity does not affect it (\cref{app:scaling}).

\subsection{The OvR-SGG leaderboard}
\label{sec:transfer:ovr}

\begin{table}[t]
\centering\small
\caption{\textbf{OvR-SGG leaderboard} (VG150 test, SGDet, 35 base / 15 novel
predicates). Our rows use the \emph{baseline's} matcher, Swin-T detector
boxes, predicted labels and ranking rule. The two rows marked \note{novel set
seen} have the 15 novel strings in their training vocabulary, so their Novel
columns are not held-out; the row above them is the retrained model satisfying
the protocol. Mean recall and the class count are measured here for both re-run systems and reported by no published row (\note{n/r}). Bold: the better of
the two measured here.}
\label{tab:ovrleaderboard}
\resizebox{\linewidth}{!}{\input{tables/ovr_leaderboard}}
\end{table}

Since \cref{sec:protocol} reproduces OvR-SGG exactly, we can also report on
the leaderboard itself, scored by the baseline's matcher rather than ours
(\cref{tab:ovrleaderboard}): VG150 in detection mode, 15 of 50 predicates held
out of training, our rows scored with a transcription of the baseline's metric
on its own detector boxes, predicted labels and ranking rule. \cref{app:ovr}
documents the two incompatible predicate splits in circulation and why the
table retains the uncorrected matcher.

\paragraph{Composition of the novel split}
Recall at $K$ on this split is a micro average, and the 15 novel predicates
carry 55\% of the test relations because \pred{on}, \pred{of} and \pred{in}
are among them and account for 93\% of the novel mass by themselves. Novel
recall is therefore, to within a few points, recall on three of the most
frequent strings in Visual Genome, and the baseline scores non-zero on 3 of
the 15. Mean recall makes this visible, which is why \cref{tab:ovrleaderboard}
adds it for the two systems we evaluate ourselves: the baseline reaches 4.1
Base+Novel and 1.9 Novel mR@50 against our 14.9 and 7.2, alongside a
Base+Novel micro recall differing by two points.

\paragraph{Holding out the novel predicates}
Our released tower has seen all 50 predicate strings of VG150, since they lie
within its training vocabulary, so its rows do not satisfy the protocol on the
novel half and are marked accordingly. We therefore retrained the tower with
the 15 novel predicates and their synonym groups removed from the training
mixture, the analogue of the rare-class deletion required by OV-LVIS
\citep{vild} (\cref{app:ovr}). This model reaches 22.4 Base+Novel and 11.8
Novel R@50, which places it ahead of the Swin-T entries without MegaSG
pre-training on Base+Novel, behind the MegaSG-pretrained and Swin-B entries,
and behind every published row but one on Novel recall. We make no claim to
lead on those columns. The gap to the full-vocabulary rows, $-5.5$ and $-12.3$
R@50, measures how much of novel performance under this protocol consists of
supervision on \pred{on}, \pred{of} and \pred{in}. On the leaderboard's own
micro metric the held-out model is mid-table; on mean recall over the same
predictions it exceeds the baseline by a factor of nearly four. This
disagreement illustrates the effect described in \cref{sec:priors}.

%% file: tables/sixaxes.tex
\begin{tabular}{@{}l >{\raggedright\arraybackslash}p{4.4cm} c c >{\raggedright\arraybackslash}p{3.5cm}@{}}
\toprule
\sffamily\bfseries axis & \sffamily\bfseries measure & \sffamily\bfseries best baseline & \sffamily\bfseries \model{} & \sffamily\bfseries note \\
\midrule
A1 transfer & VG150, F1@50 (mR@50) & 19.6 (15.2) & 36.9 (28.2) & \note{ROBIN; triplet mass 13\%} \\
 & PSG, F1@50 (mR@50) & 27.9 (22.9) & 34.7 (30.6) & \note{ROBIN; triplet mass 11\%} \\
 & IndoorVG, F1@50 (mR@50) & 22.7 (21.5) & 37.8 (29.5) & \note{ROBIN; triplet mass 7\%} \\
 & HICO-DET, F1@50 (mR@50) & 7.9 (4.5) & 18.7 (12.7) & \note{OvSGTR; zero-shot tower} \\
A2 precision & mean fAP (rare-predicate fAP) & 52.1 (44.6) & 72.6 (70.7) & \note{OvSGTR; pair coverage 89\% vs 100\%} \\
A3 open vocabulary & mR@50 over 19{,}103 strings, VG150 / PSG / IndoorVG & 20.9 / 25.1 / 25.4 & 34.5 / 28.3 / 34.6 & \note{ROBIN; OvSGTR caps at $\approx$150 strings, and ROBIN is scored through a matcher} \\
A4 detector & PSG on a shared YOLO-World, wR@50 (mR@50) & 4.0 (6.2) & 20.0 (20.5) & \note{OvSGTR; pair-recall ceiling 69.6} \\
A5 graph quality & true bits per image (share of annotation) & 13.4 (0.48) & 18.6 (0.67) & \note{OvSGTR; equal length, 199 images; annotation 27.9 bits; verdicts in \cref{sec:results:a5}} \\
A6 spatial & SpatialSense, macro AUC (pooled AUC) & 59.1 (61.7) & 69.0 (67.5) & \note{OvSGTR; chance 50.0; boxes-only baseline 68.8} \\
\midrule
\ourrow OVS composite & harmonic mean of the chance-corrected axes & 11.8 & 40.1 & \note{OvSGTR alone; A1, A2, A4, A5, A6} \\
\bottomrule
\end{tabular}

%% file: tables/a1_transfer.tex
\begin{tabular}{llcccc cccc}
\toprule
& & \multicolumn{4}{c}{\sffamily\bfseries OvSGTR \citep{ovsgtr}} & \multicolumn{4}{c}{\sffamily\bfseries \model{} (ViT-S/16+)} \\
\cmidrule(lr){3-6}\cmidrule(lr){7-10}
\sffamily\bfseries Source & \sffamily\bfseries triplet mass & R@50 & mR@50 & F1@50 & rare & R@50 & mR@50 & F1@50 & rare \\
\midrule
VG150 & 13\% & 39.9 & 10.4 & 16.5 & 0.0 & \best{53.3} & \best{28.2} & \best{36.9} & \best{42.7} \\
PSG & 11\% & 28.7 & 8.8 & 13.5 & 1.7 & \best{40.1} & \best{30.6} & \best{34.7} & \best{23.9} \\
IndoorVG & 7\% & 48.1 & 12.8 & 20.2 & 4.0 & \best{52.7} & \best{29.5} & \best{37.8} & \best{21.6} \\
HICO-DET \scriptsize(zero-shot tower) & --- & 34.0 & 4.5 & 7.9 & 0.2 & \best{35.3} & \best{12.7} & \best{18.7} & \best{4.3} \\
HICO-DET \scriptsize(+5\% HICO-DET train) & --- & 34.0 & 4.5 & 7.9 & 0.2 & \best{45.2} & \best{31.4} & \best{37.1} & \best{23.9} \\
\bottomrule
\end{tabular}

%% file: tables/baselines_psg.tex
\begin{tabular}{llccccrr}
\toprule
\sffamily\bfseries Model & \sffamily\bfseries Setting & \sffamily\bfseries R@20 & \sffamily\bfseries R@50 & \sffamily\bfseries mR@20 & \sffamily\bfseries mR@50 & \sffamily\bfseries pair cov. & \sffamily\bfseries rel/img \\
\midrule
\multicolumn{8}{l}{\itshape PSG test, vocabulary withheld --- one row per matcher, same generations} \\
\model{} & exact & 12.3 & 13.6 & 12.4 & 13.0 & 99.7 & top-$K$ \\
\model{} & A3 & 31.7 & 36.8 & 29.1 & 31.3 & 99.7 & top-$K$ \\
ROBIN-3B & exact & 25.8 & 26.2 & 19.9 & 20.0 & 45.6 & 9.4 \\
ROBIN-3B & SBERT argmax & 30.4 & 35.7 & 21.1 & 22.9 & 77.4 & 34.0 \\
ROBIN-3B & A3 & 35.1 & 37.9 & 24.1 & 25.1 & 65.6 & 18.9 \\
Qwen3-VL-32B & A3 & 15.0 & 15.6 & 5.1 & 5.2 & 35.5 & 13.8 \\
Qwen3-VL-8B & A3 & 10.2 & 10.2 & 3.3 & 3.3 & 23.6 & 9.7 \\
InternVL3.5-8B & A3 & 9.5 & 9.6 & 4.5 & 4.5 & 23.1 & 8.3 \\
GLM-4.6V-Flash & A3 & 8.9 & 9.1 & 2.1 & 2.2 & 23.4 & 11.2 \\
\bottomrule
\end{tabular}

%% file: tables/a2_precision.tex
\begin{tabular}{lccc}
\toprule
\sffamily\bfseries Metric & \sffamily\bfseries OvSGTR \citep{ovsgtr} & \sffamily\bfseries \model{} (ViT-S/16+) & \sffamily\bfseries $\Delta$ \\
\midrule
mean fAP (support $\ge$5) & 52.1 & \best{72.6} & +39.2\% \\
fAP, rare predicates & 44.6 & \best{70.7} & +58.6\% \\
fAP, common predicates & 59.5 & \best{76.0} & +27.8\% \\
mean predicate ROC-AUC (P-AUC) & 83.9 & \best{93.7} & +11.7\% \\
pair coverage (labelled cells scored) & \best{100.0} & 89.0 & -11.0\% \\
\bottomrule
\end{tabular}

%% file: tables/a3_openvocab.tex
\begin{tabular}{lccccc}
\toprule
\sffamily\bfseries Source & \sffamily\bfseries R@50 & \sffamily\bfseries mR@50 & \sffamily\bfseries rare & \sffamily\bfseries MRR & \sffamily\bfseries median rank \\
\midrule
VG150 & 56.0 & 34.5 & 39.6 & 0.68 & 1 / 19,103 \\
PSG & 30.5 & 28.3 & 20.8 & 0.39 & 10 / 19,103 \\
IndoorVG & 53.3 & 34.6 & 34.6 & 0.65 & 1 / 19,103 \\
\bottomrule
\end{tabular}

%% file: tables/a4_deployment.tex
\begin{tabular}{lcccc}
\toprule
\sffamily\bfseries Model & R@50 & mR@50 & rare & wR@50 \\
\midrule
OvSGTR \citep{ovsgtr} \scriptsize(uses detector labels) & 17.7 & 6.2 & 1.2 & 4.0 \\
\ourrow \model{} (ViT-S/16+) & \best{25.4} & \best{20.5} & \best{17.0} & \best{20.0} \\
\midrule
\note{pair-recall ceiling (both models)} & 69.6 & --- & --- & --- \\
\bottomrule
\end{tabular}

%% file: tables/a5b.tex
\begin{tabular}{@{}l ccc c@{}}
\toprule
\sffamily\bfseries model & asserted/img & accepted/img & bits/accepted & \sffamily\bfseries true bits/img \\
\midrule
\multicolumn{5}{l}{\sffamily\itshape matched depth: the top 10 pairs of every system} \\
\ourrow \model{}, vocab = pack & 9.8 & 4.07 & 4.57 & \best{18.6} \\
OvSGTR \citep{ovsgtr} & 9.8 & 4.33 & 3.09 & 13.4 \\
\multicolumn{5}{l}{\note{judge control at this depth: false acceptance 0.201 on specific corruptions ($n=369$), 0.275 pooled ($n=556$)}} \\
\midrule
\multicolumn{5}{l}{\sffamily\itshape deployed depth: each system emits what its own scores justify} \\
\ourrow \model{}, vocab = pack & 18.3 & 6.53 & 4.39 & \best{28.7} \\
OvSGTR \citep{ovsgtr} & 10.3 & 4.64 & 3.04 & 14.1 \\
\multicolumn{5}{l}{\note{judge control at this depth: false acceptance 0.135 on specific corruptions ($n=659$), 0.196 pooled ($n=986$)}} \\
\bottomrule
\end{tabular}

%% file: figures/a5_qual.tex
\input{figures/a5_qual_data}%
\newcommand{\aqidx}[1]{{\color{raMuted}\tiny\sffamily\bfseries #1}}%
\newcommand{\aqbits}[1]{{\color{raMuted}\tiny #1\,b}}%
\newcommand{\aqrel}[6]{\aqidx{#1}\,#2 \textcolor{raAccent}{\textsf{#3}} \aqidx{#4}\,#5\hfill\aqbits{#6}\par}%
\newcommand{\aqon}[6]{\aqidx{#1}\,#2 \textcolor{raMuted}{\textsf{#3}} \aqidx{#4}\,#5\hfill\aqbits{#6}\par}%
\newcommand{\aqhead}[3]{{\sffamily\bfseries\scriptsize #1}\par{\color{raMuted}\tiny #2 relations emitted; #3\,bits if all were true}\par\vspace{1.5pt}{\color{raRule}\hrule height 0.5pt}\vspace{2.5pt}}%
\begin{minipage}{\linewidth}\scriptsize
\noindent\begin{minipage}[t]{0.30\linewidth}\centering\vspace{0pt}
  \includegraphics[height=3.5cm]{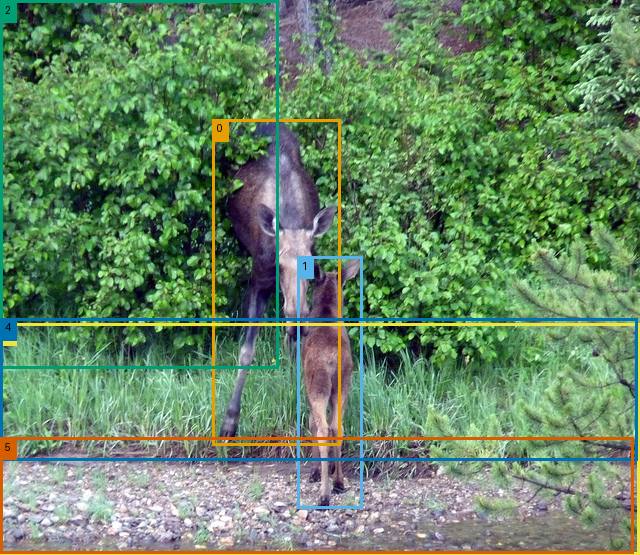}%
\end{minipage}\hfill
\begin{minipage}[t]{0.32\linewidth}\vspace{0pt}\scriptsize\setlength{\parskip}{0pt}
  \aqhead{OvSGTR}{\AqBaseNA}{\AqBaseBitsA}\AqBaseA%
\end{minipage}\hfill
\begin{minipage}[t]{0.33\linewidth}\vspace{0pt}\scriptsize\setlength{\parskip}{0pt}
  \aqhead{\model{}}{\AqOursNA}{\AqOursBitsA}\AqOursA%
\end{minipage}\par\vspace{7pt}
\noindent\begin{minipage}[t]{0.30\linewidth}\centering\vspace{0pt}
  \includegraphics[height=3.5cm]{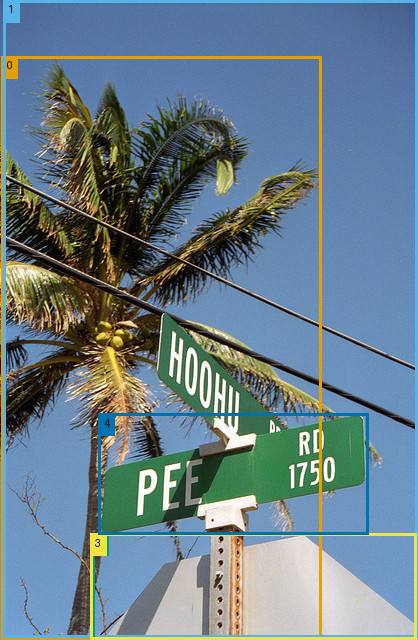}%
\end{minipage}\hfill
\begin{minipage}[t]{0.32\linewidth}\vspace{0pt}\scriptsize\setlength{\parskip}{0pt}
  \aqhead{OvSGTR}{\AqBaseNB}{\AqBaseBitsB}\AqBaseB%
\end{minipage}\hfill
\begin{minipage}[t]{0.33\linewidth}\vspace{0pt}\scriptsize\setlength{\parskip}{0pt}
  \aqhead{\model{}}{\AqOursNB}{\AqOursBitsB}\AqOursB%
\end{minipage}
\end{minipage}

%% file: tables/ovr_leaderboard.tex
\begin{tabular}{ll cccc cc c}
\toprule
& & \multicolumn{2}{c}{\sffamily\bfseries Base+Novel R} & \multicolumn{2}{c}{\sffamily\bfseries Novel R} & \multicolumn{2}{c}{\sffamily\bfseries mean recall @50} & \\
\cmidrule(lr){3-4}\cmidrule(lr){5-6}\cmidrule(lr){7-8}
\sffamily\bfseries Method & \sffamily\bfseries Backbone & @50 & @100 & @50 & @100 & B+N & Novel & \sffamily\bfseries classes hit \\
\midrule
VS3 \citep{vs3} & Swin-T & 15.6 & 17.3 & 0.0 & 0.0 & \note{n/r} & \note{n/r} & \note{n/r} \\
OvSGTR \citep{ovsgtr} & Swin-T & 20.5 & 23.9 & 13.5 & 16.2 & \note{n/r} & \note{n/r} & \note{n/r} \\
RAHP \citep{rahp} & Swin-T & 20.5 & 25.7 & 15.6 & 19.9 & \note{n/r} & \note{n/r} & \note{n/r} \\
OvSGTR + MegaSG \citep{ovsgtr} & Swin-T & 25.4 & 29.7 & 17.0 & 21.1 & \note{n/r} & \note{n/r} & \note{n/r} \\
OvSGTR \citep{ovsgtr} & Swin-B & 22.9 & 26.6 & 16.4 & 19.7 & \note{n/r} & \note{n/r} & \note{n/r} \\
INOVA \citep{inova} & Swin-B & 24.8 & 29.3 & 20.0 & 24.7 & \note{n/r} & \note{n/r} & \note{n/r} \\
\midrule
OvSGTR \citep{ovsgtr} & \note{re-run here} & 20.4 & 23.8 & 13.2 & 15.9 & 4.1 & 1.9 & 13/50 \note{(3/15 novel)} \\
\ourrow \model{} (ViT-S/16+), novel 15 held out & \note{their boxes} & 22.4 & 26.5 & 11.8 & 15.0 & \best{14.9} & \best{7.2} & 40/50 \note{(8/15 novel)} \\
\ourrow \model{} (ViT-S/16+), zero-shot & \note{novel set seen} & 27.3 & 32.0 & 22.8 & 27.0 & \note{n/r} & \note{n/r} & \note{n/r} \\
\ourrow \model{} (ViT-S/16+) & \note{novel set seen} & 27.9 & 32.6 & 24.1 & 28.5 & \note{n/r} & \note{n/r} & \note{n/r} \\
\bottomrule
\end{tabular}

%% file: sec/07_decoupling.tex
\section{Analysis: decoupling object and relation prediction}
\label{sec:decouple}
\label{sec:anyinputs}

The dominant scene-graph architecture predicts objects and relations from a
single backbone and conditions the relation head on predicted object classes
\citep{xu2017,motifs,vctree,ovsgtr}. \model{} does neither. We first measure
what the label input costs the strongest baseline
(\cref{sec:anyinputs:labels}) and show the region source to be interchangeable
(\cref{sec:anyinputs:regions}), then ask whether coupling would have helped: whether relation supervision produces relational features a detector
could share (\cref{sec:decouple:features}), and where in the input the relation logit comes from (\cref{sec:decouple:attrib}).

\subsection{Label dependence of the relation head}
\label{sec:anyinputs:labels}

\begin{table}[t]
\centering\small
\caption{\textbf{The share of each model's output that is a lookup of its object-label
pair.} VG150 test, both models on the \emph{same} boxes and predicted labels
(the baseline's Swin-T detector), top-50 pairs per image. \emph{self-lookup}:
share of predictions reproduced by a table of the model's own most frequent
predicate per label pair, fitted on the other half of images; $H(p \mid
c_s, c_o)$: conditional entropy of the prediction given the label pair (nats); \emph{freq.\ agreement}: share equal to the training majority predicate. Bottom block: labels shuffled or constant.}
\label{tab:classprior}
\resizebox{\linewidth}{!}{\input{tables/classprior}}
\end{table}

\Cref{tab:classprior} measures label dependence on the predictions both models
produce for the leaderboard of \cref{sec:transfer:ovr}. For the baseline,
88--90\% of predictions are reproduced by a table of its own most frequent
predicate per label pair, the label pair explains all but 0.4 nats of its
output entropy, and only 20--31 of the 50 predicates occur. For our model the
lookup reproduces 69--70\%, the residual entropy is three times larger, and
all 50 occur. The lower block reports an intervention: shuffling the labels
leaves the baseline's output statistics unchanged, so its predicate is a
function of the labels rather than the image, whereas our model cannot observe
them at all. Our residual 69\% is the part of relation prediction determined
by category (people wear clothes, cups rest on tables), recovered from pixels
rather than supplied as input.

\subsection{Region sources}
\label{sec:anyinputs:regions}

\begin{figure}[t]
\centering
\includegraphics[width=\linewidth]{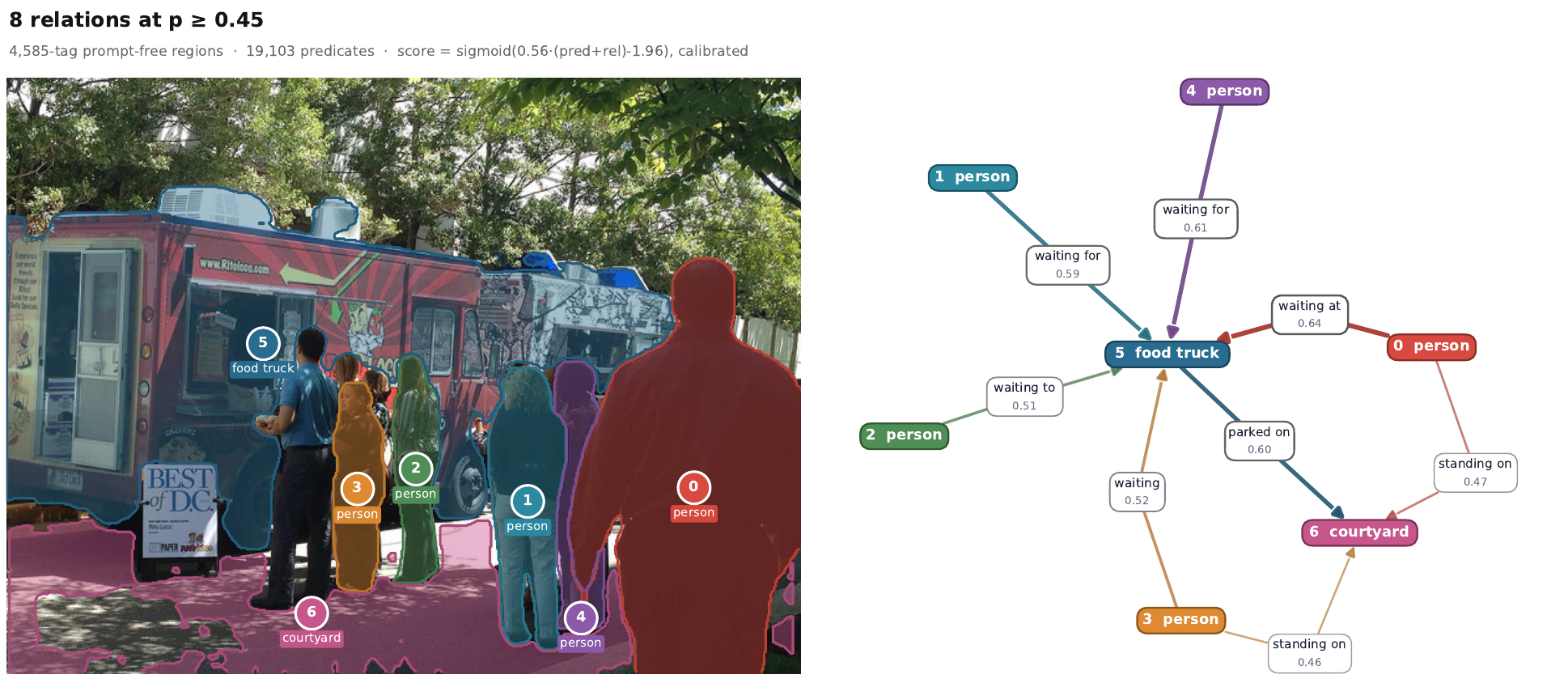}\\[6pt]
\includegraphics[width=\linewidth]{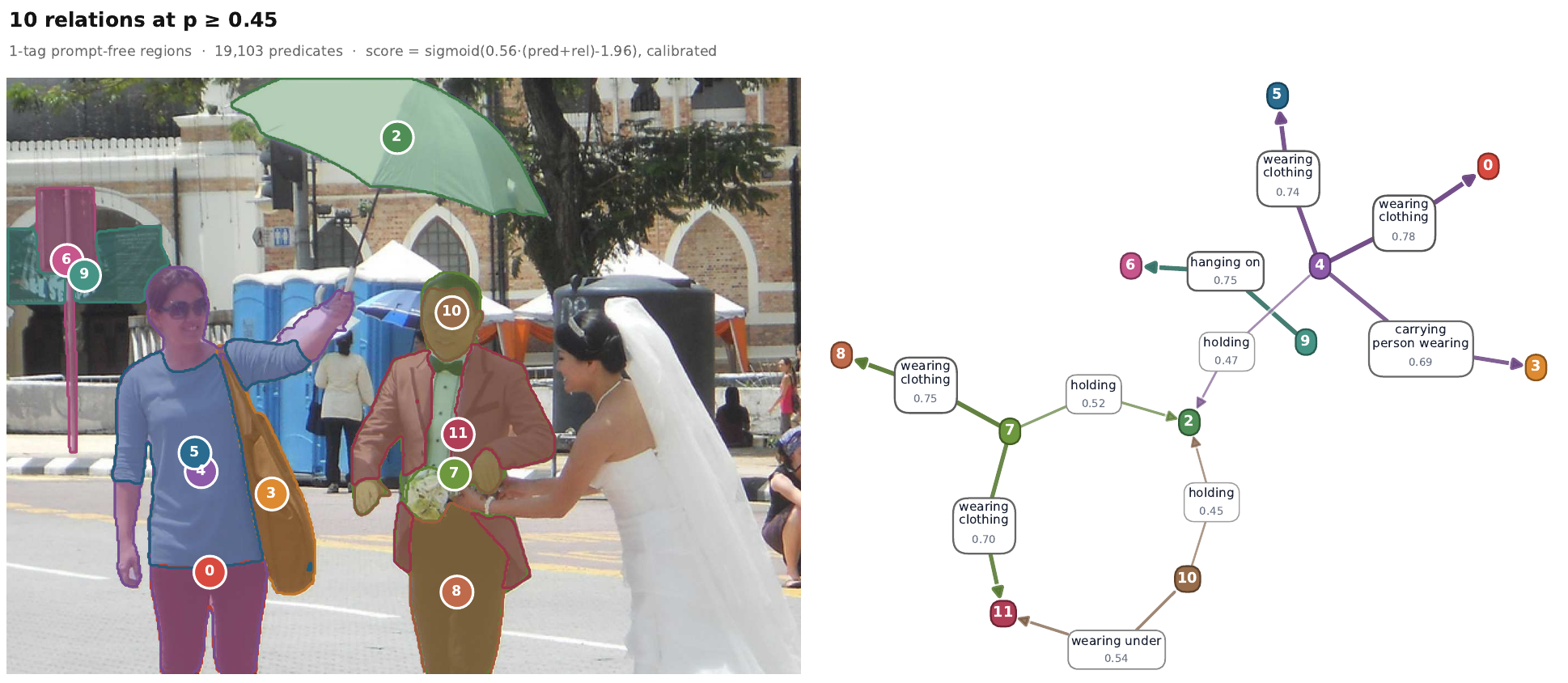}
\caption{\textbf{One relation model behind two unrelated region sources.} Both rows use
one checkpoint at the same calibrated operating point ($p \ge 0.45$, at which
it emits PSG's own density of 6.0 edges per image, \cref{app:threshold}) over
the same 19{,}103-predicate vocabulary; only the region source differs. Top:
YOLOE \citep{yoloe} run prompt-free over 4{,}585 tags returns seven named
segments, from which the model reads eight edges; \emph{food truck} and
\emph{courtyard} are names no scene-graph vocabulary contains. Bottom: FastSAM
\citep{fastsam} has a single class, so its twelve segments carry an index and
no name, and the model still returns \pred{wearing clothing}, \pred{holding}
and \pred{carrying person wearing}. Names in the top row are drawn for the
reader and never reach the model. Both panels use a proxy-scale development
checkpoint; they illustrate the interface rather than the operating quality.}
\label{fig:anyregion}
\end{figure}

Because the model receives regions rather than detections the box source is
interchangeable, and \cref{sec:transfer:deploy} showed every margin of
\cref{tab:a1} surviving a real detector with the same structure; because
labels are never an input, the detector's vocabulary is likewise
unconstrained. \Cref{fig:anyregion} shows the consequence: one checkpoint
reads relations from an open-vocabulary detector prompted with 4{,}585 tags
and from a class-agnostic segmenter supplying no names, and the edges returned
in the second case are of the same kind as in the first (see also
\cref{fig:lvisqual} in \cref{app:results}, on 1{,}198-class LVIS boxes
\citep{lvis,yoloe} passed through unchanged). The failure mode a
region-reading model inherits is visible there too: a detector
returning eleven furniture boxes yields a dense layout graph with few
interactions, since the relation model cannot be more selective than its boxes.

\subsection{Effect of relation supervision on the dense features}
\label{sec:decouple:features}

\begin{figure*}[t]
\centering
\newcommand{\cfpanel}[2]{%
  \begin{minipage}[t]{0.325\linewidth}\centering
    #1\\[2pt]
    {\scriptsize\sffamily\linespread{0.92}\selectfont #2}%
  \end{minipage}}%
\begin{minipage}[t]{0.49\linewidth}\centering
  {\footnotesize\sffamily\bfseries query on the hand (subject)}\\[3pt]
  \cfpanel{\includegraphics[width=\linewidth]{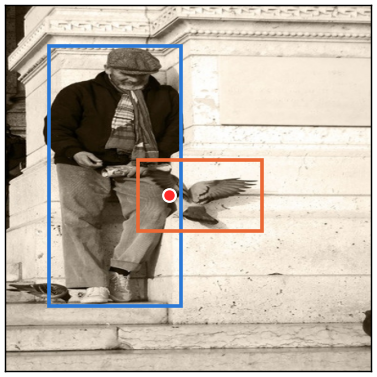}}{image\\and boxes}\hfill
  \cfpanel{\includegraphics[width=\linewidth]{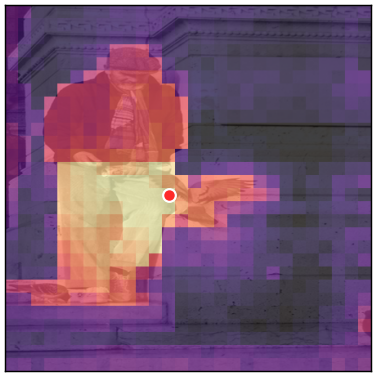}}{DINOv3\\pre-trained}\hfill
  \cfpanel{\includegraphics[width=\linewidth]{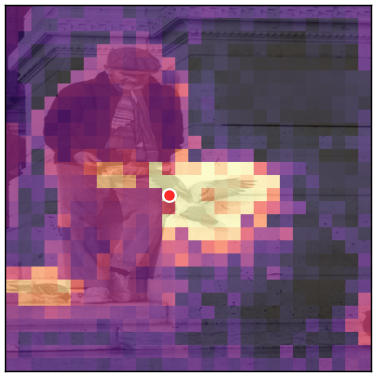}}{DINOv3\\fine-tuned}%
\end{minipage}\hfill
\begin{minipage}[t]{0.49\linewidth}\centering
  {\footnotesize\sffamily\bfseries query on the bird (object)}\\[3pt]
  \cfpanel{\includegraphics[width=\linewidth]{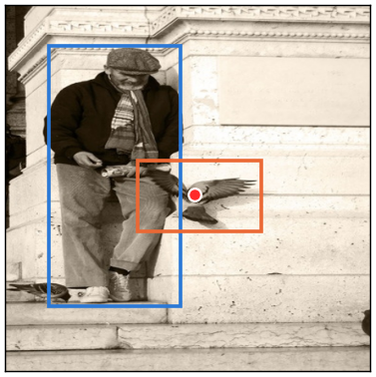}}{image\\and boxes}\hfill
  \cfpanel{\includegraphics[width=\linewidth]{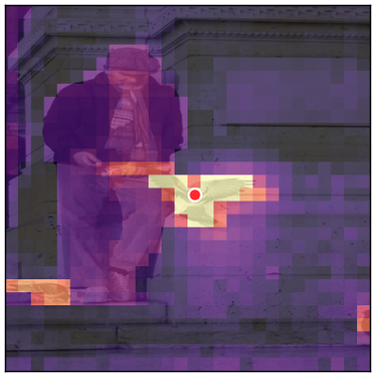}}{DINOv3\\pre-trained}\hfill
  \cfpanel{\includegraphics[width=\linewidth]{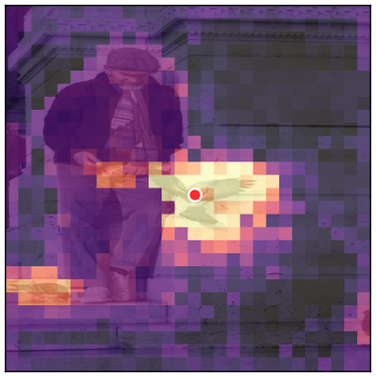}}{DINOv3\\fine-tuned}%
\end{minipage}
\caption{\textbf{The interaction region is an object feature.} Cosine similarity to the
query patch (red dot), last state, per-image mean-centred. Left three panels:
the query is the subject patch nearest the object within their intersection,
the hand. Right three: the query is on the object, the bird. Before fine-tuning the hand query highlights the whole person; after, it highlights hand
and bird, and an interaction region has emerged. Querying the bird yields the
same region, and retrieving the \emph{verb} from it fails (\cref{tab:repr} in
\cref{app:probes}): the region encodes which object is handled, not what is
being done with it.}
\label{fig:contact}
\end{figure*}

For a single backbone to serve both tasks, relation supervision would have to
induce features that encode relations. The probes in \cref{tab:repr}
(\cref{app:probes}) indicate that it does not. The backbone is fully
fine-tuned, and fine-tuning changes it substantially and predictably
(\cref{fig:backbone}): similarity to the pre-trained model falls with depth,
to a linear CKA \citep{cka} of 0.58 at the last state, while two runs under
different recipes stay at a CKA of at least 0.94 with each other, so the
destination is set by the task. What changes is within-object rather than
between-object structure, visible directly in the patch features
(\cref{fig:pca} in \cref{app:arch}).

\paragraph{Patch similarity}
Querying from a subject's patches and measuring cosine similarity across the
image, we find that fine-tuning raises class selectivity from 0.242 to 0.281
and leaves relation selectivity essentially unchanged, from 0.133 to 0.138.
The probability that a relation partner is more similar to the subject than
an unrelated object of the same class decreases from 0.469 to 0.385, against
a chance level of 0.500. Fine-tuning on relations thus made the dense features
less able to distinguish a partner from an arbitrary object of the same
category.

\paragraph{The interaction region}
A weaker form of the hypothesis is that relational structure emerges only
where the objects meet. That region does change: the contact advantage over a
distant subject patch rises from 0.356 to 0.438, and the feature shifts from
covering the whole person to covering hand and object (\cref{fig:contact}).
What it encodes, however, is object identity. Cross-image 1-NN retrieval of
the object class from that feature reaches 0.761 against a chance level of
0.068, whereas retrieval of the verb reaches 0.269 against a majority-class
chance of 0.424, that is, below chance, and fine-tuning does not change this
at any depth (\cref{fig:affordance} in \cref{app:probes}). The finding is
specific to our recipe and this family of objectives (\cref{sec:discussion}),
but it indicates that a backbone trained the way scene-graph backbones
currently are holds no relational representation for a detector to share.

\subsection{Attribution of the relation logit}
\label{sec:decouple:attrib}

\begin{table*}[t]
\centering\small
\caption{\textbf{Input-channel lesion ladder}, zero-shot tower, per-edge top-1 accuracy
over ground-truth pairs, test splits ($n =$ 152.5k / 13.6k / 29.2k edges).
Each row removes one channel and re-runs inference; the top block is absolute
accuracy for the intact model, the rest relative changes. The variance
decomposition is exact in fp32 (maximum residual $2\times10^{-5}$).}
\label{tab:attrib}
\input{tables/attribution}
\end{table*}

Given that the relation is not encoded in the object features, we ask where it
originates. \Cref{tab:attrib} answers by lesion and by exact decomposition,
and the two agree: the object-identity channel contributes almost nothing. The
head contains an optional path by which subject and object features address
the predicate vocabulary directly, gated by a learned scalar initialised at
0.1 (a separate gate from the predicate-conditioned $\alpha_p$ of
\cref{eq:gate}, not lesioned here). Training drives it to 0.014--0.044 across
all three towers, the subject and object terms account for $0.1\%$ and $0.0\%$
of the semantic logit's variance, and removing the path costs $0.1\%$ of micro
accuracy on every source, affecting only macro accuracy on VG150, by $-4.4\%$.
Pair context accounts for nearly all the remainder, 87--93\% of the variance. Scoring against the features of a different image costs 44--68\% of
micro and 85--91\% of macro accuracy, and removing box geometry 6--24\%. The micro--macro gap is the asymmetry of \cref{tab:priors} again:
a prior supplies head predicates; visual evidence contributes in the tail.

\paragraph{Summary}
Relation supervision does not create relational features
(\cref{sec:decouple:features}), the relation logit does not depend on object
identity (\cref{sec:decouple:attrib}), and object labels supply a prior that
inflates the metric by which methods are ranked and does not transfer
(\cref{sec:priors:freq}, \cref{sec:anyinputs:labels}). There is thus no
shared representation that would justify a shared backbone, and there is a
measurable cost to the shared input. The detector can consequently be replaced
without retraining the relation model, and the relation model cannot inherit
the detector's label distribution.

%% file: tables/classprior.tex
\begin{tabular}{lccccc}
\toprule
\sffamily\bfseries Model & self-lookup & $H(p \mid c_s, c_o)$ & $H$ ratio & freq.\ agreement & distinct $p$ \\
\midrule
OvSGTR Swin-T \citep{ovsgtr_eccv} & 87.7 & 0.41 & 0.17 & 65.1 & 20 / 50 \\
OvSGTR Swin-T + MegaSG \citep{ovsgtr} & 89.9 & 0.37 & 0.14 & 74.3 & 31 / 50 \\
\ourrow \model{} (ViT-S/16+), zero-shot & 68.6 & 1.24 & 0.30 & 52.8 & 50 / 50 \\
\ourrow \model{} (ViT-S/16+) & 69.7 & 1.19 & 0.30 & 53.9 & 50 / 50 \\
\midrule
\multicolumn{6}{l}{\sffamily\itshape self-lookup (\%) when the object labels handed to the model are true / shuffled across boxes / a fixed pseudo-label, 5{,}000 images} \\
OvSGTR Swin-T & 85 / 85 / 73 & & & & \\
\ourrow \model{} (ViT-S/16+) & 67 / 67 / 52 & & & & \\
\bottomrule
\end{tabular}

%% file: tables/attribution.tex
\begin{tabular}{lrrrrrr}
\toprule
& \multicolumn{3}{c}{micro Acc@1 (\% $\Delta$ vs.\ full)} & \multicolumn{3}{c}{macro Acc@1 (\% $\Delta$)} \\
\cmidrule(lr){2-4}\cmidrule(lr){5-7}
channel removed & VG150 & PSG & IndoorVG & VG150 & PSG & IndoorVG \\
\midrule
full model & 57.7 & 43.3 & 57.3 & 35.1 & 31.6 & 38.4 \\
\midrule
$-$ object$\to$text channel & -0.1 & -0.0 & -0.1 & -4.4 & -0.2 & +0.6 \\
$-$ box geometry & -16.8 & -23.6 & -6.0 & -36.1 & -33.3 & -29.6 \\
$-$ pixels (shuffled image) & -57.5 & -68.3 & -44.3 & -87.2 & -91.1 & -85.3 \\
$-$ all inputs & -99.3 & -90.5 & -98.5 & -94.3 & -94.4 & -93.0 \\
\midrule
\multicolumn{7}{l}{\emph{exact variance decomposition of the semantic logit:}} \\
pair context & 92.0\% & 87.4\% & 93.1\% & \multicolumn{3}{l}{} \\
subject / object features & 0.1 / 0.0\% & 0.1 / 0.0\% & 0.1 / 0.0\% & \multicolumn{3}{l}{} \\
\bottomrule
\end{tabular}

%% file: sec/08_datalessons.tex
\section{Analysis: in-domain versus cross-dataset measurement}
\label{sec:datalessons}

We evaluate the same models in-domain and under transfer, and test three
quantities that might be expected to predict transfer: alignment of the
training distribution with the target vocabulary, quantity of annotation when
that annotation is sparse, and backbone capacity. Each uses a single-variable
arm read against the 1.3\% noise floor of \cref{app:ablations}. A prior the
model violated that no axis could detect is analysed in
\cref{sec:transfer:wearing}.

\subsection{Corpus scale read in-domain and under transfer}
\label{sec:transfer:negative}
\label{sec:transfer:indomain}

\begin{figure}[t]
\centering
\includegraphics[width=0.74\linewidth]{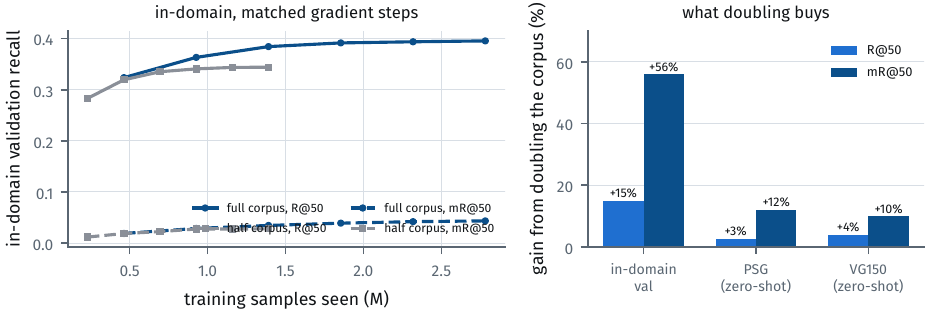}
\caption{\textbf{Doubling the corpus, read in-domain and under transfer.} Left: in-domain validation recall against training samples seen, so the half-corpus arm
is compared at matched gradient steps; the penalty for repetition grows during
training and that arm saturates by its third pass. Right: the relative gain of
the full corpus at the end of training, in-domain and on two zero-shot
benchmarks.}
\label{fig:halfdata}
\end{figure}

Both arms below hold everything else fixed and are evaluated in-domain on the
corpus's own validation split and zero-shot on VG150 and PSG test. Doubling the corpus yields $+15\%$ in-domain micro recall (R@50) and $+56\%$
in-domain macro recall (mR@50); under transfer the same change yields $+2.5$ to $+3.8\%$ and $+9.9$ to
$+11.9\%$ (\cref{fig:halfdata}), so the in-domain slope overstates the
transfer slope by approximately $5\times$. The added images teach the corpus's annotation style, which the in-domain split shares by construction, so in-domain measurement credits a corpus-bound gain.

The second arm changes sign. Removing an automatically derived left/right
excess from the mixture improves the in-domain metrics at every epoch, by
$+2.0\%$ micro and $+27.0\%$ macro, but reduces transfer macro recall by
between $-5.9$ and $-0.3\%$ across the two benchmarks. Two consequences
follow. Decisions on data scaling must be made on transfer measurements; the
extrapolated transfer gain from a further 500k annotated images was about one
point absolute, and we therefore did not generate them. The same applies to
checkpoint selection: ranking arms by their final epoch predicts
out-of-distribution behaviour better than best-epoch selection ($\rho = 0.77$
against $0.66$).

\paragraph{Distribution alignment}
A natural explanation of the mixture result is that transfer tracks the share
of training mass that falls on the target benchmark's predicates. This is not
the case. Dropping the automatically derived edges raises the training mass on
the 56 PSG test predicates from 27.3\% to 39.4\% and reduces left/right from
29.5\% to 7.4\%, as verified on the loaded relations. Transfer macro recall
nonetheless decreased, finishing 3.8 points below the mixture it replaced.

\paragraph{Annotation quantity under sparse annotation}
An Open Images V6 extension \citep{openimages} of 62{,}589 images, audited for
evaluation overlap, was tested at a deliberately amplified 25\% relation share
so that the sign of its effect would exceed the noise floor. Its effect was
negative on every axis relative to the matched arm without it (the
deformable-read arm of \cref{tab:ladder} in \cref{app:ablations}): about
$-6\%$ on the development composite and $-40.2\%$ on the worst cell, HICO-DET
tail recall. Open Images carries 4.33 relations per image against 9.03 for
\dataset{}, so its unannotated pairs enter training as false negatives under
the positive-unlabeled structure of \cref{sec:scale:pu}. Sparse annotation is
therefore worse than none, and the threshold at which this occurs is set by the loss, not the data.

\paragraph{Backbone capacity}
\label{sec:datalessons:capacity}
A matched ladder of three backbones, trained twice, once with the released
recipe and once without the HICO-DET share, is reported in \cref{app:scaling}.
Capacity does not raise the composite: on neither ladder is the composite
monotone in model size, and the largest tower, with $+114\%$ parameters over
the released one, changes it by $-0.1\%$. On both ladders the smallest towers
perform best on the spatial axis. What scaling changes is the source of the
evidence: dependence on box geometry decreases with size while dependence on
pixels increases. The advantage that the largest tower holds on individual
axes does not survive the deployment operating point, at which 5--7 edges are
emitted per image and most of them fall on boxes that no ground-truth object
matches. We therefore release ViT-S/16+ and conclude that the component to
improve next is the detector.

%% file: sec/09_efficiency.tex
\section{Inference cost}
\label{sec:efficiency}

Open-vocabulary scene-graph models are generally reported without their
computational cost. OvSGTR \citep{ovsgtr} scores every pair among about 98
detections at 800 pixels in fp32 at five frames per second on an A40.
We report our cost against that baseline under one protocol and against the
fastest closed-set real-time models we know of on a shared evaluator
(\cref{app:cost}).

\begin{table}[t]
\centering\small
\caption{End-to-end cost against the baseline under one protocol, batch 1, median
latency, eager PyTorch for both. \emph{params} counts the whole system
including the detector; 53.2M of our 231M belong to the relation model
(\cref{tab:scaling}). \note{This compares deployed configurations rather than
architectures: OvSGTR's deformable-attention kernel has no half-precision
implementation, so it is timed at tf32 against our bf16, at its native
800/1333 resize against our 448 px, scoring every pair among 98 boxes where we
sample 128 pairs from 20; the box budget is the largest of the three
differences.}}
\label{tab:costbaseline}
\input{tables/costbaseline}
\end{table}

\begin{table}[t]
\centering\small
\caption{\textbf{Generalist against specialist in the same latency class.} REACT
\citep{react} and REACT++ \citep{reactpp} are closed-set one-stage models
trained on each benchmark; ours is trained once and zero-shot on every row but
the fine-tuned one. Their rows are as published (REACT Tables 2--4, REACT++
Tables 2 and 3, plain non-DCS rows), each benchmark read at the $K$ its source
table averages over, F1@K quoted as printed. Our rows use the standard SGDet
evaluator \citep{sgbench} (graph-constrained, seeds 0--2, maximum seed
$\sigma$ 0.006) on those papers' per-dataset YOLOv8-m weights, averaged over
the same $K$. Bold: best per column and benchmark. \note{Latency measured here
for every row, batch 1 end to end on one A40 in eager mode; dashes mark rows
we did not time.}}
\label{tab:react}
\input{tables/react}
\end{table}

\paragraph{Comparison with OvSGTR}
\Cref{tab:costbaseline} places the two systems under one protocol. Our system
is $7.8\times$ faster end to end on an A40 while having more parameters than
the Swin-T baseline, because batch-1 cost is not dominated by parameter count.
The baseline scores about 9{,}500 pairs among 98 boxes at 800/1333 in fp32,
whereas we score at most 128 sampled pairs among 20 boxes at 448 px in bf16.
Under the same box budget the two would be considerably closer.

\paragraph{Comparison with closed-set real-time models}
\Cref{tab:react} places the released tower alongside the two specialists as
reported in their own papers, on the same per-dataset YOLOv8-m weights and at
the $K$ over which each source table averages. REACT++ is the faster model, at
21--23 ms against 35--36 ms under a protocol that runs our head eager and
unfused, and both specialists reach higher micro recall on every benchmark
they were trained on, while ours is higher on macro everywhere but PSG. The
two combine into an F1@K above REACT on all three, above REACT++ on VG150, 0.2
below it on IndoorVG and 3.7 below it on PSG, for a model that has seen no
annotation from any of them, answers from an arbitrary vocabulary and receives
no object labels. Fine-tuned on the benchmark it takes the highest micro
recall in the table on all three, at a cost in macro recall on VG150 and
IndoorVG. The specialist
recipe, one-stage decoding over a fixed predicate set, is available neither to
an open vocabulary nor to a swapped detector.

\paragraph{Deployment configuration} Three measurements from \cref{app:cost}
determine how the model is run. At batch 1 the relation head is bound by
kernel dispatch rather than arithmetic: the three towers cost 19--20 ms on an
A40 across a $2.5\times$ range of FLOPs. Compilation is the largest single
gain, $1.5$--$1.8\times$ on every GPU, and brings the released tower with its
detector and decoding to 20 ms per frame. Scoring 19{,}103 strings instead of
50 costs under a millisecond at batch 1 and 11--22\% of throughput when
batched. Because the text encoder lies outside the inference graph and the
detector is not part of the model, the same graph runs on a CPU: an fp16 build
reaches 7 frames per second on eight threads with 0.955 top-1 agreement
against fp32 (\cref{tab:targets}).

%% file: tables/costbaseline.tex
\begin{tabular}{lrr ccc r}
\toprule
\sffamily\bfseries System & \sffamily\bfseries params & \sffamily\bfseries boxes/img & \multicolumn{3}{c}{\sffamily\bfseries batch 1, p50 ms} & \sffamily\bfseries FPS \\
\cmidrule(lr){4-6}
& & & A40 & A100 & H100 & A40 \\
\midrule
OvSGTR Swin-T \citep{ovsgtr} & 177M & 98 & 194.0 & 179.9 & 128.1 & 5.1 \\
OvSGTR Swin-B \citep{ovsgtr} & 237M & 98 & 228.5 & 195.1 & 134.3 & 4.3 \\
\ourrow \model{} (ViT-S/16+) + YOLO-World & 231M & 20 & 25.0 & 35.0 & 25.6 & 40.0 \\
\bottomrule
\end{tabular}

%% file: tables/react.tex
\begin{tabular}{llcccr}
\toprule
\sffamily\bfseries Model & \sffamily\bfseries in-domain & R@K & mR@K & F1@K & \sffamily\bfseries ms (A40) \\
\midrule
\multicolumn{6}{l}{\sffamily\itshape VG150 test, YOLOv8-m boxes, 22,371 images; $K = 50/100$} \\
\model{} (ViT-S/16+), zero-shot & no & 26.0 & \best{15.4} & 19.3 & 35.9 \\
\ourrow \model{} (ViT-S/16+) & no & 26.7 & 15.2 & \best{19.4} & 35.9 \\
\model{} (ViT-S/16+), fine-tuned & yes & \best{30.2} & 11.1 & 16.3 & 35.9 \\
REACT \citep{react} & yes & 27.4 & 12.9 & 17.6 & --- \\
REACT++ \citep{reactpp} & yes & 28.9 & 13.2 & 18.2 & 21.3 \\
\midrule
\multicolumn{6}{l}{\sffamily\itshape PSG test, YOLOv8-m boxes, 2,177 images; $K = 20/50/100$} \\
\model{} (ViT-S/16+), zero-shot & no & 25.6 & 22.8 & 24.1 & 35.7 \\
\ourrow \model{} (ViT-S/16+) & no & 26.5 & 23.2 & 24.7 & 35.7 \\
\model{} (ViT-S/16+), fine-tuned & yes & \best{36.4} & \best{24.9} & \best{29.5} & 35.7 \\
REACT \citep{react} & yes & 30.3 & 19.8 & 23.9 & 41.3 \\
REACT++ \citep{reactpp} & yes & 33.6 & 24.7 & 28.4 & 20.9 \\
\midrule
\multicolumn{6}{l}{\sffamily\itshape IndoorVG test, YOLOv8-m boxes, 4,403 images; $K = 20/50/100$} \\
\model{} (ViT-S/16+), zero-shot & no & 25.2 & \best{22.2} & 23.6 & 35.3 \\
\ourrow \model{} (ViT-S/16+) & no & 25.9 & 21.8 & 23.7 & 35.3 \\
\model{} (ViT-S/16+), fine-tuned & yes & \best{32.2} & 18.8 & 23.7 & 35.3 \\
REACT \citep{react} & yes & 30.9 & 18.0 & 22.8 & --- \\
REACT++ \citep{reactpp} & yes & 28.1 & 20.7 & \best{23.9} & 22.6 \\
\bottomrule
\end{tabular}

%% file: sec/10_discussion.tex
\section{Discussion and limitations}
\label{sec:discussion}

\paragraph{Held-out concepts}
Every unmarked result is cross-dataset: the images, object distribution and annotation style of each benchmark are unseen. The predicate strings are not: every benchmark predicate occurs in the training vocabulary.
\Cref{sec:transfer:ovr} evaluates the held-out variant on the split that defines it; the general variant, holding concepts out across all four benchmarks, has not been run, although the architecture permits it
since the predicate matrix is never learned. Several conclusions rest on one seed; a difference within the 1.3\% noise floor of
\cref{app:ablations} is named as such.

\paragraph{Corpus precision}
Every precision claim about \dataset{} is structural: the gate rejects 11.3\%
of candidates, a rejection is a true negative up to box error, and predicates
geometry cannot constrain are named and counted rather than checked
(\cref{sec:corpus:gates}). We have no human-audited precision estimate, a
sample of generated relations read and scored by a person. That is the
measurement a reader should expect before relying on 4.3M machine-generated
annotations, and the one to add first.

\paragraph{Scope of the decoupling result}
\Cref{sec:decouple} shows that our relation supervision does not produce
relational dense features, not that none could: nothing in our recipe provides
a patch-level relational objective, which the probes suggest is what a shared
backbone would require.

\paragraph{The A5 verdicts}
A5 is scored by \cref{eq:truebits}, the information in the relations a
vision--language judge accepts, and our model leads on it. The same judge returns two verdicts favouring the baseline (\cref{sec:results:a5}): shown
whole graphs at deployed length it prefers the baseline's, and shown one
relation at a time it accepts slightly more, which a second judge repeats (\cref{app:oracle}). Per-relation precision is a genuine shortcoming: at any depth the baseline overreaches less, and a deployment that
cannot tolerate a false edge should prefer a shorter graph. Methodologically, a judge cannot perceive repetition, so informativeness must be measured against a reference distribution, not elicited.

\paragraph{Detector recall}
The dominant loss between a benchmark number and a deployed system is
object-detector recall: pair recall is quadratic in object recall and no
relation model can exceed it, while a larger relation model buys accuracy in a
part of the ranking deployment never emits (\cref{sec:transfer:capacity}).

\paragraph{Reporting practices} Each of these is inexpensive and changed a conclusion here: the shared triplet mass between corpus and benchmark, beside every recall number (\cref{sec:priors:overlap}); the matcher and whether it enforces one-to-one assignment; the detector's threshold and box cap, with their pair-recall ceiling
(\cref{sec:protocol}); whether recall is graph-constrained; and one axis on adjudicated negatives (\cref{sec:bench:axes}).

%% file: sec/11_conclusion.tex
\section{Conclusion}
\label{sec:conclusion}

Detection and segmentation became taxonomy-free; relation prediction did not.
We have argued that the obstacles were the supervision, the architecture and
the measurement, and we have removed one of each. \dataset{} supplies 4.3M
relations over 10{,}102 free-text predicates, verified against the geometry of
the boxes they name. \model{} takes regions from any source and a vocabulary
supplied at inference, and on three benchmarks evaluated cross-dataset
and a fourth evaluated zero-shot it exceeds the strongest
open-vocabulary method of comparable scale by $2.3$--$3.5\times$ in mean recall, at 20 ms per frame. \bench{} scores six axes across datasets that no prior can satisfy.

The measurement obstacle is the most consequential, because it is invisible from inside the standard protocol. Scene-graph recall as currently reported is largely a prior-matching score: a pixel-free frequency table wins the leaderboard metric, benchmarks share the
training corpus's annotation style, and scoring conventions move results by
more than the differences between methods. Two further findings follow from measuring across datasets. Coupling object and relation
prediction has no representational basis in our probes, since relation
supervision sharpens object identity without creating relational features and
the relation score is almost entirely determined by pair context. And in-domain measurement overstates transfer gains by about $5\times$ and can favour changes that reduce transfer.

Corpus, model and protocol are public. The cheapest practice we recommend is to report the shared triplet mass between corpus and benchmark beside each recall number: how much of a model's supervision asserts the same relations, of the same kinds of object, as the benchmark.

%% file: app/a_arch.tex
\section{Architecture and training details}
\label{app:arch}

\subsection{Visual path}

The backbone, the feature taps and the pooling are specified in
\cref{sec:model:arch}. Two choices there were settled by ablation
(\cref{app:ablations}). The three taps are layer-normalised individually;
without this the fusion is dominated by the larger activation scale of the
last layer. The backbone is fully fine-tuned at a learning rate $8\times$
below the head's, which at this scale outperforms both a frozen backbone and
LoRA adapters \citep{lora}, although the ordering of those two alternatives
depends on the checkpoint selection rule and inverts under best-epoch
selection.

\begin{figure}[!htb]
\centering
\includegraphics[width=0.72\linewidth]{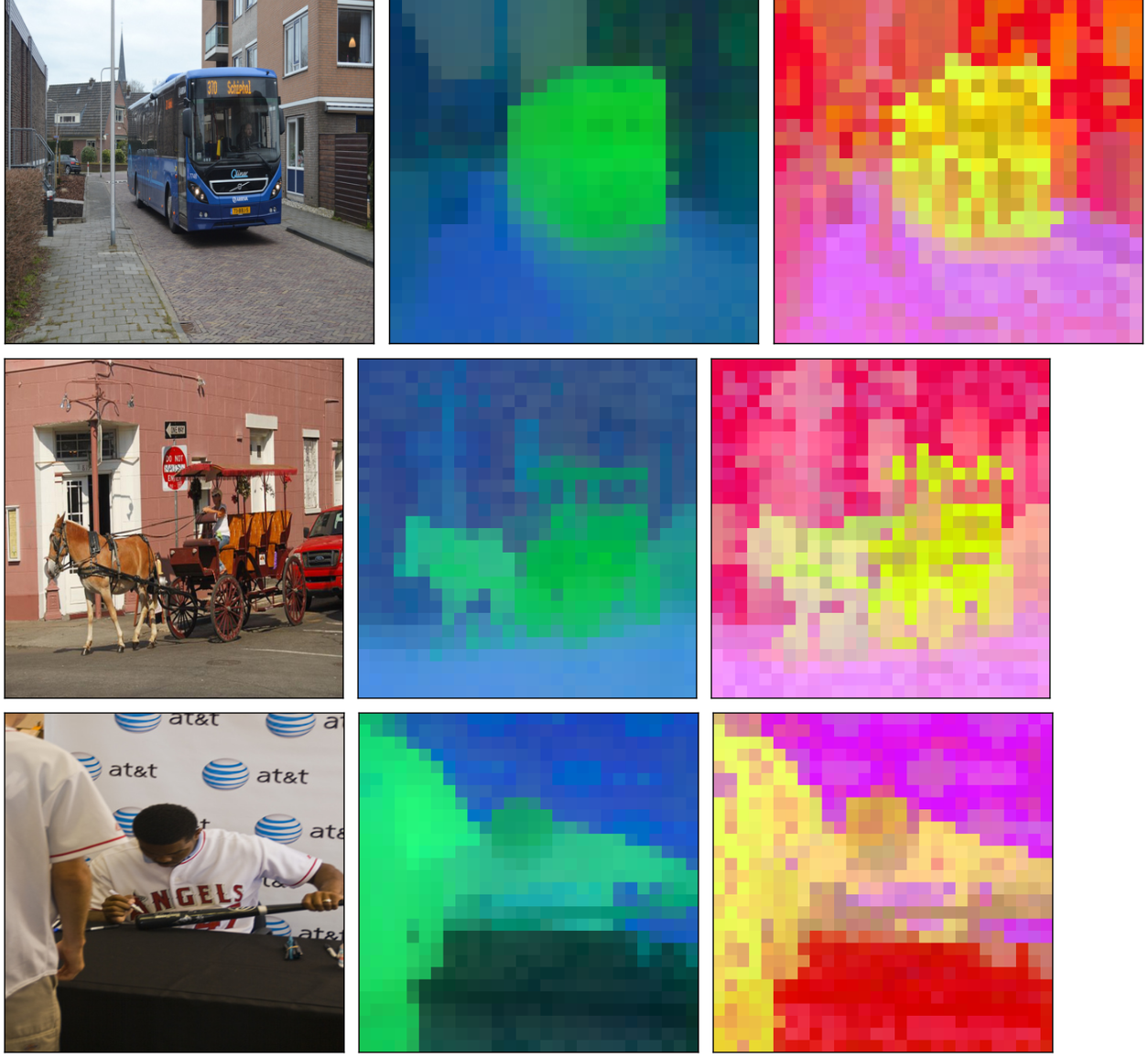}
\caption{Last-state patch features before and after relation fine-tuning,
projected by a single PCA fitted jointly per row (columns: image,
pre-trained, fine-tuned). Fine-tuning sharpens object-versus-background
boundaries and part texture within objects; it does not introduce structure
that spans an object pair. The effect is quantified in \cref{fig:backbone}
(\cref{app:probes}).}
\label{fig:pca}
\end{figure}

\subsection{Deformable scene read}
\label{app:arch:deform}

The read samples 8 heads $\times$ 4 points around each of a pair's four
anchors. Three properties of the stage matter. It is additive behind a
zero-initialised gate, so that training starts from the model without it and
the gate's own gradient decides whether the read is used; a module that
perturbed an already converged system would measure the perturbation rather
than the mechanism. Offsets are unbounded, in units of the anchor's
half-extent, because the anchors are an initialisation prior rather than a
constraint, and clamping would prevent exactly the reads outside the box that
motivate the stage. Points are initialised on rings of distinct angles and
radii per head; a variant that zero-initialised the offset predictor placed
every point of an anchor at one location, where identical gradients cannot
differentiate them.

\begin{figure*}[t]
\centering
\includegraphics[width=\linewidth]{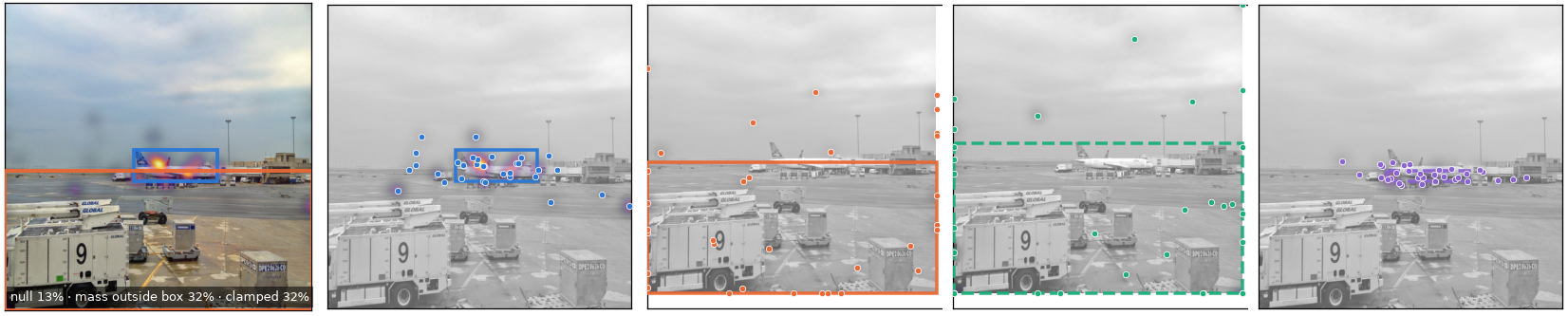}
\caption{\textbf{Sampling locations of the deformable read}, for one PSG-val
pair, \pred{airplane parked on road} at score 0.99. Left: attention mass
splatted at half-patch width. Then, per anchor (subject, object, union,
contact), the sampled points. The subject anchor carries 56\% of the mass and
reads the runway around the aircraft; the object and union anchors sample
mostly outside their boxes and carry 6\% each.}
\label{fig:deformqual}
\end{figure*}

Visualised on 300 PSG and 300 HICO-DET images (\cref{fig:deformqual}), the
trained read is dominated by the subject and contact anchors (PSG shares:
subject 0.43--0.48, contact 0.17--0.21, object 0.12--0.17, union
0.04--0.05). The union anchor contributes least because its box approaches
the whole image, so that 1.5 half-extents falls outside the frame and
42--53\% of its reads are clamped at the border. Direction is carried by the
null slots rather than by geometry: a person$\to$tie \pred{wearing} pair
scored 0.99 places 61\% of its mass on the contact anchor, whereas the same
pair with the roles reversed, scored 0.04, places 52\% on the null slots. The
eight heads weight one location rather than sampling eight distinct ones
(divergence 0.05--0.07).

\subsection{Pair head and predicate-conditioned gate}
\label{app:arch:pairhead}
\label{sec:decouple:arch}

\begin{figure}[t]
\centering
\includegraphics[width=0.85\linewidth]{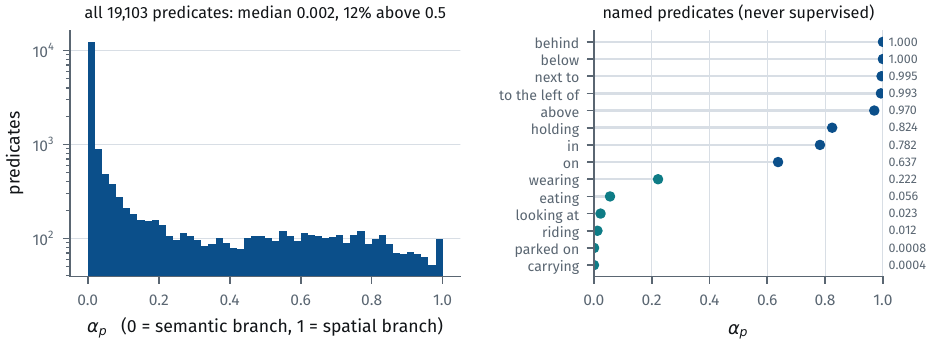}
\caption{\textbf{The predicate-conditioned gate over the vocabulary}, computed
from the text embedding alone. Left: distribution of $\alpha_p$ over all
19{,}103 training strings (log scale). Right: selected predicates. Projective
and proximity relations are routed to the geometry-dominated branch, actions
to the appearance branch, and contact relations in between; none of this is
supervised.}
\label{fig:gate}
\end{figure}

The gate of \cref{eq:gate} receives no supervision, and the weights it learns
over the training vocabulary are bimodal and semantically ordered
(\cref{fig:gate}). The median $\alpha_p$ is 0.002 and 12\% of predicates
exceed 0.5. Projective relations lie at the geometric extreme
(\pred{behind} 1.000, \pred{below} 0.9998, \pred{above} 0.970) and actions at
the other (\pred{carrying} 0.0004, \pred{parked on} 0.0008, \pred{riding}
0.012), with contact predicates between them (\pred{holding} 0.82, \pred{on}
0.64, \pred{wearing} 0.22), as is to be expected of predicates that require
both geometry and appearance.

Because the two branches are scored before they are mixed, one forward pass
yields two graphs: a layout graph over the benchmark's spatial predicates and
a content graph over its semantic ones (\cref{tab:decomposed}). The layout
graph is the easier problem on every source, and the content graph carries
the vocabulary.

\begin{table}[t]
\centering\small
\caption{\textbf{The two-graph decode}: one forward pass scored separately
over the benchmark's spatial and semantic predicates, split by the corpus
type map. Test splits, ground-truth boxes, graph-constrained.}
\label{tab:decomposed}
\input{tables/decomposed}
\end{table}

\subsection{Training recipe}

\begin{table}[h]
\centering\small
\caption{The released training recipe. The run takes 5h04m on four A100
GPUs.}
\label{tab:recipe}
\input{tables/recipe}
\end{table}

Two elements of \cref{tab:recipe} depart from common practice. The resolution
of a batch is drawn from seven values in $[0.5, 1.5] \times 448$; its
mechanism is regularisation rather than resolution (\cref{app:ablations}),
and it is what allows the released model to be evaluated at other resolutions
without retraining (\cref{tab:resolution}). Geometric augmentation is
excluded, because a horizontal flip inverts every left/right relation and the
corpus contains these in balanced quantity, so that flipping without
rewriting the labels would introduce noise on the axis that most requires
supervision. Photometric augmentation, which commutes with geometry, is
retained.

%% file: tables/decomposed.tex
\begin{tabular}{lrcc rcc}
\toprule
& \multicolumn{3}{c}{layout graph (spatial predicates)} & \multicolumn{3}{c}{content graph (semantic predicates)} \\
\cmidrule(lr){2-4}\cmidrule(lr){5-7}
benchmark & \#pred & R@50 & mR@50 & \#pred & R@50 & mR@50 \\
\midrule
VG150 & 7 & 63.1 & 30.6 & 43 & 48.6 & 30.3 \\
PSG & 4 & 60.5 & 53.9 & 52 & 40.5 & 30.3 \\
IndoorVG & 6 & 61.2 & 35.1 & 31 & 42.0 & 30.2 \\
\bottomrule
\end{tabular}

%% file: tables/recipe.tex
\begin{tabular}{ll}
\toprule
\sffamily\bfseries Setting & \sffamily\bfseries Value \\
\midrule
backbone & DINOv3 ViT-S/16+ (LVD-1689M) \\
input resolution & 448, multi-scale 0.5,1.5 over 7 rungs \\
feature taps & $-6,-3,-1$, layer-normalised \\
model width & $d=512$ \\
backbone adaptation & full fine-tune \\
deformable read & 4 points, 8 heads, 2 null slots \\
pair budget & 400 by geometry $\rightarrow$ 128 learned \\
epochs & 12 \\
batch size & 32 per GPU $\times$ 4 GPUs \\
learning rate & head $ 4\times 10^{-4} $, backbone $ 5\times 10^{-5} $ \\
dropout / augment & 0.2 / photometric 0.3, no geometric \\
loss weights & InfoNCE 0.5 $\cdot$ sigmoid 0.25 $\cdot$ swap 0.5 $\cdot$ background 0.05 $\cdot$ grounding 0.1 \\
sampler loss weights & geometry pre-scorer 1.0 $\cdot$ relatedness 1.0 \\
negative discount & estimated $\hat{P}(q \mid p)$, \cref{eq:softneg} \\
positive aggregation & weighted mean over the synonym group \\
InfoNCE negatives & 512 per anchor \\
source-aware negatives & HICO-DET train \\
EMA & 0.9998 \\
mixture & \dataset{} 72.7\%, raw Visual Genome 6.3\%, HICO-DET train 21.0\% by image \\
\bottomrule
\end{tabular}

%% file: app/b_objective.tex
\section{Training objective}
\label{app:objective}

\begin{table}[h]
\centering\small
\caption{Notation used in \cref{sec:model,sec:scale} and in this appendix.}
\label{tab:notation}
\begin{tabularx}{\linewidth}{@{}l>{\raggedright\arraybackslash}X@{}}
\toprule
symbol & meaning \\
\midrule
$I$, $B=\{b_i\}$ & the image, and the regions supplied with it \\
$\mathcal{V}$, $p$ & the predicate vocabulary supplied at inference, and one predicate in it \\
$v_i$ & pooled visual feature of region $i$ \\
$v_{\cup}$, $v_{\cap}$ & pooled features of the union box and of the contact region \\
$g_{ij}$ & 19 scale-invariant geometry features of the ordered pair $(i,j)$ \\
$x_{ij}$ & the 512-d pair representation assembled from the five above \\
$z^{\mathrm{spa}}_{ij}$, $z^{\mathrm{sem}}_{ij}$ & the pair's geometry-branch and appearance-branch embeddings \\
$e_p$ & text embedding of predicate $p$, from the frozen encoder \\
$\alpha_p = g(e_p)$ & per-predicate gate mixing the two branches, \cref{eq:gate} \\
$\tau$, $\beta$ & learned scale and bias on the predicate logit \\
$\ell_{\mathrm{pred}}$, $\ell_{\mathrm{pair}}$ & the predicate logit and the pair-relatedness logit \\
$c_s$, $c_o$ & subject and object \emph{categories}, used only in the sampler and grounding losses \\
$\hat{P}(q \mid p)$ & estimated probability that $q$ also holds when $p$ is annotated, \cref{eq:softneg} \\
\bottomrule
\end{tabularx}
\end{table}

\subsection{Auxiliary objectives and source-aware negatives}
\label{app:aux}

Four terms accompany the InfoNCE objective of \cref{sec:scale}. None is novel
in isolation. A per-cell \emph{sigmoid auxiliary} treats each (pair,
predicate) cell as an independent binary problem, with the negative cells of
a pair weighted so that their total mass equals that of its positives;
together with the
deformable read it is what moved the projective spatial band
(\cref{sec:transfer:spatial}), and it is what makes the scale of the output
head trainable. A \emph{swap hinge}
\begin{equation}
\mathcal{L}_{\mathrm{swap}} = \max\big(0,\; m - c(i,j,p) + c(j,i,p)\big)
\end{equation}
on directional predicates, where $c$ is the mixed cosine inside \cref{eq:gate}
and $m = 0.05$, makes the subject--object order contribute to the loss; each
predicate's term is weighted by its estimated probability of being
directional (\cref{sec:scale:pu}). \emph{Background suppression} lowers the top-$k$ scores of unannotated
pairs, at a small weight because an unannotated pair is not necessarily
negative; a gated variant of this term is the traced cause of the prior
violation of \cref{sec:transfer:wearing}. A light \emph{object--text
grounding} term aligns pooled object features with category text embeddings,
which is supervision on the features rather than an input. The two stages of
the pair sampler carry binary losses of their own, a plain one on the
geometry stage and a focal one on the relatedness stage; all weights are
listed in \cref{tab:recipe}.

\paragraph{Source-aware negatives}
Mixing corpora of different annotation scope makes the positive-unlabeled
problem source-specific. HICO-DET annotates verbs and never spatial
relations, so a pair annotated \pred{holding} in that source is not evidence
against \pred{on}; InfoNCE nevertheless treats it as such, and the measured
effect is a degradation of the spatial axis. Restricting the negatives to
predicates that the source could have annotated, by a per-source mask applied
on top of \cref{eq:softneg}, recovers $+12\%$ on that axis, which is 39\% of
what the HICO-DET share costs there, and gives the best composite of the
family. It is used in the released model and not in the zero-shot one
(\cref{sec:corpus:mixture}).

\subsection{Predicate text space and the synonym penalty}
\label{app:textspace}

\begin{table}[h]
\centering\small
\caption{\textbf{Predicate text-space geometry before and after
distillation}, over the 19{,}103-string training vocabulary. \emph{syn} and
\emph{inv} are mean cosine similarities over 1{,}967 synonym and 770 inverse
pairs; AUC separates the two populations; NN$_1$ is the share of predicates
whose nearest neighbour lies in their own synonym group; effective dimension
is the participation ratio; hubness is the skew of the 10-NN in-degree
distribution.}
\label{tab:textspace}
\resizebox{\linewidth}{!}{\input{tables/textspace}}
\end{table}

\subsubsection{The distillation objective}

The student of \cref{sec:scale:text} is trained against the frozen teacher
under six terms. \emph{Relational distillation} matches the student's
pairwise-cosine matrix to the teacher's over each minibatch; being defined on
cosines it is independent of the two dimensionalities, and it is what the
downstream synonym and inverse masks read. An \emph{absolute anchor}, at a
small weight, regresses the student onto a principal-component projection of
the teacher target and fixes the global frame. \emph{Neighbourhood
preservation} is a row-wise KL divergence between the teacher's and the
student's softmax over batch similarities. \emph{Antonym repulsion} is a
hinge pushing the cosine of a known inverse pair below a margin, and
\emph{synonym cohesion} a hinge holding canonical-group pairs above one; a
softer hinge at a lower margin holds modified spatial forms such as \pred{far
above} near their base predicate without merging them. The
teacher targets are de-hubbed before distillation by removing their three
leading principal directions \citep{abtt}, without which every off-diagonal
cosine sits near 0.80 and the relational term is dominated by that bulk.

Two of the six terms carry the result (\cref{tab:textspace}). Antonym
repulsion is the term a frozen teacher cannot supply, since the teacher
places \pred{to the left of} and \pred{to the right of} at 0.99 and does not
separate inverses from synonyms at all (0.92 against 0.96). Neighbourhood
preservation is what prevents the student from buying that separation
cheaply: without it, a student reaches an equally good synonym-versus-inverse
AUC (0.980 against 0.989) but collapses to an effective dimension of 24
rather than 44, and its nearest neighbour lies in its own synonym group for
18\% of predicates rather than 29\%. The shipped student places inverses near
orthogonal (0.09) while holding synonyms together at 0.71.

\subsubsection{Alignment to a synonym group}
\label{sec:scale:region}

The objective of \cref{sec:scale} aligns a pair embedding to a region of the
text space, the synonym group of the annotated predicate, rather than to a
single point. The cost is that the model frequently answers with a member of
the correct group other than the annotated one, which exact-string metrics
count as a miss. Training against the group as a whole pushes the non-winning
surface forms to ranks 90--338, and the surviving form is not necessarily the
annotated one.

\begin{figure}[t]
\centering
\includegraphics[width=0.85\linewidth]{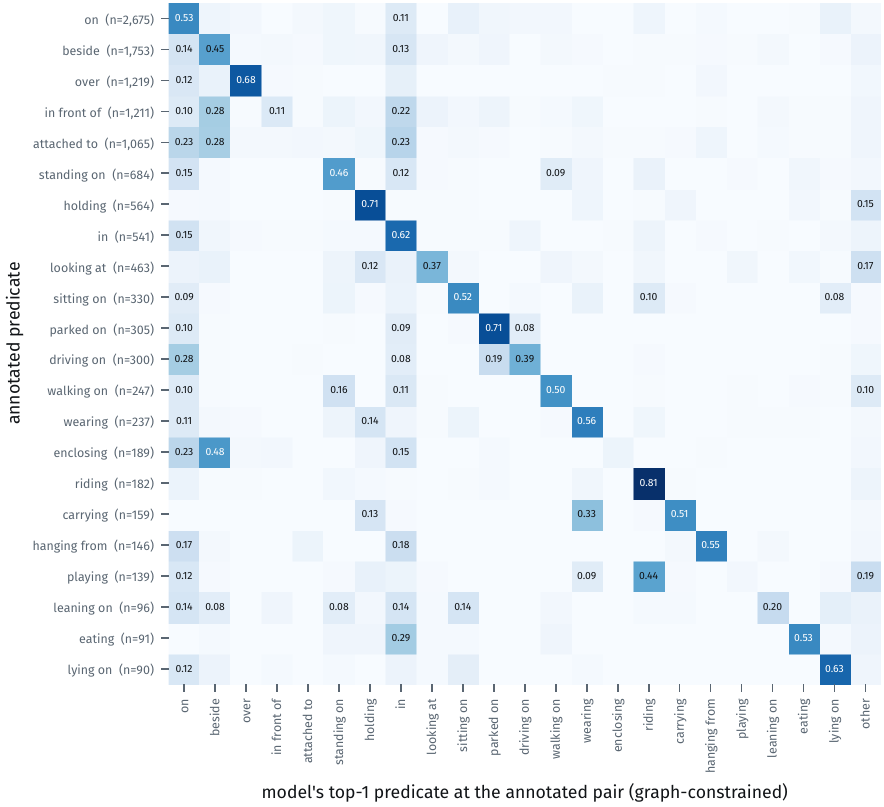}
\caption{\textbf{Confusion of exact-string top-1 predictions on PSG test}, for
the 22 most frequent annotated predicates: the model's graph-constrained
top-1 prediction at every annotated pair, with rows normalised. The
off-diagonal mass is not random: \pred{in front of} is predicted as
\pred{beside} (0.28) and \pred{in} (0.22), \pred{enclosing} as \pred{beside}
(0.48), \pred{attached to} as \pred{beside} (0.28) and \pred{on} (0.23), and
\pred{carrying} as \pred{wearing} (0.33). These are neighbours or entailments
of the annotated string, and are scored as misses.}
\label{fig:confusion}
\end{figure}

The failure is easily misdiagnosed as a directional one. \Cref{fig:confusion}
shows the substitutions on PSG. Directional strings are the clearest case:
the model answers \pred{on the right of} where an annotation reads \pred{to
the right of}, which exact-string recall counts as a miss, whereas measuring
direction directly by subject--object swap places the model's directional
accuracy at 0.97. Every closed-vocabulary number in this paper uses exact matching
and is therefore conservative for our model in a way that it is not for
closed-vocabulary baselines; A3 gives the corrected view.

%% file: tables/textspace.tex
\begin{tabular}{lrcccccc}
\toprule
encoder & dim & syn cos & inv cos & AUC$_{\mathrm{syn/inv}}$ & NN$_1$ in group & eff.\ dim & hubness \\
\midrule
dino.txt teacher & 2048 & 0.96 & 0.92 & 0.854 & 0.36 & 87 & 8.1 \\
\quad + mean removal / ABTT & 2048 & 0.49 & 0.12 & 0.828 & 0.41 & 142 & 0.9 \\
student, 768-d, no neighbourhood term & 768 & 0.93 & 0.50 & 0.980 & 0.18 & 24 & 1.5 \\
\ourrow student, 512-d, shipped & 512 & 0.71 & 0.09 & 0.989 & 0.29 & 44 & 1.3 \\
\bottomrule
\end{tabular}

%% file: app/c_corpus.tex
\section{Corpus details}
\label{app:corpus}

\subsection{Source images and generation}

\begin{figure*}[t]
\centering
\includegraphics[width=\linewidth]{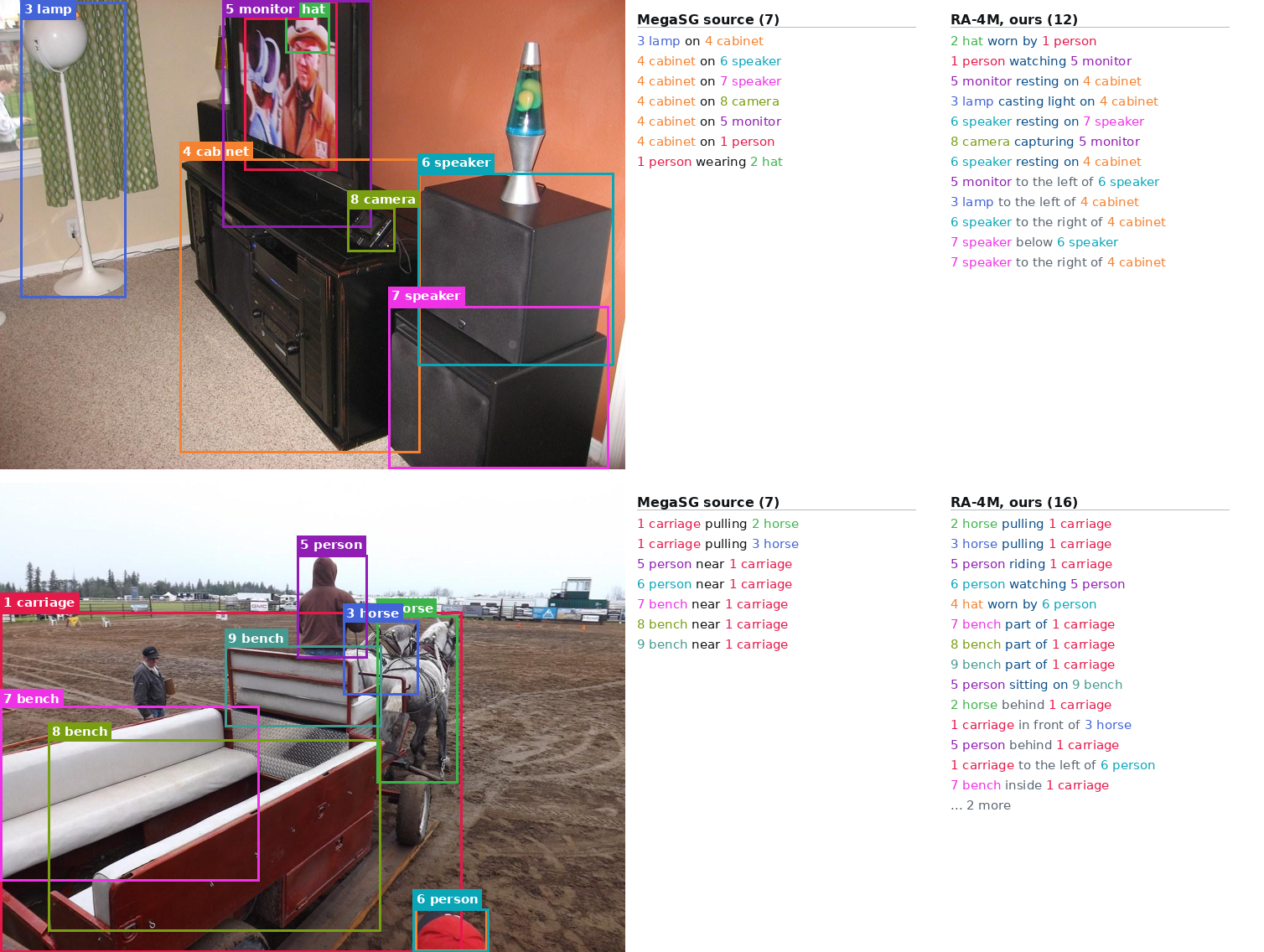}
\caption{The same images and boxes under both annotation layers. Top: the
source layer asserts \pred{cabinet on speaker}, whereas ours reads the
support relations in the correct direction and adds semantics the closed
schema cannot express (\pred{camera capturing monitor}). Bottom: the source
layer contains the inverted \pred{carriage pulling horse}, ours \pred{horse
pulling carriage}, \pred{bench part of carriage} and \pred{person sitting on
bench}. Predicates in blue are semantic-pass outputs, those in grey verified
spatial-pass outputs.}
\label{fig:qual}
\end{figure*}

We annotate the image set of the MegaSG release \citep{ovsgtr}, 500k
COCO-style images drawn from COCO, Objects365 and web sources, taking its
object boxes verbatim and replacing its relation annotations entirely. Fixing
the box set isolates the contribution of relation supervision and lets every
generated graph be joined back to the source annotations for inspection
(\cref{fig:qual}). Of the 499{,}388 images with at least two boxes,
474{,}420 form the training split and 24{,}968 the validation split; seven
training images lose every candidate to the gates, which is why
\cref{tab:corpus} counts 474{,}413.

\begin{figure*}[t]
\centering
\resizebox{\linewidth}{!}{\input{figures/pipeline.tex}}
\caption{The generation pipeline. Three passes of an open-weight
vision--language model propose relations against numbered box markers; a
deterministic layer verifies what geometry can verify, rejects what it
contradicts and passes what it cannot constrain. A salience-ranked geometric top-up fills the
few images that remain below the spatial-density floor.}
\label{fig:pipeline}
\end{figure*}
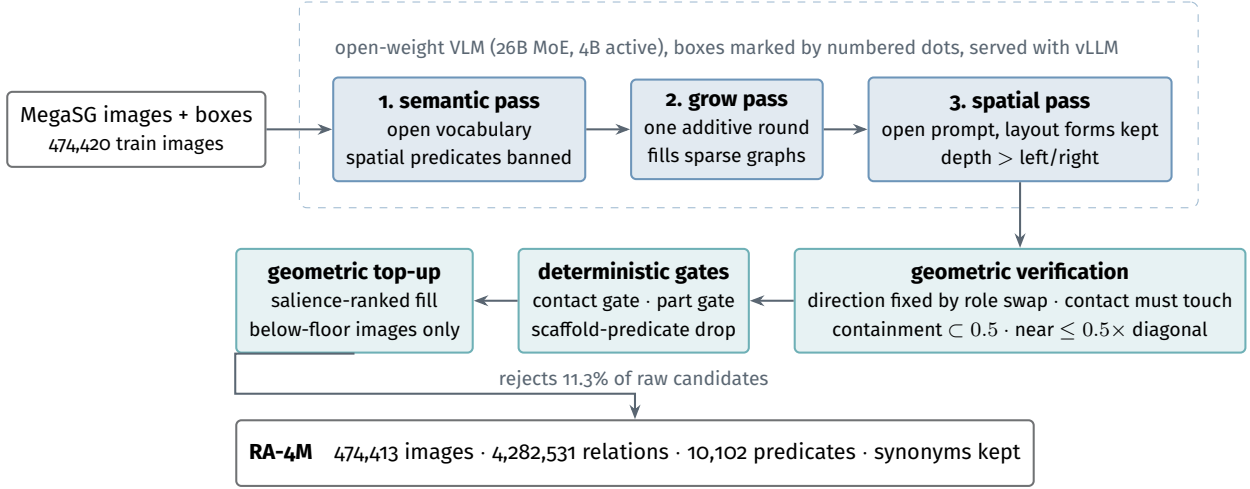

Generation uses an open-weight vision--language model \citep{gemma} (26B
parameters, mixture-of-experts, 4B active per token) served with vLLM
\citep{vllm}, in three passes per image (\cref{fig:pipeline}). The image is
presented with a numbered marker at the centre of each object box, so that
the model refers to instances rather than to categories. The \emph{semantic pass} proposes
actions and interactions with a fully open vocabulary and an explicit
prohibition on spatial predicates; its prompt (\cref{app:datagen}) is written
for coverage rather than precision. The \emph{grow pass} runs one additive
round on the same image to fill sparse graphs; it contributes most of the
corpus's density and most of its errors, and the body-part gate exists
because of it, its characteristic failure being \pred{man wearing human
nose}. The \emph{spatial pass} proposes layout relations under an open prompt
that asks for depth, contact and proximity ahead of the left/right relations
that dominate purely geometric annotation; every proposal is checked against
box geometry, and proposals whose predicate does not reduce to one of 21
layout forms (left/right, above/below, front/behind, \pred{inside},
\pred{near} and thirteen contact phrasings) are dropped, since actions belong
to the semantic passes. Images that remain below a
spatial-density floor receive a salience-ranked geometric top-up, which
contributes 18.4\% of spatial relations.

Two design comparisons contradicted the expected outcome. A
verify-before-assert prompt, which asks the model to justify each relation
before emitting it, was evaluated on a fixed 1{,}000-image calibration sample
and rejected: it raised precision by reducing semantic predicate diversity
from 404 distinct types to 135, whereas the deterministic gates reach the
same precision at no cost in diversity. Conversely, a purely geometric
spatial layer was free but monotone (6 surface forms, 58\% left/right, almost
no depth), whereas propose-then-verify yields 20 surface forms at 32.5\%
depth, 29\% left/right, 14\% proximity and 13\% contact, with left/right and
front/behind balanced to within 0.4\% at full scale.

\subsection{Verification gate}

\begin{figure}[h]
\centering
\includegraphics[width=0.82\linewidth]{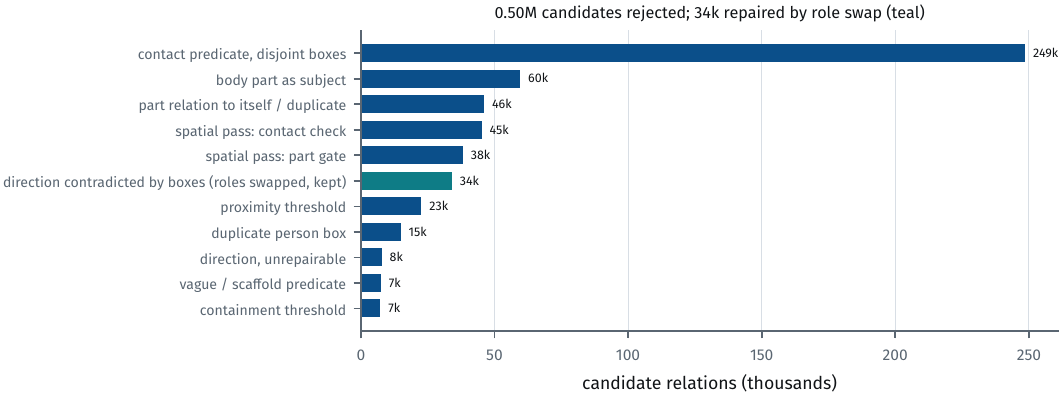}
\caption{\textbf{Candidates rejected by each gate} over the full generation
run. Contact predicates on disjoint boxes dominate; the direction gate is the
one that repairs rather than rejects.}
\label{fig:gates}
\end{figure}

\begin{algorithm}[h]
\small
\DontPrintSemicolon
\KwIn{relation $(s, p, o)$ with boxes $b_s, b_o$, from the spatial pass or a semantic pass; canonical map $c(\cdot)$}
\KwOut{\textsc{keep}, \textsc{keep with roles swapped}, or \textsc{reject}}
\lIf{$s$ and $o$ are both person boxes with $\mathrm{IoU}(b_s, b_o) \ge 0.7$}{\Return \textsc{reject}}
\eIf{the relation is from the spatial pass}{
  \lIf{$s$ or $o$ is a body part}{\Return \textsc{reject}}
  \lIf{$c(p)$ is not one of the 21 layout forms}{\Return \textsc{reject}}
  \If{$c(p)$ requires contact \textnormal{(\pred{on}, \pred{sitting on}, \pred{hanging from}, \ldots)}}{
    \lIf{$b_s$ and $b_o$ do not touch}{\Return \textsc{reject}}
  }
  \If{$c(p)$ is containment}{
    \lIf{$|b_s \cap b_o| < 0.5\,|b_s|$}{\Return \textsc{reject}}
  }
  \If{$c(p)$ is proximity \textnormal{and} $b_s$, $b_o$ do not touch}{
    \lIf{$\mathrm{gap}(b_s, b_o) > 0.5 \max(\mathrm{diag}(b_s), \mathrm{diag}(b_o))$}{\Return \textsc{reject}}
  }
  \If{$c(p)$ is left/right or above/below \textnormal{and} the box centres contradict $p$}{
    \lIf{$p$ has an inverse form}{\Return \textsc{keep with roles swapped}}
    \Return \textsc{reject} \textnormal{(support phrasings such as \pred{atop})}\;
  }
}{
  \If{$c(p) \ne$ \pred{part of}}{
    \lIf{$s$ is a body part}{\Return \textsc{reject}}
    \lIf{$o$ is a body part with at least 60\% of its area inside $b_s$}{\Return \textsc{reject}}
  }
  \If{$c(p)$ requires contact or containment \textnormal{(\pred{wearing}, \pred{holding}, \pred{sitting on}, \ldots)}}{
    \lIf{$b_s \cap b_o = \emptyset$}{\Return \textsc{reject}}
  }
}
\Return \textsc{keep}
\caption{Geometric verification of one generated relation. The canonical map
$c(\cdot)$ groups surface forms for gating only; the emitted string is the
annotator's own. After verification, kept left/right, above/below and
front/behind relations are restated from the other endpoint with probability
one half (\cref{sec:corpus:gates}).}
\label{alg:verify}
\end{algorithm}

\subsection{Statistics and comparison}

\begin{table}[h]
\centering\small
\caption{\textbf{Shared triplet mass, and the mixture over which it is
computed}: the share of a training source's relation \emph{instances} whose
$\langle$subject category, predicate, object category$\rangle$ triple the
benchmark also annotates. The two share
columns differ because the sources differ in density, a share by image being
multiplied by the source's relations per image; HICO-DET's 21.0\% of images
is 5.0\% of relations at 1.94 relations per image against 9.02 for the
leakage-filtered \dataset{}, and the relation share is what the loss observes.
\emph{\#pred} counts the distinct predicate strings among the relations loaded
for training after the leakage filter and is therefore below the corpus totals
of \cref{tab:corpus}. The 19{,}103 on the
mixture rows is the deployed bank, which is not the union of these three
sources but the wider answer space described below, and one bank serves the
whole tower family. No image overlap is involved; leakage is
handled separately (\cref{sec:corpus:leak}). VG150 train, the baseline's
fine-tuning corpus, is the comparison row of \cref{tab:massdecomp}.}
\label{tab:overlap}
\resizebox{\linewidth}{!}{\input{tables/overlap}}
\end{table}

\begin{table}[h]
\centering\small
\caption{\textbf{Shared triplet mass split into its two components}, over the
same cells as \cref{tab:overlap}. \emph{pred} is the share of a mixture's
relations whose predicate string the benchmark also uses, \emph{obj} the share
whose two object-category strings both appear in its taxonomy, and
\emph{triple} the share whose whole triple the benchmark annotates. The
components are printed because the triple charges for all three strings at
once, so a low value can mean either annotation-style or taxonomy mismatch.
The baseline row is the clear case of the second: against IndoorVG it keeps
95.7\% of its predicate mass and 12.3\% of its object-pair mass, because
IndoorVG is indoor-only and VG150's relation mass falls on \cat{sky},
\cat{building} and \cat{street}. Against its own benchmark it loses nothing on
either component, which is the asymmetry \cref{sec:priors:overlap} is
about.}
\label{tab:massdecomp}
\resizebox{\linewidth}{!}{\input{tables/massdecomp}}
\end{table}

\begin{table*}[t]
\centering\small
\caption{Machine-generated relation supervision: how existing corpora are
produced, and where \dataset{} differs. \emph{Grounding} is how a generated
relation is attached to specific object instances; \emph{verification} is
whether the annotator's output is checked. Literature rows are quoted from
the cited papers; dashes denote quantities the papers do not state.}
\label{tab:gen}
\resizebox{\linewidth}{!}{\input{tables/gen_approaches}}
\end{table*}

The use of models to write relation annotations is not new (\cref{tab:gen}),
and existing approaches differ chiefly in how a relation is grounded. The
earliest parses image captions with a language scene parser, which yields
triplets that are ungrounded, since the parser cannot determine which of four
chairs the caption refers to, and that inherit caption bias \citep{gpt4sgg}.
RLIPv2 \citep{rlipv2} scales this to 1.9M images by generating captions with
BLIP and assigning the parsed relation texts to region pairs with a learned
tagger, so that grounding becomes a prediction of the same kind the
downstream model is expected to learn and tagger errors are
indistinguishable from supervision. GPT4SGG \citep{gpt4sgg} and MegaSG
\citep{ovsgtr} instead prompt a multimodal model directly, on 113k COCO
images and 1M images respectively, which grounds relations in the image but
accepts the model's output without verification.

\begin{figure*}[t]
\centering
\includegraphics[width=0.88\linewidth]{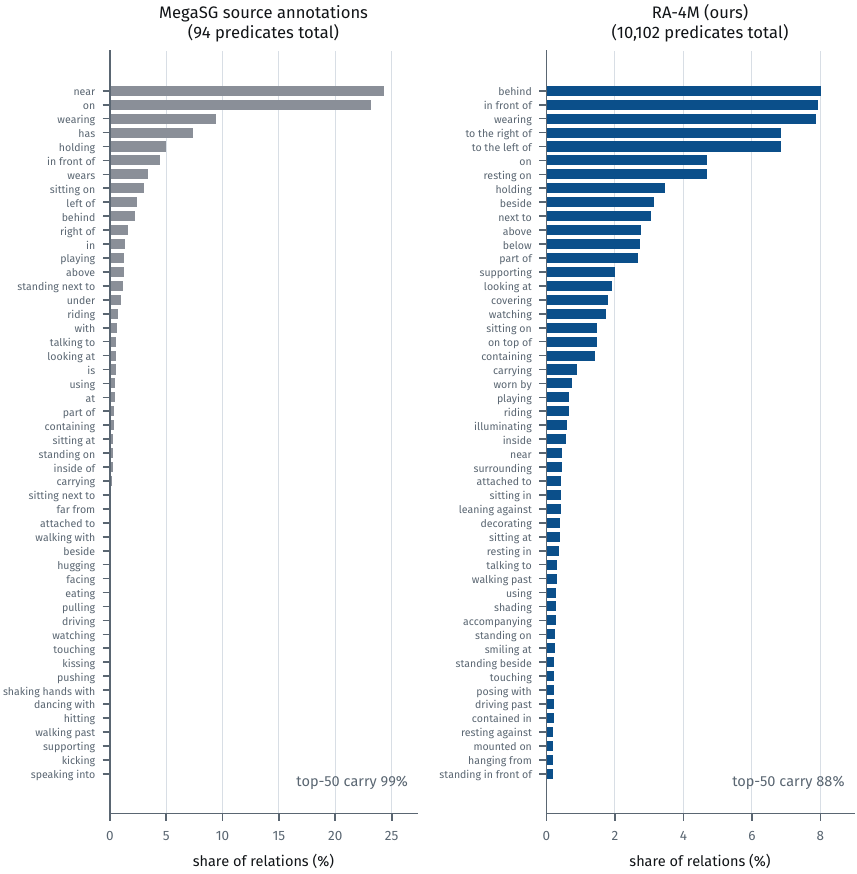}
\caption{Top-50 predicates by relation share, MegaSG source annotations (left)
against \dataset{} (right), on the same images and the same boxes. The head of
the source distribution is \pred{near}/\pred{on} at 24.3\% and 23.2\%; ours is
spatial and direction-balanced (\pred{behind} 8.0\% versus \pred{in front of}
7.9\%, \pred{to the right of} 6.9\% versus \pred{to the left of} 6.8\%), with
no predicate above 8\%. The $x$-scales differ: the difference lies in
concentration as well as in vocabulary size.}
\label{fig:top50}
\end{figure*}

\begin{table}[h]
\centering\small
\caption{The corpus against its source annotations and raw Visual Genome.
Entropy is computed over the predicate distribution, in nats. The MegaSG
release's own annotations are machine-generated (Gemini~1.5 Flash
\citep{ovsgtr}) and projected onto 94 predicate classes.}
\label{tab:corpus}
{\setlength{\tabcolsep}{4.5pt}\input{tables/corpus}}
\end{table}

In rank--frequency, \dataset{} lies between the 94-class closed vocabulary of
its source and the 36{,}549 unconstrained strings of raw Visual Genome: it
has a long free-text tail of 4{,}505 hapax predicates, but an order of
magnitude more mass in ranks 10--100, which is where non-generic but
learnable predicates are found.

\subsection{Design decisions}
\label{app:corpus:decisions}

\paragraph{Surface forms}
\label{sec:corpus:synonyms}
Synonyms are not canonicalised in the emitted corpus, and canonical groups
exist only within the loss (\cref{sec:scale:region}) and the gates. The
reason is that a model whose deployment vocabulary is arbitrary must be
trained on targets that populate the text space densely: collapsing
\pred{grasping} onto \pred{holding} at annotation time teaches the model that
\pred{grasping} does not exist. This is what makes the normalisation of GQA to
roughly 300 predicate classes \citep{gqa}, and the projection of MegaSG onto
94, unusable for our purpose. The cost is the exact-string penalty quantified
in \cref{sec:scale:region}.

\paragraph{Leakage control}
\label{sec:corpus:leak}
Every evaluation split used anywhere in this paper is excluded from training
by image identity. The same photograph exists under two names, since Visual
Genome keys images by VG id and MegaSG by zero-padded COCO id, so a raw
filename intersection between an evaluation split and the corpus returns zero
and is incorrect; the collisions appear only after mapping through an
explicit VG$\leftrightarrow$COCO table of 51{,}498 entries, in both
directions. Removing perceptual near-duplicates of the VG150 and PSG
validation and test images removes 8{,}687 corpus images, and the same
procedure applied to the raw Visual Genome source removes 32{,}456; IndoorVG,
which was added later, contributed a further 506 and 12. All exclusions are
applied by explicit id list.

\paragraph{Second source}
The mixture adds the raw Visual Genome relation annotations, with all of
their free-text predicate strings rather than the 50 of VG150, restricted to
the images that survive the leakage filter: 40{,}615 images, 758{,}547
relations and 17{,}742 predicate strings, of the 36{,}549 raw Visual Genome
carries before filtering. Human free-text annotation has a
different style and a different error profile from machine annotation, and a
small admixture measurably improves transfer. Mixture fractions are applied
per image whereas the loss is computed per relation, so only the relation
share describes the supervision, and the two differ (\cref{tab:overlap}).

\paragraph{Rejected sources, retained vocabularies}
Two corpora were considered as relation sources and rejected. The scene graphs
of GQA are the Visual Genome annotations normalised to roughly 300 predicate
classes \citep{gqa}, which removes the surface-form diversity discussed above,
and raw Visual Genome covers 99.2\% of the GQA images in any case. Synthetic
Visual Genome \citep{svg}, at 49.2 relations per image, was rejected because
half of its object names fall outside the union taxonomy, which invalidates
the per-category-pair annotation statistics the sampler loss depends on
(\cref{sec:scale:pu}).

Their predicate \emph{strings} are kept even so, and this is the distinction
the design turns on. Supervision and answer space are separate: a string may
be a column of the bank without any relation in the corpus asserting it, since
nothing on the text side is learned (\cref{sec:decouple:contract}). The
deployed bank is therefore \dataset{}'s 10{,}102 strings followed by the
9{,}001 further strings that GQA, SpatialSense \citep{spatialsense} and
Synthetic Visual Genome contribute, 19{,}103 in all, which is wider than the
mixture it scores and independent of it.

%% file: figures/pipeline.tex
\begin{tikzpicture}[
  font=\sffamily\footnotesize,
  node distance=4mm and 6mm,
  stage/.style={draw=raRule, fill=raPanel, rounded corners=2pt, thick,
                align=center, inner sep=5pt, minimum height=9mm},
  vlm/.style={stage, fill=raAccent!12, draw=raAccent!60},
  det/.style={stage, fill=raRef!10, draw=raRef!60},
  data/.style={stage, fill=white, draw=raInk!60},
  lbl/.style={font=\sffamily\scriptsize, text=raMuted, align=center},
  arr/.style={-{Stealth[length=2mm]}, thick, draw=raMuted},
  drop/.style={-{Stealth[length=1.8mm]}, thick, draw=BrickRed!70, dashed},
]
\node[data] (img) {MegaSG images + boxes\\ \scriptsize 474{,}420 train images};
\node[vlm, right=9mm of img] (sem) {\textbf{1. semantic pass}\\ \scriptsize open vocabulary\\ \scriptsize spatial predicates banned};
\node[vlm, right=of sem] (grow) {\textbf{2. grow pass}\\ \scriptsize one additive round\\ \scriptsize fills sparse graphs};
\node[vlm, right=of grow] (spa) {\textbf{3. spatial pass}\\ \scriptsize open prompt, layout forms kept\\ \scriptsize depth $>$ left/right};
\node[lbl, above=1.5mm of grow] (vlmlbl) {open-weight VLM (26B MoE, 4B active), boxes marked by numbered dots, served with vLLM};
\begin{scope}[on background layer]
\node[draw=raAccent!40, dashed, rounded corners=3pt, inner sep=3.5mm,
      fit=(sem)(grow)(spa)(vlmlbl)] (vlmbox) {};
\end{scope}
\node[det, below=9mm of spa] (verify) {\textbf{geometric verification}\\ \scriptsize direction fixed by role swap $\cdot$ contact must touch\\ \scriptsize containment $\subset 0.5$ $\cdot$ near $\le 0.5\times$ diagonal};
\node[det, left=of verify] (gates) {\textbf{deterministic gates}\\ \scriptsize contact gate $\cdot$ part gate\\ \scriptsize scaffold-predicate drop};
\node[det, left=of gates] (backstop) {\textbf{geometric top-up}\\ \scriptsize salience-ranked fill\\ \scriptsize below-floor images only};
\node[data, below=9mm of backstop.south west, anchor=north west] (out)
  {\textbf{RA-4M}\quad 474{,}413 images $\cdot$ 4{,}282{,}531 relations $\cdot$ 10{,}102 predicates $\cdot$ synonyms kept};
\draw[arr] (img) -- (sem);
\draw[arr] (sem) -- (grow);
\draw[arr] (grow) -- (spa);
\draw[arr] (spa) -- (verify);
\draw[arr] (verify) -- (gates);
\draw[arr] (gates) -- (backstop);
\draw[arr] (backstop.south) |- (out.west |- backstop.south) -- ++(0,-0.55) -| (out.north);
\node[lbl, below=0.8mm of gates.south] {rejects 11.3\% of raw candidates};
\end{tikzpicture}

%% file: tables/overlap.tex
\begin{tabular}{lrcccccc}
\toprule
& & \multicolumn{2}{c}{\sffamily\bfseries share of mixture} & \multicolumn{4}{c}{\sffamily\bfseries shared triplet mass (\%)} \\
\cmidrule(lr){3-4}\cmidrule(lr){5-8}
\sffamily\bfseries Training source & \sffamily\bfseries \#pred & \sffamily\bfseries by image & \sffamily\bfseries by relation & VG150 & PSG & IndoorVG & Haystack \\
\midrule
\dataset{} & 9,848 & 72.7 & 80.6 & 10.5 & 9.4 & 7.0 & 4.5 \\
raw Visual Genome \citep{vg}, leakage-filtered & 17,352 & 6.3 & 14.4 & 19.0 & 4.0 & 3.6 & 0.3 \\
HICO-DET \citep{hicodet} train & 116 & 21.0 & 5.0 & 31.9 & 50.1 & 8.5 & 24.3 \\
\midrule
\ourrow \textbf{released tower}, the mixture scored everywhere below & 19,103 & 100.0 & 100.0 & 12.8 & 10.7 & 6.6 & 4.9 \\
zero-shot tower \note{no HICO-DET; 91.9/84.6, 8.1/15.4 by image/relation} & 19,103 & 100.0 & 100.0 & 11.8 & 8.6 & 6.5 & 3.9 \\
\bottomrule
\end{tabular}

%% file: tables/massdecomp.tex
\begin{tabular}{l ccc ccc ccc ccc}
\toprule
& \multicolumn{3}{c}{\sffamily\bfseries VG150} & \multicolumn{3}{c}{\sffamily\bfseries PSG} & \multicolumn{3}{c}{\sffamily\bfseries IndoorVG} & \multicolumn{3}{c}{\sffamily\bfseries Haystack} \\
\cmidrule(lr){2-4}\cmidrule(lr){5-7}\cmidrule(lr){8-10}\cmidrule(lr){11-13}
\sffamily\bfseries Training corpus & pred & obj & \bfseries triple & pred & obj & \bfseries triple & pred & obj & \bfseries triple & pred & obj & \bfseries triple \\
\midrule
\ourrow \textbf{released tower} mixture & 53.4 & 44.4 & 12.8 & 38.7 & 36.7 & 10.7 & 49.5 & 26.8 & 6.6 & 38.7 & 30.7 & 4.9 \\
zero-shot tower mixture & 53.0 & 43.6 & 11.8 & 37.2 & 33.9 & 8.6 & 49.4 & 27.4 & 6.5 & 37.2 & 27.6 & 3.9 \\
\midrule
VG150 train \note{the baseline's} & 100.0 & 100.0 & 90.9 & 57.3 & 22.1 & 8.6 & 95.7 & 12.3 & 10.4 & 57.3 & 7.1 & 0.6 \\
\bottomrule
\end{tabular}

%% file: tables/gen_approaches.tex
\begin{tabular}{llllll}
\toprule
\sffamily\bfseries Corpus & \sffamily\bfseries annotator & \sffamily\bfseries images & \sffamily\bfseries predicates & \sffamily\bfseries grounding & \sffamily\bfseries verification \\
\midrule
caption parsing & language parser & --- & --- & none (ungrounded) & none \\
GQA \citep{gqa} & normalised VG & 113k & 310 & human boxes & none \\
RLIPv2 \citep{rlipv2} & BLIP + R-Tagger & $>$1.9M & free text & learned tagger & none \\
GPT4SGG \citep{gpt4sgg} & GPT-4 & 113k & free text & prompted & none \\
MegaSG \citep{ovsgtr} & Gemini 1.5 Flash & 1M & 94 (release) & prompted & none \\
\ourrow \dataset{} (ours) & open 26B VLM, 3 passes & 474k & 10,102 free text & numbered box markers & deterministic geometry \\
\bottomrule
\end{tabular}

%% file: tables/corpus.tex
\begin{tabular}{lrrrrr}
\toprule
\sffamily\bfseries Corpus & \sffamily\bfseries images & \sffamily\bfseries relations & \sffamily\bfseries rel/img & \sffamily\bfseries predicates & \sffamily\bfseries entropy \\
\midrule
\ourrow \dataset{} (this report) & 474,413 & 4,282,531 & 9.03 & 10,102 & 3.99 \\
MegaSG source annotations (same images) & 474,420 & 2,510,905 & 5.29 & 94 & 2.56 \\
raw Visual Genome \citep{vg} & 108,077 & 2,315,906 & 22.10 & 36,549 & 3.81 \\
\quad after our leakage filter (secondary source) & 40,615 & 758,547 & 18.70 & 17,742 & --- \\
\bottomrule
\end{tabular}

%% file: app/d_protocol.tex
\section{Protocol details}
\label{app:protocol}

\subsection{Relation score}
\label{app:score}

The deployed relation score is
\begin{equation}
s = \sigma\!\big(a\,(\ell_{\mathrm{pred}} + w\,\ell_{\mathrm{pair}}) + b\big),
\label{eq:score}
\end{equation}
in which the pair-relatedness logit $\ell_{\mathrm{pair}}$ is fused
additively; one definition is shared by the evaluator and the deployed
pipeline. Additive and multiplicative fusion rank differently. The
annotation-derived benchmarks favour the multiplicative form, because
multiplication up-weights $\ell_{\mathrm{pair}}$, the annotation-propensity
term; on adjudicated negatives the ordering inverts (additive 0.911 AUC,
multiplicative 0.904, relatedness alone 0.748). We report the additive form
throughout.

\subsection{Sensitivity of the composite}

\begin{table}[h]
\centering\small
\caption{\textbf{Leave-one-axis-out analysis of the composite.} OVS is
recomputed with each axis withheld in turn, from the same chance-corrected
per-axis scores that \cref{tab:sixaxes} aggregates.}
\label{tab:ovsloo}
\input{tables/ovs_loo}
\end{table}

\Cref{tab:ovsloo} recomputes the composite with each axis withheld. The
ordering of the two systems does not depend on any single axis, including
A5, whose scoring \cref{sec:results:a5} discusses at greatest length.
Withholding A4 narrows the gap most, because A4 is the axis on which the
baseline is furthest behind.

\subsection{Annotation propensity in the score}
\label{sec:priors:relatedness}

\begin{figure}[h]
\centering
\includegraphics[width=0.85\linewidth]{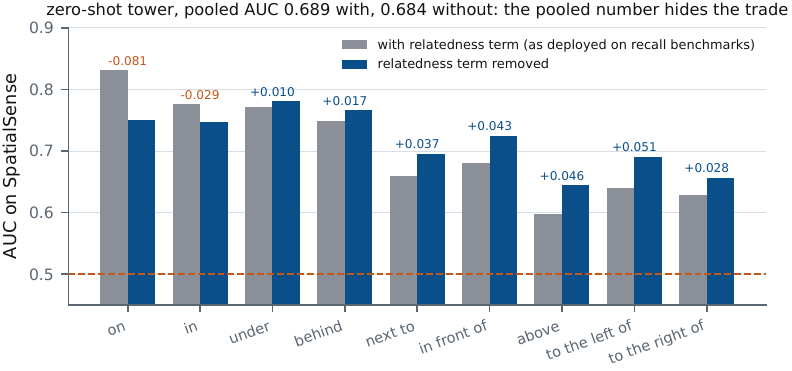}
\caption{\textbf{Effect of the relatedness term on adjudicated spatial
negatives.} Per-predicate AUC on the adjudicated true/false pairs of
SpatialSense with the learned pair-relatedness logit fused into the score, as
on every recall benchmark, and with it removed. The pooled AUC moves by 0.005
and reflects none of this. The probe uses the zero-shot tower, so its pooled
AUC differs from the 0.675 that \cref{sec:transfer:spatial} reports for the
released tower.}
\label{fig:relspatial}
\end{figure}

The pair-relatedness score of \cref{sec:model:arch} is the
annotation-propensity term of \cref{sec:priors:propensity}. It is present by
construction, so its contribution can be measured on both benchmark
families, and its sign differs between them. On the annotation-derived
benchmarks, retaining it improves recall. On adjudicated negatives, removing
it at evaluation time improves the result: every projective predicate gains
between $+0.03$ and $+0.05$ AUC and only the two contact predicates lose,
which is the expected behaviour of a contact prior (\cref{fig:relspatial}).
Recall against sparse annotation and truth judgement against adjudicated
negatives are therefore different quantities, and one term can improve the
first while degrading the second.

\subsection{Calibration}

\begin{figure}[t]
\centering
\includegraphics[width=0.85\linewidth]{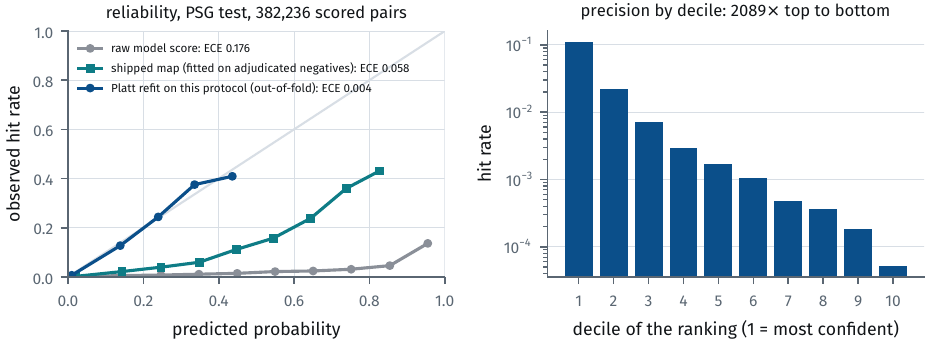}
\caption{\textbf{Calibration of the relation score.} Left: reliability diagram
of the relation score on all 382k scored PSG-test pairs, where a hit is an
annotated triple. The raw score is over-confident (ECE 0.176); two Platt
\citep{platt} parameters fitted out-of-fold under this protocol reduce this
to 0.004; the released map, which is fitted on adjudicated negatives, lies
above the diagonal here because an unannotated true relation counts as a
miss (ECE 0.058). Right: hit rate by decile of the ranking, which spans three
orders of magnitude.}
\label{fig:calibration}
\end{figure}

The model ranks well but on an uncalibrated scale (\cref{fig:calibration}).
Its logit scale and bias are trained against a mass-balanced objective, that
is, an implicit 50/50 prior, whereas a real frame is 0.2--4\% positive, so
scores accumulate near 1 and a threshold has no fixed meaning; the ranking
behind those scores nonetheless separates the top decile from the bottom by
three orders of magnitude in precision. The two parameters $(a, b)$ of
\cref{eq:score} are obtained by a Platt fit \citep{platt}, which is monotone,
so every ranking metric is unchanged and only the meaning of a threshold
changes.

What the fit is made against determines what the number means. Fitted on
annotation hits, the score estimates the probability that a relation appears
in an annotation and reaches an ECE of 0.004 under the same protocol. The
released map is instead fitted on adjudicated negatives and estimates the
probability that a person judges the relation true, which is the relevant
question at deployment; it is conservative in the useful direction, reporting
a precision of 0.77 at the released operating point where the adjudicated
precision is 0.90. Per-predicate calibration was evaluated and rejected, as
it fits in-domain and does not transfer.

\subsection{Output at the operating threshold}
\label{app:threshold}

\begin{table}[h]
\centering\small
\caption{\textbf{Output at three thresholds of the calibrated score.} PSG
test: 2{,}173 images, 13{,}077 annotated relations, 6.02 per image, every
candidate pair scored in the deployment configuration of \cref{app:cost}.
Recall, $P_\text{ann}$ and $P_\text{cal}$ are percentages. $P_\text{ann}$
counts an unannotated prediction as wrong and is therefore a lower bound;
$P_\text{cal}$ is the mean calibrated score over the retained set, which
estimates precision without reference to the annotation. The head is the
mask-gated variant rather than the released one, so the absolute values are
not release figures; what \cref{sec:priors:propensity} uses is the
decomposition below.}
\label{tab:threshold}
\input{tables/threshold}
\end{table}

Because $(a, b)$ are fitted against adjudicated cells rather than against
annotation, a threshold on the resulting score is a probability rather than a
rank cut, and the qualitative figures in this paper specify a threshold
rather than a top-$k$. \Cref{tab:threshold} reports what such a threshold
costs and returns. Three defensible criteria disagree by a factor of three in
output volume: maximising F1 against the calibrated precision emits 18.2
edges per image, maximising it against the annotated precision emits 7.9, and
matching PSG's own annotation density emits 6.1 at $\tau = 0.45$. The last is
the operating point used for the qualitative panels of
\cref{sec:anyinputs:regions}, so those are produced at the corpus's density
rather than at a budget chosen to suit them.

The two precision columns differ by a factor of 2.8 at that threshold, and
the gap can be counted. Of the 39{,}516 edges emitted at the calibrated-F1
threshold, 4{,}376 (11.1\%) match an annotated triple, 4{,}201 (10.6\%) fall
on a pair that PSG annotated and assign a different predicate to it, and
30{,}939 (78.3\%) fall on a pair that PSG never annotated in either
direction. Only the middle group is a predicate error in the ordinary sense;
the largest group is unscored rather than wrong. This is the propensity prior
of \cref{sec:priors:propensity} in counted form, and the reason
\cref{sec:bench:axes} scores one axis against adjudicated negatives. The
predicate errors are more often a vocabulary difference than a mistake: among
the 8{,}577 edges on an annotated pair, 51.0\% carry PSG's own predicate, and
the remainder concentrate on near-synonyms and hypernyms, namely \pred{in}
for an annotated \pred{on} (231 cases), \pred{on} (148) and \pred{beside}
(137) for \pred{attached to}, and \pred{beside} for \pred{in front of} (132).
A recall metric counts every one of these as a miss.

\subsection{Detector operating point}
\label{app:boxbudget}

\begin{table}[h]
\centering\small
\caption{\textbf{Pair-recall ceiling at two detector operating points.} Each
row uses identical detector weights; only the confidence threshold and box
cap differ. \emph{ceiling} is pair recall, the fraction of ground-truth
relations for which both endpoints are detected, which is the upper bound on
relation recall for any model given those boxes. The baseline's
GroundingDINO applies no threshold and retains every query whose label is not
background, which is why it appeared to be the stronger detector.}
\label{tab:boxbudget}
\input{tables/boxbudget}
\end{table}

\Cref{tab:boxbudget} is the table underlying \cref{fig:boxbudget}: identical
detector weights per row, two operating points, and the pair-recall ceiling
each implies.

\subsection{OvR-SGG reproduction}
\label{app:ovr}

\begin{table}[h]
\centering\small
\caption{\textbf{Reproduction of the OvR-SGG leaderboard, and the effect of
the matcher.} VG150 test, SGDet, graph-constrained. Published rows are the
baseline's own; reproduced rows re-run its released checkpoints through our
inference code and score them with a transcription of its metric. The third
row of each block changes only the matcher, by adding the one-to-one
ground-truth-to-detection assignment that object detection has required
since COCO.}
\label{tab:protocol}
\resizebox{\linewidth}{!}{\input{tables/protocol}}
\end{table}

\Cref{tab:protocol} is the reproduction underlying \cref{sec:protocol}: the
baseline's released checkpoints, run through our inference code and scored
with a transcription of its metric, recover its twelve published numbers to
within 0.35 points, and the third row of each block changes only the matcher.

\paragraph{Scope of the leaderboard rows}
Our model has never seen a VG150 relation annotation, so on Base+Novel it
performs a strictly harder task than the published rows. The premise of the
split, however, is that the 15 novel predicates were unseen in any
annotation, and \dataset{} contains those strings. Open-vocabulary detection
establishes where the line is drawn. OV-LVIS \citep{vild} deletes the
annotations of the 337 rare LVIS categories from the training set and
retrains on the remaining 866, so the split is a property of the training set
rather than a column of the results table. What the protocol does not require
is that the novel concepts be unseen everywhere, since the classifier for the
held-out categories is built from an image--text model whose web corpus
contains their names, and Detic \citep{detic} trains on ImageNet-21k
image-level labels for an overlapping vocabulary while removing the
rare-class boxes. Our supervision for the novel predicates is not of the
permitted kind, because it is grounded and instance-level, which corresponds
to retaining those boxes. The compliant version removes those predicates and
their synonym groups from the training mixture and retrains; it is the
\note{novel 15 held out} row of \cref{tab:ovrleaderboard}, and the
full-vocabulary rows below it are marked as non-compliant because the
difference between them measures how much of novel performance under this
protocol is supervision on three frequent strings. Neither depends on a
single box set: the two full-vocabulary variants differ by less than the gap
to the next entry, and the numbers move by less than 0.1 when the boxes of
the baseline's stronger checkpoint are substituted. Scoring every candidate
pair rather than our deployed pair budget, which covers 43\% of them, adds
2.1 Base+Novel and 3.7 Novel R@50.

\paragraph{Predicate split and matcher}
Two incompatible predicate splits are in circulation. We use that of OvSGTR
\citep{ovsgtr_eccv} throughout, with 35 base and 15 novel predicates, because
every row of \cref{tab:ovrleaderboard} uses it and because it is the only one
published as an enumerated list and can therefore be reproduced exactly. PGSG
\citep{pgsg} instead ``randomly select[s] 50\% predicate categories as
novel'' (30\% in its own supplement), names none of them, and reports
novel-class mR@K where the leaderboard reports R@K. Recent work tabulates
both under one ranking with a footnote \citep{vlirm}; we do not, because a
25/25 partition changes the training supervision, the test denominator and
the metric simultaneously, and the resulting columns are not commensurable.
The whole table is scored with the uncorrected matcher, the duplicate-
crediting one that \cref{sec:protocol} argues against, since this is the only
setting in which our rows and the published rows have the same meaning.
Applying the correction to the baseline's rows (\cref{tab:protocol}) reduces
Base+Novel by 11--13\% and Novel by 13--15\% at every $K$, and applying it to
ours as well leaves the ordering unchanged.

\subsection{Graph-quality oracle (A5)}
\label{app:oracle}

A1--A4 score predictions against annotations and inherit their blind spots: a
relation that is true but unlabelled counts as a false positive, and a graph
can be no better than the corpus it is compared against. A5 removes the
ground truth from the question. A vision--language judge is shown the
photograph with its numbered boxes and asked about the relations a system
emits. The judge is Qwen3-VL-8B \citep{qwen3vl}. It is not a Gemma model,
because \dataset{} was written by one and a Gemma judge would favour our own
output distribution, and it is not a reasoning variant, because the verdict
is parsed from the last JSON object within a 300-token budget and a long
chain of thought would exhaust it. The rubric, the prompt and the rendering
are released with the benchmark.

\begin{table}[h]
\centering\small
\caption{\textbf{The whole-graph oracle}, PSG test, the shared YOLO-World
boxes of A4, judge Qwen3-VL-8B. Top: graph shape, which requires neither
ground truth nor a judge. Bottom: the oracle verdicts as our win rate among
decisive comparisons, stratified by which graph was longer. \emph{pack}
denotes our head reparameterised to the benchmark's 56 predicates and
\emph{train} the full vocabulary. The controls certify the judge rather than
any single comparison, and are pooled across runs of one judge
configuration.}
\label{tab:a5}
\resizebox{\linewidth}{!}{\input{tables/a5}}
\end{table}

\paragraph{Whole-graph comparison and its controls}
In the whole-graph arm reported in \cref{tab:a5}, two graphs are judged in
both presentation orders, with several sampled votes per order, all votes
pooled, and a supermajority required for a winner. Six controls are enforced.
The share of votes cast for whichever graph was shown first is reported as
the primacy rate, where 0.5 is unbiased; a single greedy decode measures 0.70
on this task. Graph length is measured rather than suppressed: each model
emits every pair scoring within 70\% of its best pair on that image, up to
20, so that a model describing more of the scene is permitted to do so, and
the judge's own preference for longer text is measured by a \emph{padding}
control that compares a graph against itself with fabricated relations added.
A \emph{scramble} control compares a graph against a copy with its predicates
permuted across pairs; a judge that cannot prefer the intact graph is not
measuring anything, and its verdicts are discarded. Vocabulary breadth is a
confound, so two arms are run, one with our head reparameterised to the
benchmark's predicates, the same range as the baseline, and one with the full
vocabulary.

\paragraph{Certification}
A run is admitted only if the pooled controls satisfy three conditions at
once: scramble accuracy of at least 0.75, primacy of at most 0.60, and at
least ten decisive control comparisons. The last matters because a judge that
answers ``tie'' to a scrambled graph has demonstrated nothing, and a high
accuracy over three decisions is not evidence. The reported judge satisfies
all three (1.00, 0.19, 56 of 103). A smaller judge of the same family,
Qwen2.5-VL-7B \citep{qwen25vl}, was also evaluated and reaches the same
verdict at a lower scramble accuracy (0.95) and a lower decisive fraction,
which is why the larger one is reported.

\subsubsection{Per-relation oracle (A5b)}
\label{app:oracle:relation}

The pairwise oracle returns one verdict per pair of graphs and so cannot
indicate whether a losing graph lost because its additional relations were
false or because the judge prefers shorter text. A5b scores each asserted
relation separately: the judge sees one relation at a time and answers
whether the claim is true of the photograph.

\paragraph{Rendering}
Two renderings were evaluated. The plain rendering presents the judge with
the same numbered image used by the pairwise oracle, carrying up to a hundred
identically coloured boxes; the highlighted rendering draws a separate image
per relation with the subject and object boxes in different colours. Which is
better is decided by the control: a fixed share of judged relations have
their predicate replaced by another drawn from the same image, and the run
reports how often the judge accepts the corruption. That rate is 0.201 under
the plain rendering and 0.135 under the highlighted one, which indicates that
the plain rendering suffered from a grounding problem. The rate is an upper
bound rather than an exact false-positive rate, since a swapped predicate is
sometimes true by coincidence.

\paragraph{Depth}
Two depths were evaluated and neither supersedes the other. The deployed
graph is what each system emits when it asserts as much as its own confidence
spread justifies. The matched graph is the top ten pairs of each system with
no threshold, which removes length as a variable: at equal depth, the number
of accepted relations equals precision multiplied by depth.

\paragraph{Why informativeness is measured rather than elicited}
The judge was initially asked a second question, how much a relation conveys
beyond what the two object names already imply, because truth alone rewards a
degenerate system: a model answering \pred{on} for most pairs is usually
correct and conveys almost nothing. That question did not survive its own
audit. Over the 200-image run the two questions were nearly the same axis,
with $P(\text{info}=0 \mid \text{false})=0.998$ under both renderings and
$P(\text{info}\ge2 \mid \text{true})$ at 0.92 and 0.96, so a rate combining
them was very nearly precision multiplied by a constant; and repetition went
unpriced, the judge rating the baseline's \pred{on} relations at 1.05 on the
0--3 scale against 0.72 for its other predicates. The defect is not in the
wording of the rubric. The redundancy of a relation is a property of the
distribution in which it occurs, and a judge shown one relation at a time
cannot observe that distribution. Informativeness is therefore measured as
surprisal, $-\log_2 p(\text{pred})$, summed over the relations the judge
accepted, as in \cref{eq:truebits}.

\paragraph{Estimators}
The referenced estimator uses the PSG training marginal, which is shared by
both systems, so neither is scored against its own habits, and restricted to
the 56 predicates both emit, so no smoothing constant affects the comparison;
the out-of-vocabulary mass is reported alongside and never folded in.
\Cref{tab:a5b} reports the restricted-vocabulary arm, the like-for-like
comparison. An open-vocabulary arm was run as a control for vocabulary
breadth: at matched depth the judge accepts its relations at the same rate,
and it reaches 22.6 reference-free bits per image against 16.0. Since 73\% of
its predicates fall outside the reference, \cref{eq:truebits} cannot be
evaluated for it without a smoothing constant determining the comparison, and
it is excluded rather than scored on a biased subsample.

The reference-free estimator is a Krichevsky--Trofimov \citep{kt} description
length over each system's own stream of accepted predicates: the first
\pred{on} is expensive and the five-hundredth nearly free, and the code
charges for vocabulary size, which answers the objection that a larger
predicate list creates headroom by itself. It is stable in the threshold at
which rare predicates are pooled, yielding 25.0 / 24.8 / 23.7 bits per image
for the restricted arm over merge thresholds of 1, 3 and 10 occurrences
against 7.2 for the baseline at all three, and at matched depth 16.0 against
8.0. It therefore agrees with \cref{eq:truebits} on the ordering, at a
slightly wider ratio. The same asymmetry appears without any judge or
reference, in the predicate distributions themselves: entropy 4.00 against
2.14, and 1.9 distinct accepted predicates per graph against 1.5.
\Cref{tab:a5b} reports none of these, because the axis is defined as
\cref{eq:truebits} and a table of competing estimators would invite the
reader to choose one.

\paragraph{Normalisation}
\Cref{eq:truebits} is unbounded, so \cref{tab:sixaxes} and
\cref{fig:overview} divide it by the same quantity evaluated on the
ground-truth relations of the same images, which yields the share of the
information a human annotator recorded that a system delivers. Two properties
bound this reading. The ground truth counts every annotated relation,
including those whose endpoints the shared detector never found, so both
systems carry the detector's recall identically and the denominator does not
depend on its operating point. And the ratio can exceed one because the
annotation is sparse: at deployed depth our model reaches 1.03, asserting
18.3 relations per image against 6.33 in the annotation and being credited
for true relations that were never annotated. The matched-depth figure, at
which both systems are below one, is the one reported.

%% file: tables/ovs_loo.tex
\begin{tabular}{@{}l c c c@{}}
\toprule
\sffamily\bfseries axis withheld & \sffamily\bfseries OvSGTR & \sffamily\bfseries \model{} & \sffamily\bfseries ratio \\
\midrule
\ourrow none \note{the composite as reported} & 11.8 & 40.1 & 3.39$\times$ \\
\midrule
A1 transfer & 11.6 & 43.3 & 3.74$\times$ \\
A2 precision & 10.0 & 36.2 & 3.63$\times$ \\
A4 detector & 22.9 & 45.3 & 1.98$\times$ \\
A5 graph quality & 10.0 & 36.4 & 3.66$\times$ \\
A6 spatial & 10.9 & 40.6 & 3.73$\times$ \\
\bottomrule
\end{tabular}

%% file: tables/threshold.tex
\begin{tabular}{l ccccc}
\toprule
\sffamily\bfseries Criterion & $\tau$ & edges/img & recall & $P_\text{ann}$ & $P_\text{cal}$ \\
\midrule
max F1 (calibrated $P$) & 0.24 & 18.2 & 33.5 & 11.1 & 41.2 \\
max F1 (annotated $P$) & 0.41 & 7.9 & 23.4 & 17.9 & 53.2 \\
count-matched (6.0\,/img) & 0.45 & 6.1 & 20.1 & 19.8 & 56.1 \\
\midrule
\multicolumn{6}{l}{\note{AUC 0.929 $\cdot$ AP 0.090 $\cdot$ ECE 0.061 over all 377{,}986 scored pairs}} \\
\bottomrule
\end{tabular}

%% file: tables/boxbudget.tex
\begin{tabular}{ll cccccc}
\toprule
& & \multicolumn{3}{c}{\sffamily\bfseries conf 0.05, $\le$60 boxes} & \multicolumn{3}{c}{\sffamily\bfseries conf 0.001, $\le$100 boxes} \\
\cmidrule(lr){3-5}\cmidrule(lr){6-8}
\sffamily\bfseries Detector & \sffamily\bfseries Bench & box/img & obj\,rec & ceiling & box/img & obj\,rec & ceiling \\
\midrule
YOLOv8-m & VG150 & 27.0 & 69.5 & 63.8 & 99.4 & 85.2 & \best{82.9} \\
YOLO-World & VG150 & 26.2 & 58.6 & 50.1 & 99.3 & 78.7 & \best{74.8} \\
YOLOv8-m & PSG & 23.1 & 77.8 & 82.8 & 84.7 & 86.3 & \best{92.7} \\
YOLO-World & PSG & 22.7 & 66.4 & 69.6 & 95.8 & 80.3 & \best{87.6} \\
\midrule
\note{GroundingDINO (baseline's own)} & VG150 & \multicolumn{3}{c}{\note{no threshold applied}} & 97.1 & 83.9 & 81.6 \\
\bottomrule
\end{tabular}

%% file: tables/protocol.tex
\begin{tabular}{l cccccc}
\toprule
& \multicolumn{3}{c}{\sffamily\bfseries Base+Novel} & \multicolumn{3}{c}{\sffamily\bfseries Novel} \\
\cmidrule(lr){2-4}\cmidrule(lr){5-7}
\sffamily\bfseries Model / matcher & R@20 & R@50 & R@100 & R@20 & R@50 & R@100 \\
\midrule
OvSGTR Swin-T \citep{ovsgtr}, \note{as published} & 15.8 & 20.5 & 23.9 & 10.2 & 13.5 & 16.2 \\
\quad reproduced, \note{their matcher} & 16.0 & 20.4 & 23.8 & 10.1 & 13.2 & 15.9 \\
\quad + one-to-one assignment & 13.9 & 18.0 & 21.2 & 8.6 & 11.4 & 13.8 \\
\midrule
OvSGTR Swin-T + MegaSG \citep{ovsgtr}, \note{as published} & 19.4 & 25.4 & 29.7 & 12.2 & 17.0 & 21.1 \\
\quad reproduced, \note{their matcher} & 19.5 & 25.5 & 29.9 & 12.4 & 17.2 & 21.3 \\
\quad + one-to-one assignment & 17.2 & 22.6 & 26.7 & 10.8 & 14.9 & 18.6 \\
\bottomrule
\end{tabular}

%% file: tables/a5.tex
\begin{tabular}{lccccc}
\toprule
\multicolumn{6}{l}{\sffamily\itshape graph shape, no ground truth and no judge} \\
\sffamily\bfseries model & rel./graph & distinct pred. & modal share & vocab used & at cap \\
\midrule
\ourrow \model{}, vocab = pack & 18.3 & 4.5 & 0.52 & 56 & 81\% \\
OvSGTR \citep{ovsgtr} & 10.3 & 1.7 & 0.85 & 26 & 24\% \\
\ourrow \model{}, vocab = train & 18.9 & 8.4 & 0.33 & 939 & 88\% \\
\midrule
\multicolumn{6}{l}{\sffamily\itshape oracle verdicts, ours against the baseline} \\
\sffamily\bfseries arm & compared & decisive & win rate & longer: win rate ($n$) & equal: win rate ($n$) \\
\midrule
vocab = pack & 140 & 58 & 0.16 & 0.10 (101) & 0.40 (39) \\
vocab = train & 141 & 46 & 0.24 & 0.16 (106) & 0.62 (35) \\
vocab = pack, top 10 each & 142 & 64 & 0.62 & --- & 0.62 (142) \\
\midrule
\multicolumn{6}{l}{\note{controls: scramble accuracy 1.00 (56 decisive of 103), padding resistance 1.00 (5 of 30), primacy 0.19}} \\
\multicolumn{6}{l}{\note{judge Qwen/Qwen3-VL-8B-Instruct, 6 votes per comparison, majority 0.67, both presentation orders}} \\
\multicolumn{6}{l}{\note{matched-depth run, same judge: scramble accuracy 1.00 (24 decisive of 30), padding resistance 1.00 (15 of 28), primacy 0.23}} \\
\bottomrule
\end{tabular}

%% file: app/e_results.tex
\section{Additional results}
\label{app:results}
\label{app:perclass}

\subsection{Per-predicate results}

\begin{figure*}[!htb]
\centering
\includegraphics[width=0.88\linewidth]{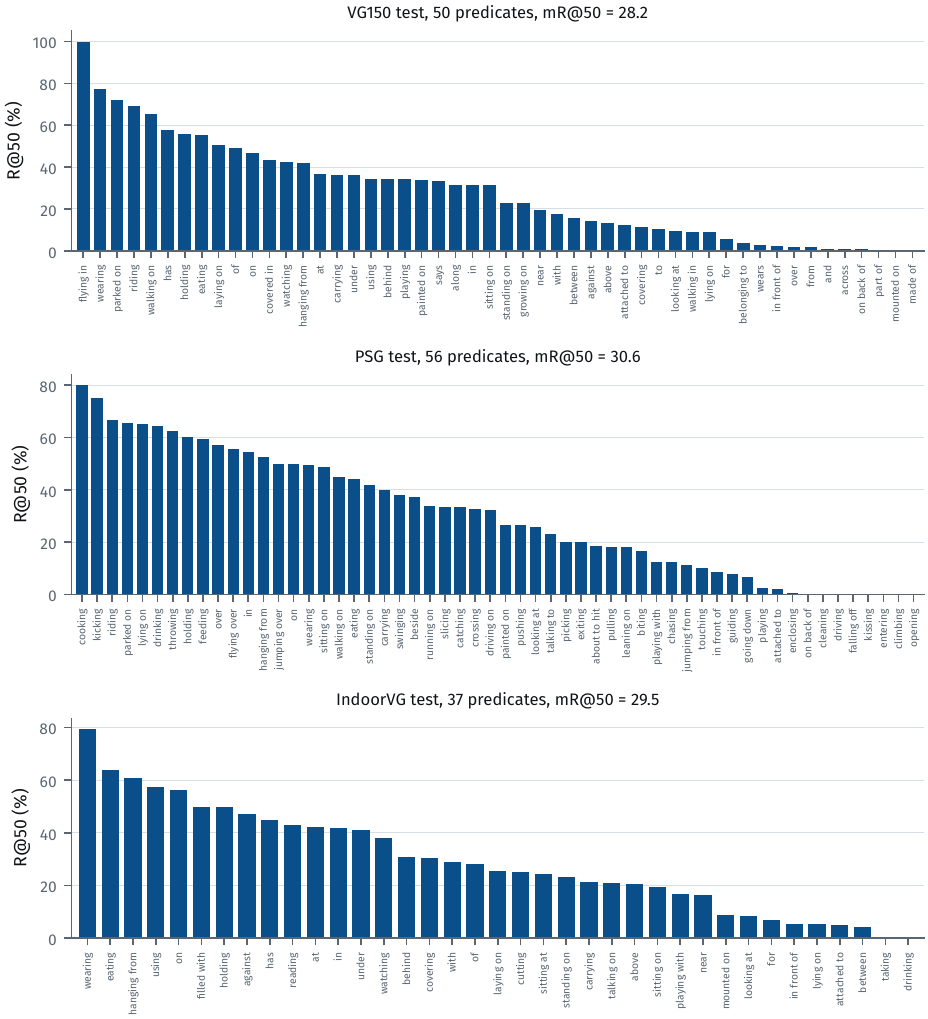}
\caption{Per-predicate R@50 of the released model on the three
closed-vocabulary transfer benchmarks (test splits, ground-truth boxes,
graph-constrained, exact-string matching). Predicates at zero are, almost
without exception, members of a synonym group whose other surface form the
model predicts instead (\cref{sec:scale:region}).}
\label{fig:perclass}
\end{figure*}

\Cref{fig:perclass} gives every per-predicate recall underlying the
mean-recall columns of \cref{tab:a1}. Two of the patterns visible there are
emission rules rather than recognition failures. HICO-DET annotates several
verbs per pair and scores each of them while the model emits only one, so the
general verb occupies the single slot: \pred{holding} absorbs
\pred{wielding}, \pred{carrying} and \pred{reading}, and \pred{riding}
absorbs \pred{straddling} and \pred{sitting on}, so that a specific verb the
model ranks second counts as a miss. The remedy is a multi-label emission
rule for multi-label sources rather than additional capacity. On PSG, where
the model emits exactly as many edges as the image has annotations, the
misses are the informative part: the annotation records \pred{attached to}
between umbrellas where the model predicts \pred{hanging from} and \pred{in
front of}. Unannotated true relations receive no credit, and near-synonyms
count as wrong.

\subsection{Priors join and detector retention}
\label{app:closed}

\begin{figure}[h]
\centering
\includegraphics[width=0.85\linewidth]{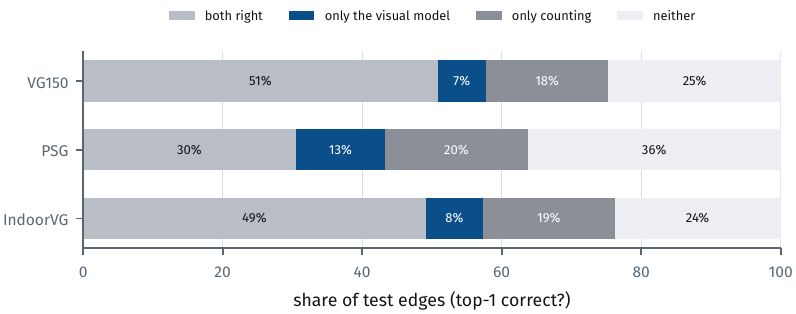}
\caption{\textbf{Edge-by-edge join of the visual model against the frequency
baseline}, top-1 correct or not, on the same edges as \cref{tab:priors}. The
frequency baseline is correct where the model is wrong about $2.4\times$ as
often as the reverse, but the model-only share is 7--13\% of all edges, so
the two are correct on partly different edges rather than the model being a
weaker version of the prior.}
\label{fig:join}
\end{figure}

\begin{figure}[h]
\centering
\includegraphics[width=0.78\linewidth]{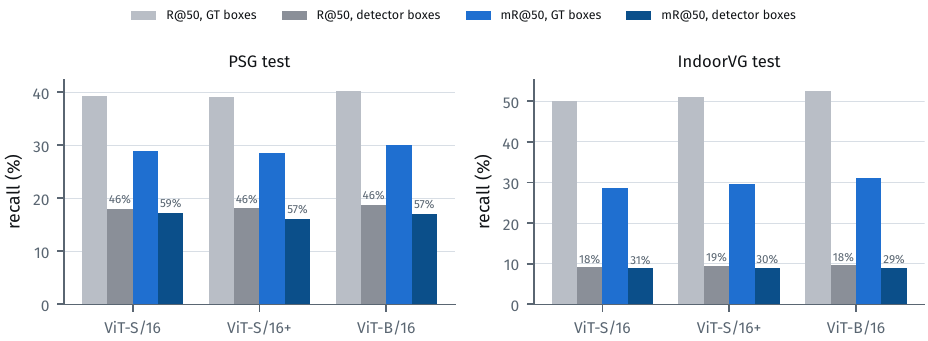}
\caption{\textbf{Retention under a real detector for each tower}, YOLO-World
at a confidence threshold of 0.05, with one-to-one, class-aware (strict) matching as in
\cref{sec:protocol} and a pair budget of 500 on both sides; percentages are
retention relative to ground-truth boxes. The micro ladder survives
detection, the macro ladder flattens, and IndoorVG, whose categories the
detector covers poorly, loses 70--80\% of the model's performance to
detection.}
\label{fig:detbox}
\end{figure}

\begin{figure*}[!htbp]
\centering
\includegraphics[width=\linewidth]{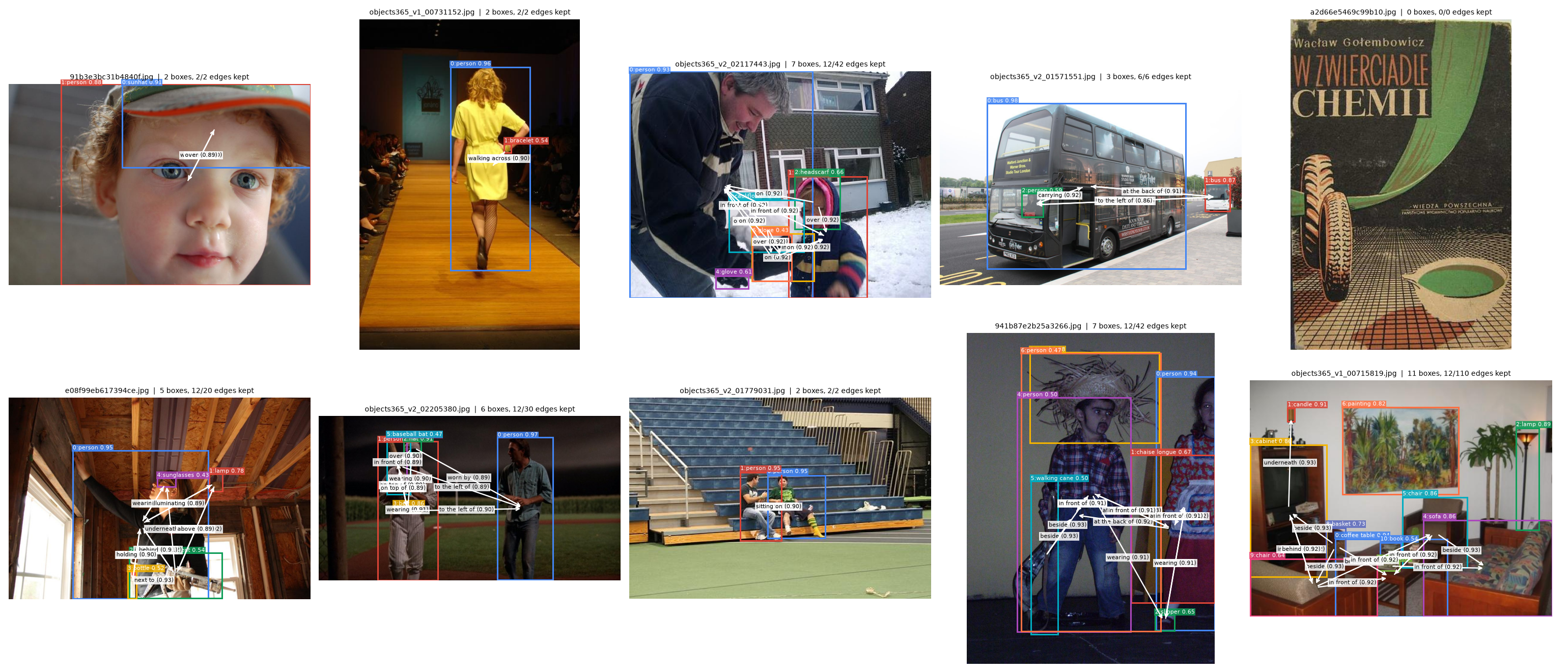}
\caption{\textbf{Regions from a detector prompted with the LVIS vocabulary},
which the relation model never saw. YOLOE \citep{yoloe} with all 1{,}198 LVIS
classes \citep{lvis} on MegaSG validation images, with the boxes passed to the
released model unchanged and a 35-predicate vocabulary. The names on the
boxes are the detector's and are shown for the reader only; the edges are
those retained at the released operating point.}
\label{fig:lvisqual}
\end{figure*}

\Cref{fig:join} is the edge-by-edge join underlying \cref{tab:priors},
\cref{fig:detbox} shows what each tower retains under a real detector
(\cref{sec:transfer:deploy}), and \cref{fig:lvisqual} shows the released
model behind a detector prompted with the LVIS vocabulary
(\cref{sec:anyinputs:regions}).

\subsection{Bounds on the spatial-axis claim}
\label{app:closed:spatial}

\begin{figure*}[t]
\centering
\includegraphics[width=0.75\linewidth]{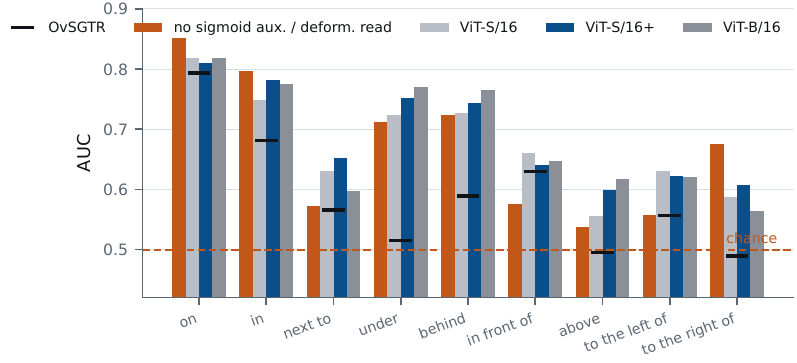}
\caption{\textbf{Axis 6: per-predicate AUC on SpatialSense}, whose
adversarially collected negatives defeat co-occurrence priors. The ablated
arm is the recipe without the sigmoid auxiliary and the deformable read; its
projective predicates (\pred{above}, \pred{to the left of}, \pred{in front
of}) lie at 0.54--0.58, and \pred{to the right of} is the one projective
predicate it already handles. The released family raises that set. Black
dashes mark the A6 baseline, OvSGTR on the same cells under the same metric
implementation, which is at or below chance on \pred{above} and \pred{to the
right of}.}
\label{fig:spatial}
\end{figure*}

\begin{figure}[t]
\centering
\includegraphics[width=0.85\linewidth]{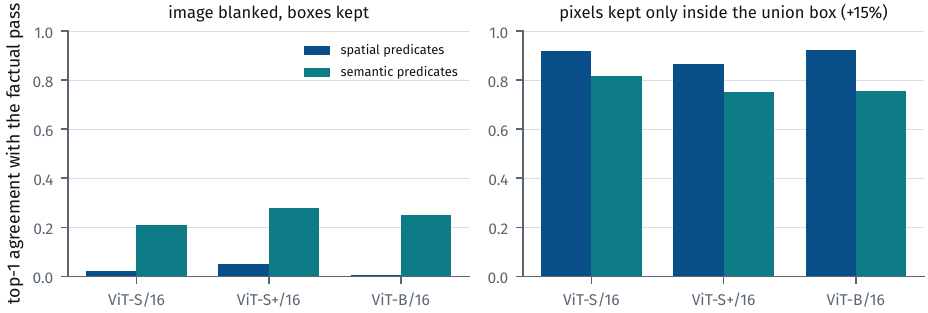}
\caption{\textbf{Grounding probes across the ladder}, PSG val, with the
sampler frozen so that the factual and counterfactual passes score the same
pairs. Left: with the image blanked, spatial predicates retain 1--5\% of
their top-1 decisions and semantic ones 21--28\%, which indicates that the
model reads geometry from the boxes and appearance from the pixels. Right:
with pixels retained only inside the union box, 75--92\% of top-1 decisions
survive; the scene context added by the deformable read rarely changes the
answer on PSG.}
\label{fig:counterfactual}
\end{figure}

\Cref{fig:spatial} gives the per-predicate view of A6 and
\cref{fig:counterfactual} the grounding probes. Three facts bound the claim
made in \cref{sec:transfer:spatial}. The boxes-only baseline of the
benchmark's authors scores 68.8 without seeing the image, so a pooled AUC
near 0.69 is not by itself evidence of visual spatial reasoning. A
counterfactual spatial-grounding probe varies between 0.01 and 0.81 as a
function of the backbone learning rate alone, which invalidates any grounding
claim resting on a single arm, including ours. And an
appearance-versus-geometry probe refutes the spatial-shortcut hypothesis in
the opposite direction, since object appearance moves the predicate by $+34$
points relative to the control. The supervision-ceiling reading of
\cref{sec:transfer:spatial} is tested against capacity in
\cref{sec:transfer:capacity}.

\subsection{Closed-set and localisation legs of the VLM comparison}
\label{app:baselines}

\begin{table}[h]
\centering\small
\caption{\textbf{Free-text systems across three benchmarks}, each under its own
protocol: ours answers in the benchmark's own words, ROBIN-3B in its own under
its published SBERT rule. Ground-truth regions, graph-constrained, on the
image subsets ROBIN was run on (VG150 5k, IndoorVG in full, PSG test).
\emph{Pair cov.} is the share of annotated pairs a system reaches at all.}
\label{tab:baselinescross}
\resizebox{\linewidth}{!}{\input{tables/baselines_cross}}
\end{table}

\Cref{tab:baselinescross} is the cross-dataset comparison summarised in
\cref{sec:results:a3}. The margin is smallest on PSG, $1.1\times$ on micro
recall, and widens to $1.9\times$ on VG150 and $2.2\times$ on IndoorVG; PSG is
also the only one of the three on which free-text systems are routinely
reported.

\begin{table}[h]
\centering\small
\caption{\textbf{Closed-set and localisation legs of the VLM comparison.} PSG
test, 2{,}179 images, graph-constrained, the same scorer as
\cref{tab:baselines}. Top: the model is given the 56 PSG predicates instead of
answering in its own words, with ground-truth regions throughout. Bottom: the
regions are withheld and the model must localise as well; \emph{strict}
requires the predicted object class to match, \emph{lenient} matches on IoU
alone, and mAP@50 is COCO-style detection AP over the 133 PSG object
categories, defined only in this block.}
\label{tab:baselinesclosed}
\resizebox{\linewidth}{!}{\input{tables/baselines_psg_closed}}
\end{table}

\Cref{tab:baselinesclosed} gives the two legs omitted from
\cref{tab:baselines}. Of three ways of indicating the regions to a prompted
model, namely pixel boxes listed in the prompt, numbered overlays, and boxes
drawn on the image, every reported run uses the last. Prompting these models with the benchmark's own 56
predicates rather than letting them answer freely changes their R@20 by
between $-2.3$ and $+1.8$ points and raises the mean recall of every one of
them, by $+0.5$ to $+3.1$, leaving their order on R@20 unchanged; on mean
recall InternVL3.5-8B and Qwen3-VL-8B exchange places.
This is the setting in which A1 places a model, but it is not the setting of
a deployed open-vocabulary model, which is why the main text reports the
open-vocabulary leg.

The lower block withholds the ground-truth regions. The general models fall
to 0.6--1.5 R@20, and their localisation is not the reason: 23.5 and 25.6
mAP@50 over 133 categories is genuine detection. The two facts are consistent
because the pairs a model can reach scale with the square of its box recall
(at IoU box recall of 0.32 and 0.38, only 10\% and 14\% of annotated pairs
have both endpoints present), and because these models emit only 1.1 to 2.3
relations per image. This is the ceiling \cref{fig:boxbudget} measures for
detectors, at a harsher operating point, and the reason A4 evaluates every
system on a single shared detector.

%% file: tables/baselines_cross.tex
\begin{tabular}{lllccccrr}
\toprule
\sffamily\bfseries Benchmark & \sffamily\bfseries Model & \sffamily\bfseries Setting & \sffamily\bfseries R@20 & \sffamily\bfseries R@50 & \sffamily\bfseries mR@20 & \sffamily\bfseries mR@50 & \sffamily\bfseries pair cov. & \sffamily\bfseries rel/img \\
\midrule
VG150 & \model{} & closed set & 43.6 & 51.2 & 23.1 & 27.1 & --- & top-$K$ \\
 & ROBIN-3B & SBERT argmax & 23.3 & 27.3 & 12.4 & 15.2 & 61.4 & 25.5 \\
\midrule
PSG & \model{} & closed set & 34.7 & 40.1 & 28.4 & 30.6 & --- & top-$K$ \\
 & ROBIN-3B & SBERT argmax & 30.4 & 35.7 & 21.1 & 22.9 & 77.4 & 34.0 \\
\midrule
IndoorVG & \model{} & closed set & 44.7 & 52.9 & 24.0 & 29.6 & --- & top-$K$ \\
 & ROBIN-3B & SBERT argmax & 19.4 & 24.2 & 17.5 & 21.5 & 64.3 & 33.1 \\
\bottomrule
\end{tabular}

%% file: tables/baselines_psg_closed.tex
\begin{tabular}{llcccccc}
\toprule
\sffamily\bfseries Model & \sffamily\bfseries Setting & \sffamily\bfseries R@20 & \sffamily\bfseries R@50 & \sffamily\bfseries mR@20 & \sffamily\bfseries mR@50 & \sffamily\bfseries mAP@50 & \sffamily\bfseries rel/img \\
\midrule
\multicolumn{8}{l}{\itshape Closed set --- the model is prompted with PSG's 56 predicates} \\
RelateAnything ViT-S/16\textsuperscript{+} (53M) & closed-set & 34.7 & 40.1 & 28.4 & 30.6 & --- & top-$K$ \\
Qwen3-VL-32B & closed-set & 14.2 & 15.2 & 8.2 & 8.6 & --- & 19.9 \\
Qwen3-VL-8B & closed-set & 12.0 & 12.3 & 5.6 & 5.7 & --- & 14.8 \\
InternVL3.5-8B & closed-set & 9.1 & 9.1 & 5.0 & 5.0 & --- & 10.8 \\
GLM-4.6V-Flash & closed-set & 6.6 & 6.8 & 3.3 & 3.4 & --- & 13.7 \\
\midrule
\multicolumn{8}{l}{\itshape SGDet --- the model predicts its own boxes} \\
Qwen3-VL-8B & strict & 0.6 & 0.6 & 0.3 & 0.3 & 23.5 & 1.1 \\
Qwen3-VL-8B & lenient & 1.0 & 1.0 & 0.5 & 0.5 & 23.5 & 1.7 \\
Qwen3-VL-32B & strict & 0.8 & 0.8 & 0.3 & 0.3 & 25.6 & 1.3 \\
Qwen3-VL-32B & lenient & 1.5 & 1.5 & 0.8 & 0.8 & 25.6 & 2.3 \\
\bottomrule
\end{tabular}

%% file: app/f_probes.tex
\section{Representation probes}
\label{app:probes}
\label{app:arch:finetune}

\begin{table*}[!htb]
\centering\small
\caption{\textbf{Effect of relation fine-tuning on the dense backbone
features}, for the pre-trained backbone and for the two towers of
\cref{sec:corpus:mixture}. \emph{Patch similarity} (PSG val, last state,
per-image mean-token subtraction) queries from subject-box patches and
measures the mean cosine similarity to the relation partner, to another
object of the same class, and to unrelated regions; \emph{class} and
\emph{rel.\ sel.} are the corresponding selectivity margins, and
$P(\text{partner}>\text{same class})$ is the per-query win rate against a
chance level of 0.500. \emph{Interaction region} (HICO-DET test, 736
hand--verb relations) queries the subject patch nearest the object within the
intersection: \emph{contact adv.} is its similarity advantage over a distant
subject patch, and the 1-NN columns give cross-image retrieval accuracy for
the \emph{verb} and for the \emph{object class} from that same feature.}
\label{tab:repr}
\resizebox{\linewidth}{!}{\input{tables/representation}}
\end{table*}

\begin{figure}[h]
\centering
\includegraphics[width=0.78\linewidth]{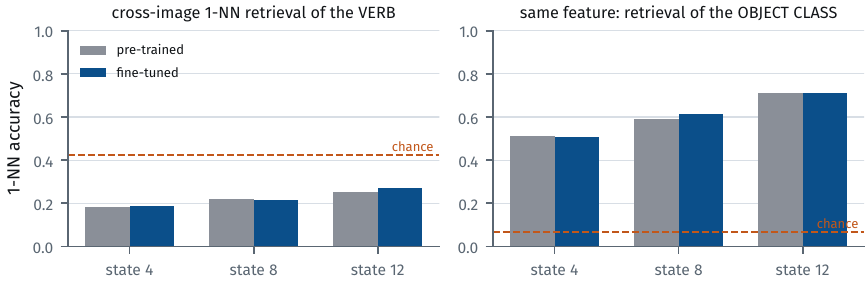}
\caption{\textbf{Verb and object retrieval from the contact-region feature.}
From the contact-region feature of 736 HICO-DET hand--verb relations,
cross-image nearest-neighbour retrieval of the verb remains below the
majority-class chance level at every depth, before and after fine-tuning,
whereas retrieval of the handled object's class from the same feature is
$10\times$ chance.}
\label{fig:affordance}
\end{figure}

\begin{figure*}[t]
\centering
\includegraphics[width=0.9\linewidth]{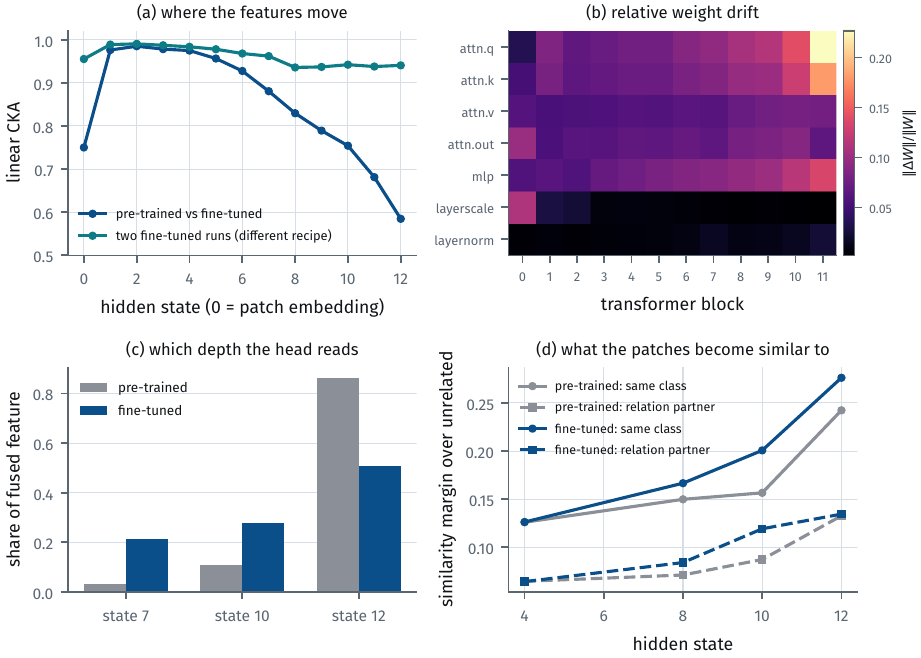}
\caption{\textbf{Effect of relation fine-tuning on the DINOv3 backbone}
(ViT-S/16+, 200 PSG-val images). (a)~Linear CKA between the pre-trained and
fine-tuned features decreases monotonically with depth, to 0.58 at the last
state, whereas two fine-tuned runs under \emph{different} recipes agree at
0.94, which indicates that the task rather than the recipe determines where
the features go. (b)~Relative weight drift is small and uniform (6--9\% per
block for attention and MLP, 1\% for the normalisation parameters) except in
the last block, where the query and key projections move by 23\% and 18\%,
so the change is concentrated in last-block routing. (c)~The learned fusion
over the three taps shifts from 86\% on the last state to 51\%, recruiting
the mid-depth taps. (d)~Patch similarity margins over unrelated regions: the
margin to \emph{same-class} objects grows at every depth, whereas the margin
to the \emph{relation partner} is unchanged.}
\label{fig:backbone}
\end{figure*}

\Cref{tab:repr}, \cref{fig:affordance} and \cref{fig:backbone} are the probes
underlying \cref{sec:decouple:features}. Beyond the findings reported there,
what fine-tuning adds is small and lies elsewhere. A linear probe on frozen
box-pooled tokens shows that relatedness beyond object identity is already
present in the pre-trained backbone, with a within-class-pair AUC of 0.73 on
PSG against 0.70 for a geometry-only baseline, and fine-tuning does not
increase it ($0.73 \to 0.72$). What increases is the linear decodability of
the predicate, with macro-F1 rising from 0.445 to 0.475 on PSG and from 0.264
to 0.305 on VG150. This is the gradient of the pair head reaching the
features it reads, rather than a relational representation forming within
them.

%% file: tables/representation.tex
\begin{tabular}{lccc ccc}
\toprule
& \multicolumn{3}{c}{patch similarity (PSG, $n=300$)} & \multicolumn{3}{c}{interaction region (HICO-DET, $n=736$)} \\
\cmidrule(lr){2-4}\cmidrule(lr){5-7}
backbone & class sel. & rel.\ sel. & partner $>$ same-cls & contact adv. & verb 1-NN & object 1-NN \\
\midrule
DINOv3, pre-trained & 0.242 & 0.133 & 0.469 & 0.356 & 0.251 & 0.713 \\
+ relation fine-tuning & 0.276 & 0.134 & 0.387 & 0.377 & 0.270 & 0.713 \\
+ relation fine-tuning (shipped) & 0.281 & 0.138 & 0.385 & 0.438 & 0.269 & 0.761 \\
\midrule
\note{chance} & \note{---} & \note{---} & \note{0.500} & \note{---} & \note{0.424} & \note{0.068} \\
\bottomrule
\end{tabular}

%% file: app/g_scaling.tex
\section{Backbone scaling}
\label{app:scaling}

\begin{table*}[!htb]
\centering\small
\caption{The backbone ladder, matched on corpus, schedule, batch size and
every other hyper-parameter, for the released family and for the zero-shot
trio trained without the HICO-DET share and without source-aware negatives.
Axis scores are chance-corrected (\cref{sec:bench}). \textbf{The OVS column
here is computed over A1, A2, A4 and A6}, not the five-axis OVS of
\cref{tab:sixaxes}: A5 costs one judge run per arm and was run for the
released tower only, so the two numbers are not comparable. A3 is reported
alongside the composite axes. Throughput and memory are measured on the
training hardware (4$\times$A100). The released tower is highlighted.}
\label{tab:scaling}
\input{tables/scaling}
\end{table*}

\begin{figure}[h]
\centering
\includegraphics[width=0.85\linewidth]{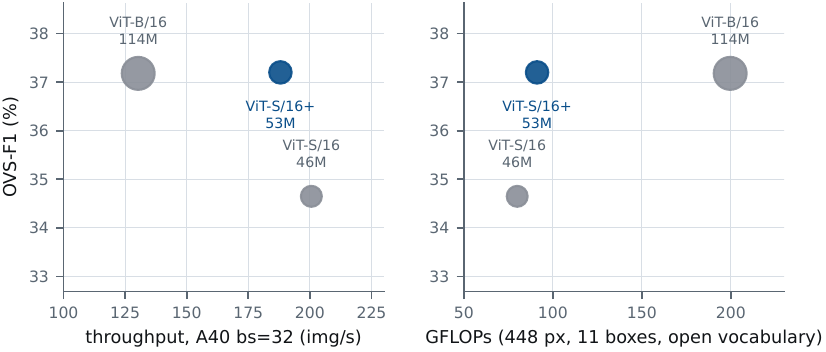}
\caption{The released family on the cost--accuracy plane; bubble area is
proportional to parameter count, and both axes come from the same benchmark
as \cref{tab:latency}. Left: batched throughput (batch-1 latency is
dispatch-bound and nearly identical across towers, 19.3--20.0 ms on an A40).
Right: measured GFLOPs (dense operations, 448 px, 11 boxes, full vocabulary).
Relative to the released tower, ViT-B costs $2.2\times$ the FLOPs and $-31\%$
throughput for $-0.02$ OVS-F1, and the advantage it holds on individual axes
does not survive the deployment operating point.}
\label{fig:family}
\end{figure}

\Cref{tab:scaling} and \cref{fig:family} support
\cref{sec:datalessons:capacity}. One observation belongs with them. On the
zero-shot ladder the train--validation NCE gap at the final epoch is $+0.083
/ +0.094 / +0.159$ and the best development epoch moves earlier (12 / 8 / 7),
so the ladder brackets the optimum: ViT-S is still improving when training
ends and ViT-B has begun to overfit, which is why ``capacity-limited''
applies to the S class only. Separately, a DINOv3-ConvNeXt family trained for
a latency ladder places its tuned tiny variant in the ViT-S accuracy class
and makes it the fastest of all at batch size one; its stage fusion is
dominated by the coarse stages by an order of magnitude, and forcing the fine
stages up changes nothing. It remains an option where dispatch latency
dominates.

\subsection{Effect of backbone capacity on transfer}
\label{sec:transfer:capacity}

The ladder of \cref{tab:scaling} was trained twice, once with the released
recipe and once without the HICO-DET share, changing nothing but the tower.
Capacity does not raise the composite, and on neither ladder is the composite
monotone in model size. On the released ladder, S $\to$ S+ costs $+15\%$
parameters and yields $+7.4\%$, whereas S+ $\to$ B costs $+114\%$ parameters,
$1.33\times$ memory and $1.22\times$ wall-clock time and yields $-0.1\%$; the
largest tower does not outperform the released one. On the zero-shot ladder
the middle rung is the weakest of the three, $-1.5\%$ below S, before B
recovers $+3.1\%$. Three observations matter more than the ordering.

\emph{Spatial axis.} On both ladders ViT-S/16+ is the best of the three on
A6, and ViT-B is worse than a model of less than half its size. This is the
deciding evidence for the supervision-ceiling reading of
\cref{sec:transfer:spatial}: capacity cannot substitute for supervision that
is absent.

\begin{figure}[t]
\centering
\includegraphics[width=0.85\linewidth]{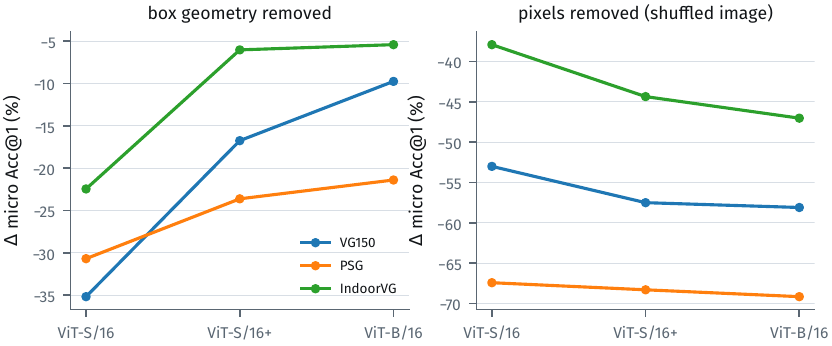}
\caption{\textbf{Lesion ladder across backbone sizes.} The lesion ladder of
\cref{tab:attrib} at all three sizes: the cost of removing box geometry
decreases with size while the cost of removing pixels increases, in every
source. Greater robustness would have moved both in the same direction.}
\label{fig:lesion}
\end{figure}

\begin{figure}[t]
\centering
\includegraphics[width=0.85\linewidth]{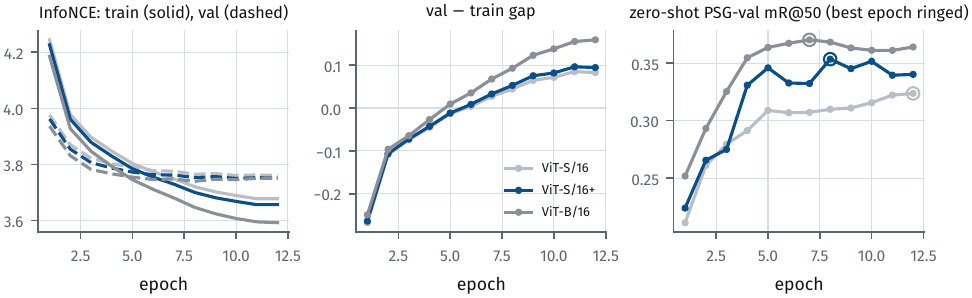}
\caption{\textbf{Training dynamics of the ladder.} Train and validation
InfoNCE, their gap, and the zero-shot selection metric per epoch. The gap
grows with width and the best epoch arrives earlier, so the ladder brackets
the optimum.}
\label{fig:laddercurves}
\end{figure}

\emph{Source of the evidence.} The lesion ladder of \cref{tab:attrib}, run at
all three sizes (\cref{fig:lesion}), shows that dependence on box geometry
decreases with size (from $-35.2\%$ to $-9.8\%$ on VG150) while dependence on
pixels increases (from $-53.0\%$ to $-58.1\%$), monotonically and in all
three sources. The two move in opposite directions, which rules out the
simpler explanation that larger models are more robust to any lesion. Larger
backbones recover from pixels what smaller ones read from explicit box
coordinates: a substitution rather than an addition, and consistent with the
training dynamics of \cref{fig:laddercurves}.

\emph{Deployment.} Run through the deployment path, with detector boxes, its
own calibration fit and its own count-matched threshold, the two larger
towers tie, and S+ achieves the higher value on the only precision we can
measure against adjudicated negatives (0.903 against 0.893). Benchmark recall
reads 50 ranked slots per image, whereas the operating point emits 5--7
edges, about 70\% of which fall on boxes that no ground-truth object matches.
The capacity advantage therefore lies in a part of the ranking that
deployment never emits, which is the evidence behind the choice of tower in
\cref{sec:datalessons:capacity}.

%% file: tables/scaling.tex
\begin{tabular}{lcccccccrrrr}
\toprule
\sffamily\bfseries Backbone & \sffamily\bfseries OVS-F1 & \sffamily\bfseries A1 & \sffamily\bfseries A2 & \sffamily\bfseries A3 & \sffamily\bfseries A4 & \sffamily\bfseries A6 & \sffamily\bfseries HICO F1 & \sffamily\bfseries params & \sffamily\bfseries img/s & \sffamily\bfseries peak GB & \sffamily\bfseries wall \\
\midrule
\multicolumn{12}{l}{\sffamily\itshape released family: RA-4M + 5\% HICO-DET train, source-aware negatives} \\
ViT-S/16 & 34.6 & 27.4 & 59.6 & 30.6 & 24.3 & 35.1 & 36.4 & 46.1M & 114 & 25.1 & 4:52 \\
\ourrow ViT-S/16+ & \best{37.2} & 28.6 & 60.8 & 32.5 & \best{27.4} & \best{37.9} & 37.1 & 53.2M & 109 & 30.0 & 5:04 \\
ViT-B/16 & \best{37.2} & \best{28.9} & \best{64.4} & \best{33.4} & 26.8 & 37.2 & \best{37.5} & 113.8M & 87 & 39.8 & 6:09 \\
\midrule
\multicolumn{12}{l}{\sffamily\itshape zero-shot trio: RA-4M + raw Visual Genome, no HICO-DET share} \\
ViT-S/16 & 34.7 & 23.2 & 53.0 & 31.1 & \best{27.0} & 38.7 & 18.5 & 46.1M & 115 & 25.0 & 3:58 \\
ViT-S/16+ & 34.2 & 23.4 & 53.1 & 32.5 & 24.7 & \best{40.8} & 18.7 & 53.2M & 108 & 29.9 & 4:12 \\
ViT-B/16 & \best{35.3} & \best{24.7} & \best{56.0} & \best{33.4} & 25.8 & 39.8 & \best{20.0} & 113.8M & 87 & 39.8 & 5:10 \\
\bottomrule
\end{tabular}

%% file: app/h_ablations.tex
\section{Ablations and negative results}
\label{app:ablations}

\subsection{Selection protocol and reproducibility}
\label{sec:transfer:repro}

Because a full-scale arm costs several GPU-hours, the recipe was developed on
a 50k-image proxy of the corpus and promoted according to criteria fixed in
advance. The reliability of that proxy is itself a measured result. Its
\emph{ranking} of arms settles early: on the selection metric, the sign of
the difference between two arms agrees with the twelve-epoch outcome for
82--86\% of the comparisons that exceed the noise floor, at every epoch from
the first. Its \emph{levels} do not settle, the composite of the winning arm
rising from 0.339 on the proxy to 0.371 at full scale. The noise floor
between seed-matched reruns is 1.3\%, and every difference reported in this
paper should be read against it.

Two properties of the selection are worth stating, because both could have
biased the outcome. The recipe was chosen on a composite over A1, A2, A3 and
A6, before A4 and A5 were defined. Recomputing every selection decision on
the A1/A2/A4/A6 composite changes no outcome: the same three towers pass, the
same one is released, and the backbone-learning-rate comparison still
resolves to the default arm, by $-0.012$ OVS-F1 against a $+0.003$ threshold
where the earlier composite gave $-0.006$. The selection is therefore not an
artefact of which axes are summed. Selection also reads the final epoch
rather than the best one, which predicts out-of-distribution behaviour better
(\cref{sec:transfer:indomain}) and inverts the ordering of some arms,
including the comparison between backbone adaptation schemes.

\begin{figure}[t]
\centering
\includegraphics[width=0.85\linewidth]{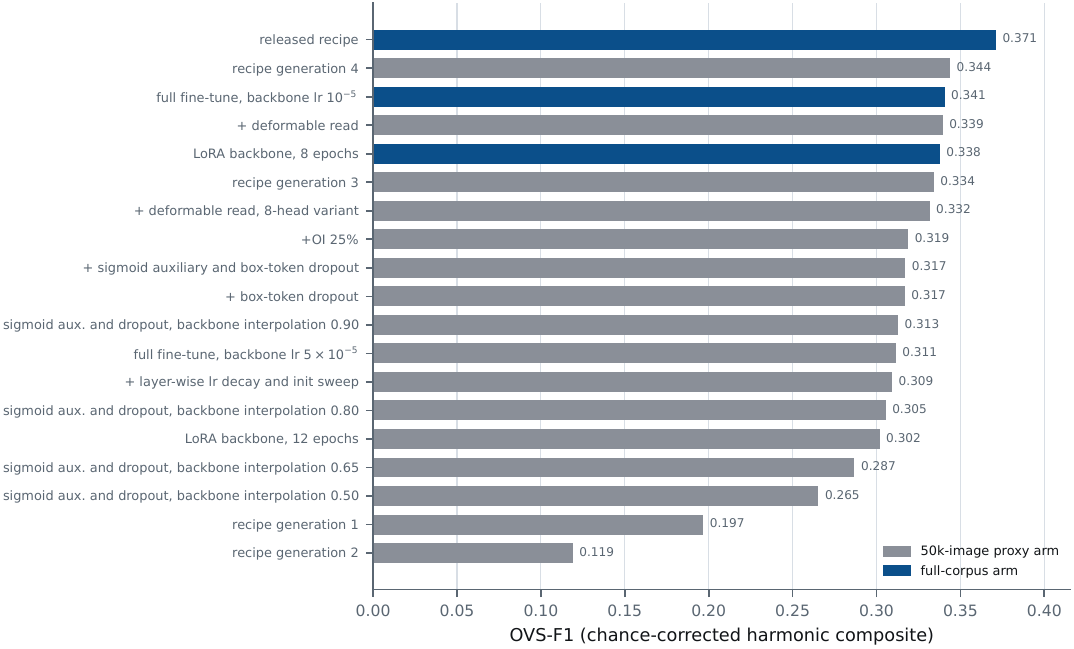}
\caption{\textbf{The recipe ladder}: every arm with a complete axis set,
ordered by composite. Development arms are scored on the axis set that
existed while the recipe was being chosen (A1, A2, A3, A6); none of them was
run in detection mode, so none has an A4 cell, and the ladder is internally
comparable but not comparable with \cref{tab:sixaxes}. Grey arms are
50k-image proxies and blue arms are full-corpus runs. The proxy winner became
the full recipe.}
\label{fig:proxy}
\end{figure}

\begin{table}[h]
\centering\small
\caption{\textbf{The recipe ladder in tabular form}: every arm with a complete
axis set, the four development axes chance-corrected, and their harmonic mean
OVS-dev. This is the number on which the recipe was selected, and it is not
comparable with the five-axis OVS of \cref{tab:sixaxes}, since no development
arm has an A4 or A5 cell. The Open Images arm should be read against the
deformable-read arm, the matched arm without that share.}
\label{tab:ladder}
{\footnotesize\input{tables/ladder}}
\end{table}

Every number in the tables and figures of this paper is generated by script
from evaluation artefacts; the only hand-entered values are literature rows,
each of which carries its source. All recall is graph-constrained, evaluation
is deterministic, and claims that rest on a single seed are identified as
such. The corpus, the model weights, the benchmark tooling, and the
generation and evaluation code are released.

\subsection{Attributions}

The following attributions come from the ladder of \cref{fig:proxy} and
\cref{tab:ladder}, each from a single-variable arm against a matched control,
and should be read against the 1.3\% noise floor.

\begin{itemize}\itemsep3pt
\item \textbf{Sigmoid auxiliary}: moves the projective spatial band at a small
  transfer cost, and makes the scale of the output head trainable.
\item \textbf{Deformable scene read}: improves every axis and recovers the
  transfer cost of the sigmoid auxiliary; the two together constitute the
  full-scale recipe.
\item \textbf{Layer-normalised feature taps}: free, and equivalent to a
  multi-level read costing $3\times$ as much, which was therefore not
  adopted.
\item \textbf{Multi-scale sampling}: $+3.7\%$ on the development composite
  with 9 of 11 cells improving, and a regulariser rather than a resolution
  effect: it reduces the proxy's train--validation gap by 60\%, so its gain
  at full scale is correspondingly smaller, as that mechanism predicts.
\item \textbf{Same-predicate feature mixing}: $+10.1\%$ zero-shot tail mean
  recall, reproduced with two seeds, whereas the identical mechanism with
  \emph{random} partners yields $-2.8\%$. The class-conditional matching
  rather than the mixing is the mechanism; both mixing arms have a lower
  training loss than the control, which rules out an augmentation effect and
  points to prototype smoothing.
\item \textbf{Source-aware negative masking}: $+12\%$ on the spatial axis
  when corpora of different annotation scope are mixed (\cref{app:aux}), and
  the best composite of the family.
\end{itemize}

\subsection{A prior violation traced to one loss term}
\label{sec:transfer:wearing}

One diagnostic is reported here because no axis of \cref{sec:bench} can
detect it. The corpus teaches that a person wears a bounded set of garments
attached to their body. Under a gated variant of the background-suppression
term, the trained model violated that prior $22\times$ more often than its
training data does, asserting \pred{wearing} between disjoint boxes at a rate
of 11.04\%.

Three explanations were tested and rejected. A vocabulary explanation fails,
since 85\% of the type errors involve body parts absent from the source
ontology, so the model is not choosing incorrectly among available
categories. A graph-reasoning explanation fails, since the fan-in is 97\%
co-reference, that is, the same garment attributed to several boxes of one
person, which requires no relational inference. A spatial-shortcut
explanation fails in the opposite direction, since probes show that the
entity features already identify body parts correctly, with 66.9\% assigned
to the correct family. Removing the gate reduced the disjoint-box
\pred{wearing} rate to 0.00\% at no cost elsewhere, and the released model
does not use it. None of A1--A3 could have detected this failure, because the
benchmarks' own annotation never asserts the violated pairs in either
direction. Checking a model's output against a prior stated by its training
data is an inexpensive measurement that is not currently made.

\subsection{Negative results}
\label{app:closed:directions}

Each direction in \cref{tab:closed} was evaluated and not adopted on the basis
of the measurement listed. None supports a claim in the main text, and every
number is a single arm read against the 1.3\% noise floor.
\Cref{tab:resolution} reports one of them, evaluation resolution, in full.

\begin{table}[!htb]
\centering\small
\caption{Evaluation resolution for the released tower, trained multi-scale at
$[0.5, 1.5]\times 448$, without retraining. Test splits, ground-truth boxes,
graph-constrained.}
\label{tab:resolution}
\input{tables/resolution}
\end{table}

\begin{table}[p]
\centering\footnotesize
\caption{Directions evaluated and not adopted, with the measurement that
decided each.}
\label{tab:closed}
\begin{tabularx}{\linewidth}{@{}>{\raggedright\arraybackslash}p{2.9cm}>{\raggedright\arraybackslash\hsize=1.05\hsize}X>{\raggedright\arraybackslash\hsize=0.95\hsize}X@{}}
\toprule
direction & measurement & conclusion \\
\midrule
\multicolumn{3}{@{}l}{\sffamily\itshape modelling and training} \\
stochastic depth \citep{stochasticdepth} & train--validation gap reduced $8.8\times$; benchmark change approximately zero & the generalisation gap and the transfer gap are different quantities \\
WiSE-FT \citep{wiseft} toward the pre-trained backbone & monotone loss at every interpolation weight & the transfer cost is not backbone drift; a jointly trained head invalidates the premise \\
model soups over seeds & averaging same-recipe, different-seed checkpoints halves A1 & a new seed re-randomises the head, so the arms lie in mismatched basins; shared initialisation untested \\
masks as training input & $-18\%$ under the released box input; $+6.6\%$ only when masks are supplied at evaluation to a box-trained model & train on boxes, accept masks at inference \\
multiplicative (Tucker) pair queries & used by the network (norm grows), composite-neutral over two seeds; one robust cell, zero-shot PSG macro recall $+4$ to $+7\%$ & not adopted \\
head width 768 & $-2.7\%$ on the largest backbone; the predicate bank has an effective rank of 43.7 & 12M head parameters already suffice for a target of rank about 44 \\
\midrule
\multicolumn{3}{@{}l}{\sffamily\itshape text space and decoding} \\
CSLS hubness correction \citep{csls} at the synonym matcher & negative; the cosine threshold it was calibrated against accepted 0.6\% of true synonyms in the released space & every similarity threshold is calibrated to one embedding space \\
commit-late decoding & failed with three aggregators for one cause & no certified cross-lemma synonym resource exists to aggregate over \\
predicate box embeddings & entailment geometry on frozen queries trades against ranking: VG150 mR 0.37 / 0.18 / 0.13 against Spearman 0.17 / 0.55 / 0.60 as the entailment weight rises & a Pareto trade-off rather than an improvement; joint training untested \\
\midrule
\multicolumn{3}{@{}l}{\sffamily\itshape evaluation and deployment} \\
evaluation-resolution tiering & flat, and shrinks with model size (\cref{tab:resolution}); $1.5\times$ resolution yields $+10\%$ rare-bucket recall and nothing on micro recall or the spatial axis & multi-scale training already provides the robustness \\
letterboxing & corrects a genuine $5.2^\circ$ angle error; projective spatial axis $+0.1\%$ & the model does not use precise angles \\
CUDA-graph capture & $3.45\times$ faster and silently incorrect; graph compilation moves evaluation metrics by $40\times$ the noise floor & no reported number uses either \\
per-predicate calibration & fits in-domain and does not transfer & two global Platt parameters transfer, per-predicate ones do not \\
\bottomrule
\end{tabularx}
\end{table}

%% file: tables/ladder.tex
\begin{tabular}{llccccc}
\toprule
\sffamily\bfseries arm & \sffamily\bfseries scale & \sffamily\bfseries A1 & \sffamily\bfseries A2 & \sffamily\bfseries A3 & \sffamily\bfseries A6 & \sffamily\bfseries OVS-dev \\
\midrule
\ourrow released recipe & \note{full corpus} & 26.4 & 50.0 & 33.0 & 51.1 & \best{37.1} \\
recipe generation 4 & \note{full corpus} & 23.4 & 49.4 & 31.5 & 46.2 & 34.4 \\
full fine-tune, backbone lr $10^{-5}$ & \note{full corpus} & 27.1 & 46.7 & 35.1 & 32.6 & 34.1 \\
+ deformable read & \note{50k proxy} & 23.7 & 48.4 & 29.5 & 47.4 & 33.9 \\
LoRA backbone, 8 epochs & \note{full corpus} & 26.8 & 46.2 & 34.2 & 33.0 & 33.8 \\
recipe generation 3 & \note{full corpus} & 25.8 & 46.6 & 33.8 & 33.4 & 33.4 \\
+ deformable read, 8-head variant & \note{50k proxy} & 23.1 & 48.5 & 29.4 & 44.2 & 33.2 \\
+ Open Images share, 25\% & \note{50k proxy} & 21.7 & 43.2 & 28.9 & 46.0 & 31.9 \\
+ sigmoid auxiliary and box-token dropout & \note{50k proxy} & 21.7 & 47.3 & 26.9 & 46.1 & 31.7 \\
+ box-token dropout & \note{50k proxy} & 24.3 & 44.7 & 29.5 & 34.5 & 31.7 \\
sigmoid aux. and dropout, backbone interpolation 0.90 & \note{50k proxy} & 21.4 & 47.1 & 26.1 & 46.3 & 31.3 \\
full fine-tune, backbone lr $5\times10^{-5}$ & \note{50k proxy} & 24.7 & 43.4 & 30.4 & 31.2 & 31.1 \\
+ layer-wise lr decay and init sweep & \note{50k proxy} & 21.1 & 46.6 & 25.3 & 47.9 & 30.9 \\
sigmoid aux. and dropout, backbone interpolation 0.80 & \note{50k proxy} & 20.8 & 47.0 & 25.1 & 45.8 & 30.5 \\
LoRA backbone, 12 epochs & \note{50k proxy} & 23.2 & 43.9 & 27.9 & 32.4 & 30.2 \\
sigmoid aux. and dropout, backbone interpolation 0.65 & \note{50k proxy} & 19.2 & 46.4 & 23.0 & 45.0 & 28.7 \\
sigmoid aux. and dropout, backbone interpolation 0.50 & \note{50k proxy} & 17.2 & 45.3 & 20.8 & 44.3 & 26.5 \\
recipe generation 1 & \note{full corpus} & 15.9 & 35.8 & 13.7 & 25.1 & 19.7 \\
recipe generation 2 & \note{full corpus} & 13.4 & 33.7 & 5.3 & 22.2 & 11.9 \\
\bottomrule
\end{tabular}

%% file: tables/resolution.tex
\begin{tabular}{l ccc ccc ccc c}
\toprule
& \multicolumn{3}{c}{VG150} & \multicolumn{3}{c}{PSG} & \multicolumn{3}{c}{IndoorVG} & A6 \\
\cmidrule(lr){2-4}\cmidrule(lr){5-7}\cmidrule(lr){8-10}
eval px & R@50 & mR@50 & rare & R@50 & mR@50 & rare & R@50 & mR@50 & rare & AUC \\
\midrule
448 & 52.2 & 28.1 & 36.2 & 39.2 & 28.4 & 18.7 & 51.1 & 29.7 & 22.0 & 0.689 \\
560 & 52.3 & 27.7 & 29.0 & 39.5 & 28.7 & 18.7 & 51.5 & 29.9 & 22.2 & 0.686 \\
672 & 52.3 & 28.5 & 39.7 & 39.7 & 29.5 & 20.6 & 51.7 & 30.8 & 24.1 & 0.686 \\
\bottomrule
\end{tabular}

%% file: app/i_cost.tex
\section{Inference cost details}
\label{app:cost}

Cost was measured on three GPU classes under one protocol: batch 1 on real
VG150 test images, CUDA-event timed, median of 200 frames after 20 warm-up
frames. Batch 1 is the quantity relevant to deployment; batched throughput is
reported alongside it because the two disagree.

\begin{table}[h]
\centering\small
\caption{Cost of the released family, relation head only, with an open
vocabulary of 19{,}103 strings, in bf16. GFLOPs count dense operations at 448
px with 11 boxes and are GPU-independent. \note{The vocabulary matrix used
here is a shape-matched stand-in for the distilled student. Cost depends only
on the shape of that matrix, so the timings are exact, but nothing in this
table is an accuracy claim; \cref{tab:deploy} uses the actual bank.}}
\label{tab:latency}
\input{tables/latency}
\end{table}

\begin{table}[h]
\centering\small
\caption{\textbf{Cost of the released tower on three GPU classes}, with the
actual 243-predicate release vocabulary for the end-to-end rows and the full
19{,}103-string vocabulary for the relation-head rows, batch 1 unless stated.}
\label{tab:gpucost}
\input{tables/gpucost}
\end{table}

\paragraph{Dispatch-bound latency at batch 1}
\Cref{tab:latency} gives the cost of the released family and
\cref{tab:gpucost} the cost of the released tower across three GPU classes.
The three towers cost 19.3--20.0 ms on an A40 despite a $2.5\times$ range in
parameters and FLOPs between the ends of the ladder, so at batch 1 the model
is bound by kernel dispatch rather than by arithmetic. The ordering appears
only at batch 32 (201 / 188 / 130 img/s), which is why the argument for the
smaller tower rests on throughput rather than on batch-1 latency.

\paragraph{Cost of the open vocabulary}
Scoring against the full 19{,}103-string vocabulary instead of a
50-predicate one costs $0.7$--$0.8$ ms per frame on an A40 and, within noise,
nothing on the A100 and H100. At batch 32 the same head costs 11--22\% of
throughput on every tower. Supplying the vocabulary as an input therefore has
no cost on a live frame and a moderate cost on an offline batched workload.

\paragraph{Host dependence of batch-1 latency}
The A100 is the slowest of the three GPUs at batch 1 (27.5--28.4 ms against
18.9--20.0 ms on the A40 and H100) while delivering $1.9$--$2.0\times$ the
throughput of the A40 at batch 32. This is not contention, since it
reproduces on an idle second A100 node (27.2 / 28.1 / 27.2 ms). The A100 is
attached to a Zen3 host and the A40 and H100 to Zen4 hosts, and a
dispatch-bound workload is a host-CPU workload; the same pattern appears in
the host-bound graph-inference stage of the baseline. At batch 1, a latency
on a named GPU therefore does not identify an operating point unless the host
CPU is also named, which is why the figures here are given per GPU.

\begin{table}[h]
\centering\small
\caption{The deployment path of the released tower, end to end with an
open-vocabulary detector, batch 1, with the actual 243-predicate release
vocabulary. \note{The compiled rows are deployment figures only: compilation
changes the reduction order and moves evaluation metrics by far more than the
noise floor, so none of the accuracy tables uses compiled inference.}}
\label{tab:deploy}
\input{tables/deploy}
\end{table}

\paragraph{Compilation and concurrency}
\Cref{tab:deploy} times the same model end to end. Compilation contributes
more than any other change evaluated, at 20.3 ms on an A40 and 18.1 ms on an
H100 against 30.5 and 32.1 ms for eager execution, and it is the fastest
configuration on all three GPUs. Since the relation backbone reads only
pixels, it could in principle run concurrently with the detector. Measured,
that overlap recovers 2.6 ms on an A40 under eager execution and costs 27 ms
on an H100, and costs time on both once the model is compiled; the pattern
held across five configurations, two thread settings and three detectors, so
the sequential path is used. \Cref{tab:targets} reports the same pipeline on
each deployment device.

\begin{table}[h]
\centering\small
\caption{\textbf{Deployment targets} for the released ViT-S/16+ tower,
detector + relation head + decoding, batch 1; \emph{agreement} is the top-1 /
top-20 Jaccard agreement of the emitted triplets with the fp32 graph.
\note{Every row names its device; the CPU rows are measured on a cluster
node.}}
\label{tab:targets}
\resizebox{\linewidth}{!}{\input{tables/targets}}
\end{table}

%% file: tables/latency.tex
\begin{tabular}{lr ccc ccc}
\toprule
\sffamily\bfseries Backbone & \sffamily\bfseries GFLOPs & \multicolumn{3}{c}{\sffamily\bfseries batch 1, ms} & \multicolumn{3}{c}{\sffamily\bfseries batch 32, img/s} \\
\cmidrule(lr){3-5}\cmidrule(lr){6-8}
& & A40 & A100 & H100 & A40 & A100 & H100 \\
\midrule
ViT-S/16 & 80 & 19.5 & 27.6 & 18.9 & 201 & 387 & 708 \\
\ourrow ViT-S/16+ & 91 & 20.0 & 28.4 & 19.4 & 188 & 367 & 678 \\
ViT-B/16 & 200 & 19.3 & 27.5 & 18.9 & 130 & 261 & 504 \\
\bottomrule
\end{tabular}

%% file: tables/gpucost.tex
\begin{tabular}{llccc}
\toprule
\sffamily\bfseries Stage & \sffamily\bfseries Configuration & A40 & A100 & H100 \\
\midrule
relation head, batch 1 (ms) & bf16, 19{,}103-string vocabulary & 19.3 & 28.1 & 19.3 \\
relation head, batch 32 (img/s) & same & 188 & 367 & 678 \\
\midrule
detector + head + decode, batch 1 (ms) & PyTorch eager & 30.5 & 43.1 & 32.1 \\
\ourrow detector + head + decode, batch 1 (ms) & PyTorch, \texttt{torch.compile} & 20.3 & 24.4 & 18.1 \\
detector + head + decode, batch 1 (ms) & ONNX Runtime, CUDA & 29.7 & 27.6 & 21.2 \\
\bottomrule
\end{tabular}

%% file: tables/deploy.tex
\begin{tabular}{l cc cc cc}
\toprule
\sffamily\bfseries Configuration & \multicolumn{2}{c}{\sffamily\bfseries A40} & \multicolumn{2}{c}{\sffamily\bfseries A100} & \multicolumn{2}{c}{\sffamily\bfseries H100} \\
\cmidrule(lr){2-3}\cmidrule(lr){4-5}\cmidrule(lr){6-7}
& p50 ms & FPS & p50 ms & FPS & p50 ms & FPS \\
\midrule
eager, sequential & 30.5 & 33 & 43.1 & 23 & 32.1 & 31 \\
eager, concurrent backbone & 27.9 & 36 & 41.9 & 24 & 59.3 & 17 \\
\ourrow torch.compile, sequential & 20.3 & 49 & 24.4 & 41 & 18.1 & 55 \\
torch.compile, concurrent & 22.3 & 45 & 31.0 & 32 & 55.1 & 18 \\
CUDA graphs, concurrent & 21.5 & 47 & 29.9 & 33 & 50.8 & 20 \\
ONNX Runtime, CUDA EP & 29.7 & 34 & 27.6 & 36 & 21.2 & 47 \\
\bottomrule
\end{tabular}

%% file: tables/targets.tex
\begin{tabular}{lllrrc}
\toprule
\sffamily\bfseries Target & \sffamily\bfseries Runtime & \sffamily\bfseries Device & ms & FPS & \sffamily\bfseries agreement \\
\midrule
\ourrow GPU, A40 & PyTorch 2.13, torch.compile, bf16 & NVIDIA A40 (host AMD EPYC 9334) & 20 & 49 & --- \\
GPU, A40 & ONNX Runtime, CUDA EP & NVIDIA A40 & 30 & 34 & --- \\
CPU & OpenVINO IR, fp16 & Intel Xeon 6338, 8 threads & 137 & 7 & 0.955 / 0.971 \\
CPU & OpenVINO IR, int4 weights & Intel Xeon 6338, 8 threads & 126 & 8 & 0.727 / 0.771 \\
\bottomrule
\end{tabular}

%% file: app/j_prompts.tex
\section{Generation prompts}
\label{app:datagen}

The three passes of \cref{app:corpus} use the prompts below verbatim. They are
reproduced in full because the calibration study that selected them, coverage
over verify-before-assert, can only be interpreted against the actual text.

\subsection{Semantic pass}
\input{prompts/iter_20.tex}

\subsection{Grow pass}
\input{prompts/sgg26b_grow_v1.tex}

\subsection{Spatial pass}
\input{prompts/sgg26b_spatial_v3.tex}

%% file: prompts/iter_20.tex
\begin{Verbatim}
You are a precise visual scene analyst. Extract visually grounded relationships between the {n} objects marked with coloured numbered dots in the image.

Objects (number → label):
{object_list}

Follow these steps carefully:

STEP 1 — Observe what each object IS DOING: Before reasoning about pairs, note the activity, state, or role of every object.
  What is it doing? (grazing, running, sitting, flying, eating, carrying a load, casting a shadow, etc.)
  What is it interacting with in the scene? (the ground, another object, a person, food, etc.)

STEP 2 — For each pair, find the MOST INFORMATIVE relationship:
  Priority order — stop at the first that applies:
  (a) Activity / locomotion — what is one object doing relative to the other or its environment?
      Open vocabulary: use any precise verb phrase (grazing on, eating from, running across,
      standing on, walking through, casting shadow on, drinking from, pecking at, chasing,
      herding, pulling, pushing, swimming in, flying over, climbing, digging in …)
  (b) Physical interaction — direct manipulation, support, containment:
      holding, wearing, riding, carrying, sitting on, resting on, leaning against,
      attached to, part of, mounted on, hanging from, covering, surrounding, inside
  (c) Directed attention / social:
      looking at, watching, talking to, pointing at, following, leading

  Critical rules:
  • DO NOT output any spatial or directional predicate — spatial layout will be handled separately
  • Prefer open-vocabulary action phrases: describe WHAT objects are doing, not WHERE they are
  • Use DIRECT predicates — no "is", "are", "was", "were" prefix
  • Each (subject_id, object_id) pair may appear AT MOST ONCE in your output

STEP 3 — Select relations:
  • Aim for 6-10 relations total — quality over quantity
  • Every numbered object should appear in at least one relation
  • If an object genuinely has no informative relation to any other marked object, skip it
  • DO NOT pad with spatial or directional predicates

STRICTLY BANNED predicates — your output is invalid if you use any of these:
  has, near, next to, alongside, by, with, on, beside, and, same scene,
  overlapping, adjacent, associated, related, interacts, features, involves,
  appears, is positioned, is situated, is located, faces, supporting, displaying,
  above, below, behind, in front of, to the left of, to the right of, between,
  standing behind, standing in front of, standing beside, standing next to,
  sitting behind, sitting in front of, sitting beside,
  parked behind, parked in front of, parked beside,
  walking behind, walking in front of, flying above, hanging above,
  positioned, located, placed, arranged, oriented

STEP 4 — Output: After your reasoning, output a single JSON block with no extra text outside it:
```json
{{"relations": [{{"subject_id": <int>, "subject_label": "<str>", "predicate": "<str>", "object_id": <int>, "object_label": "<str>"}}]}}
```
\end{Verbatim}

%% file: prompts/sgg26b_grow_v1.tex
\begin{Verbatim}
Here is an image and a PARTIAL scene graph for it — relationships already found between the numbered objects.

Objects (number → label):
{object_list}

Relationships already found:
{draft}

Look carefully at the image and ADD relationships that are genuinely present but MISSING from the list above — find what the previous pass overlooked:
  • Objects that appear in FEW or NONE of the relationships above — what is each doing, or
    what is done to it?
  • A genuinely DIFFERENT relationship for a pair already listed (e.g. a rider both
    "sitting on" AND "holding" a horse) — but NEVER a synonym or rephrasing of one already there.
  • Any action, physical contact, support, use, or attention you can see that is not yet listed.

Rules:
  • Output ONLY NEW relationships. Do NOT repeat any relationship already in the list, and
    do NOT output the inverse of one already there.
  • Use the MOST PRECISE open-vocabulary predicate; direct form (no "is/are"); never use
    vague words (near, next to, with, on alone, has, positioned, located, situated).
  • Report only what is genuinely visible. If nothing is missing, output an empty list.
  • The dots, numbers and outlines are markers added to the photo — never describe them.

Output a single JSON block with ONLY the new relationships, nothing else:
```json
{{"relations": [{{"subject_id": <int>, "subject_label": "<str>", "predicate": "<str>", "object_id": <int>, "object_label": "<str>"}}]}}
```
\end{Verbatim}

%% file: prompts/sgg26b_spatial_v3.tex
\begin{Verbatim}
Here is an image with numbered objects and their SEMANTIC relationships already found.

Objects (number → label):
{object_list}

Semantic relationships already found:
{draft}

Your task: describe the SPATIAL LAYOUT of the scene — where the objects are relative to each other. Choose the pairs whose arrangement best DEFINES the scene's layout, most informative first.

Spatial vocabulary (pick the most precise fit — close natural synonyms are also fine):
  depth        in front of / behind                    → who is closer to the camera, who occludes whom
  contact      on / resting on / leaning against / hanging from / attached to
                                                       → physical support or contact you can SEE
  proximity    next to / beside / near                 → clearly close together in the scene
  containment  inside                                  → one object genuinely within another (food inside a bowl)
  vertical     above / below / beneath / on top of
  horizontal   to the left of / to the right of

Rules:
  1. PREFER depth, contact and proximity — they say the most about a scene. Use
     "to the left of" / "to the right of" only when the horizontal offset is the single
     most informative thing about that pair.
  2. COVER the scene: every major object should appear in at least one layout relation.
     Skip only pairs whose arrangement is genuinely ambiguous or adds nothing.
  3. One relation per pair, judged from the VIEWER's perspective. Never also output the
     inverse (if A is in front of B, do not add B behind A).
  4. Do NOT use "between" (it needs two reference objects). Do NOT use action verbs —
     actions were covered above; this pass is layout only.
  5. Do NOT place body-part objects (human face, hand, arm, leg …) — their position is
     implied by the person they belong to.
  6. Do NOT repeat a pair already covered above unless the layout adds genuinely new
     information (e.g. "man riding horse" already implies he is on it).
  7. Target 6–12 relations; fewer only for genuinely sparse scenes (2–3 objects).
  8. The dots, numbers and outlines are markers added to the photo — never describe them.

Output a single JSON block with ONLY the new spatial relationships, nothing else:
```json
{{"relations": [{{"subject_id": <int>, "subject_label": "<str>", "predicate": "<str>", "object_id": <int>, "object_label": "<str>"}}]}}
```
\end{Verbatim}